%% file: main.tex
\documentclass[runningheads]{llncs}

\usepackage{eccv}

\usepackage{eccvabbrv}

\usepackage{graphicx}
\usepackage{booktabs}
\usepackage{multirow} 
\usepackage{wrapfig}
\usepackage{tabularx}
\usepackage{makecell}
\usepackage{arydshln}
\usepackage{multibib}
\usepackage{array}
\newcites{supp}{Supplementary References}

\usepackage[accsupp]{axessibility}  
\usepackage[pagebackref,breaklinks,colorlinks,citecolor=eccvblue]{hyperref}

\usepackage{orcidlink}

\begin{document}

\title{ProGVC: Progressive-based Generative Video Compression via Auto-Regressive Context Modeling} 

\titlerunning{Preprint}

\author{Daowen Li\inst{1}$^{\dagger}$ \and
Ruixiao Dong\inst{1}$^{\dagger}$ \and
Kai Li\inst{1}$^{*}$ \and
Ying Chen\inst{1}$^{*}$ \and
Ding Ding\inst{1} \and
Li Li\inst{2}}

\authorrunning{D.~Li, R.~Dong \etal}
% If there are more than two authors, '\etal' is used.

\institute{
Alibaba Group, China \\
\email{\{lidaowen.ldw,dongruixiao.drx,kaishi.lk,YingChen,liangjie.dd\}@taobao.com}\\
\and
Institute of Artificial Intelligence, Hefei Comprehensive National Science Center\\
\email{lil1@ustc.edu.cn}
}

\maketitle

\begingroup
\renewcommand\thefootnote{}
\footnotetext{$^{\dagger}$ Equal contribution.}
\footnotetext{$^{*}$ Corresponding author.}
\endgroup
\setcounter{footnote}{0}

\begin{abstract}
Perceptual video compression leverages generative priors to reconstruct realistic textures and motions at low bitrates. 
However, existing perceptual codecs often lack native support for variable bitrate and progressive delivery, and their generative modules are weakly coupled with entropy coding, limiting bitrate reduction. 
Inspired by the next-scale prediction in the Visual Auto-Regressive (VAR) models, we propose \textbf{ProGVC}, a \textbf{Pro}gressive-based \textbf{G}enerative \textbf{V}ideo \textbf{C}ompression framework that unifies progressive transmission, efficient entropy coding, and detail synthesis within a single codec. 
ProGVC encodes videos into hierarchical multi-scale residual token maps, enabling flexible rate adaptation by transmitting a coarse-to-fine subset of scales in a progressive manner. 
A Transformer-based multi-scale autoregressive context model estimates token probabilities, utilized both for efficient entropy coding of the transmitted tokens and for predicting truncated fine-scale tokens at the decoder to restore perceptual details. 
Extensive experiments demonstrate that as a new coding paradigm, ProGVC delivers promising perceptual compression performance at low bitrates while offering practical scalability at the same time.
\end{abstract}

\section{Introduction}
\label{sec:intro}
Driven by the explosive growth of multimedia consumption, advanced video compression technologies have become increasingly essential to reduce the bitrate cost of storing and transmitting video data. 

Depending on the target application, video compression can be designed for human viewing or for machine-oriented visual analysis~\cite{tian2024free, tian2026smc,duan2020video}, with this work focusing on the former. Traditional video compression standards, such as H.264/AVC~\cite{wiegand2003overview}, H.265/HEVC~\cite{sullivan2012overview} and H.266/VVC~\cite{bross2021overview}, have achieved remarkable success by employing carefully designed modules optimized towards pixel-wise distortion metrics (PSNR, SSIM~\cite{1284395}, MS-SSIM~\cite{1292216}, \textit{etc}.). In recent years, end-to-end Neural Video Compression (NVC) models, including ~\cite{lu2019dvc, li2023neural, li2024neural, sheng2025bi, bian2025augmented} have also demonstrated competitive performance against traditional standards. While these codecs are effective at preserving signal fidelity, they often produce visually unpleasing reconstruction results at low bitrates. These annoying visual artifacts include blurring, blocking, and ringing, indicating that human perception diverges from such distortion measures.

To improve perceptual quality, researchers have explored generative video compression, leveraging powerful generative models to synthesize realistic details. GAN-based methods~\cite{zhang2021dvc, yang2022perceptual,li2023high} adopted generative adversarial network (GAN)~\cite{goodfellow2020generative}, which yielded sharp-looking results but often suffered from unstable adversarial training and low-fidelity reconstructions. Recently, the advent of Vision Foundation Models, particularly diffusion models~\cite{rombach2022high, peebles2023scalable, podell2023sdxl}, has inspired diffusion-based generative codecs~\cite{ma2025diffusion, mao2025generative, ma2025diffvc} that exploit powerful priors to generate realistic textures and motions. 

Despite these advances, existing perceptual codecs still face several limitations. First, most methods lack support for scalability and variable bitrate adaptation, which are crucial under fluctuating network bandwidth. Second, diffusion-based codecs typically use diffusion model as decoding-time denoiser for the transmitted pixels or latents. Since the diffusion process is largely decoupled from entropy coding, spatio-temporal priors of diffusion model are less utilized to remove signal redundancy. Moreover, diffusion model requires iterative denoising, which increases decoding latency and hinders real-time applications.

In this paper, we aim to address these issues by introducing Visual Auto-Regressive (VAR)~\cite{tian2024visual, han2025infinity, liuinfinitystar} models for generative video compression. The next-scale prediction mechanism in VAR utilizes a multi-scale residual quantizer to encode a video sequence into hierarchical visual token maps (\ie, scales). These scales are then predicted progressively utilizing a Transformer-based Auto-Regressive (AR) model. This coarse-to-fine paradigm naturally supports scalability and bitrate adaptability: the encoder can transmit the first few coarse scales and progressively refine quality by sending additional high-frequency details. Moreover, the autoregressive model provides accurate conditional probabilities for each token that can directly serve as an entropy context model. Furthermore, unlike diffusion models that rely on iterative denoising loops, VAR enables efficient sampling in much fewer sampling steps, leading to substantially lower decoding latency.

Motivated by these insights, we propose \textit{ProGVC} (Progressive-based Generative Video Compression via Auto-Regressive Context Modeling), a perceptual video coding paradigm that unifies progressive scalability, efficient entropy modeling, and detail synthesis in a single autoregressive pipeline. Specifically, ProGVC employs a causal video VAE to extract continuous intra frame features for the first intra frame and inter frame features for subsequent frames. At encoder, a multi-scale residual quantization scheme encodes these features into $K$ hierarchical multi-scale discrete token maps. To achieve scalability and variable bitrate control, the encoder transmits all intra frame scales, as they are crucial for leveraging temporal priors and incur only a small bitrate overhead, and the first $k$ inter scales ($k \le K$), where the different choices of $k$ correspond to different bitrates. Moreover, to maximize the compression efficiency for the transmitted scales, we introduce the multi-scale autoregressive context model that acts as the entropy context model. For intra frame scales, the model captures cross-scale spatial dependencies; for inter frame scales, it explicitly conditions on both previously decoded intra and inter frame scales to exploit global spatio-temporal redundancy. The obtained discrete distributions for each token then drive arithmetic coding to minimize bitrate. At the decoder, ProGVC decodes the transmitted scales losslessly and utilizes the autoregressive model to generate the discarded $K-k$ high-frequency inter scales, enabling high-quality perceptual reconstruction with minimal additional rate.

To the best of our knowledge, this is the \textit{first} study to systematically investigate generative video coding built upon an visual autoregressive generative model. We hope this work can shed some light on future research in this field.
In summary, our core contributions are as follows:
 \begin{itemize}
     \item We introduce ProGVC, the first progressive generative video coding paradigm built upon the visual autoregressive model, enabling native scalability and variable bitrate by transmitting different subsets of hierarchical scales.
     \item We design a task-specific multi-scale autoregressive context model for the intra and inter token coding, unifying accurate probability estimation for entropy coding of the transmitted tokens and generative sampling for the un-transmitted details within a single framework, thereby reducing bitrate while preserving reconstruction quality.
     \item Extensive experimental results demonstrate that ProGVC exhibits superior perceptual quality over fidelity-oriented approaches and achieves competitive compression performance against existing perceptual video codecs.
 \end{itemize}

\section{Related works}

\subsection{Perceptual Video Compression}
To improve perceptual quality of the reconstructed videos, generative models have been utilized in perceptual video compression methods. Among them, GAN-based codecs adopt adversarial learning to produce more realistic reconstructions. Mentzer \etal~\cite{mentzer2022neural} first introduced GAN into video compression. Yang \etal~\cite{yang2022perceptual} designed recurrent conditional GAN and adversarial loss functions. Qi \etal~\cite{qi2025generative} presented a latent generative coding scheme, performing transform coding with a vector-quantized variational auto-encoder.

Recently, diffusion-prior-based coding paradigms have emerged, where the diffusion foundation model is used as the denoiser in latent or pixel space to synthesize missing details and thus enhance the perceptual quality of the reconstructions. Li \etal~\cite{li2024extreme} proposed a hybrid approach that combined image compression and a diffusion model for video compression. Ma and Chen~\cite{ma2025diffusion} integrated the Stable Diffusion model~\cite{rombach2022high} as a latent denoiser into the conditional video codec. Mao \etal ~\cite{mao2025generative} leveraged a video diffusion transformer model to jointly enhance intra and inter frame latents through sequence-level denoising.

Despite their encouraging visual results, prior perceptual video codecs have limited practicality, mainly due to the lack of native support for variable bitrate and scalable bitstream, resulting in the fact that they mostly require training separate models for different rates. In addition, diffusion models are typically leveraged as the decoder-side refiners to suppress quantization noise, rather than being tightly coupled with entropy modeling. Consequently, their powerful spatio-temporal priors are not directly translated into reducing transmitted bits. Furthermore, diffusion-based decoding typically relies on multiple iterative denoising steps, which increases computational cost and limits their applicability in low-delay or real-time streaming scenarios.

\subsection{Visual Auto-Regressive Generative Models}
Unlike diffusion models which rely on iterative denoising, Auto-Regressive (AR) models generate images and videos by sequentially predicting visual tokens conditioned on prefix context. Early raster-scan AR methods operate directly in pixel domain~\cite{van2016conditional, van2016pixel}. To improve efficiency and generation quality, Taming Transformers~\cite{esser2021taming} and VideoGPT~\cite{yan2021videogpt} proposed a two-stage framework: first leverages a discrete tokenizer mapping visual data to latent tokens via a learned codebook, which is conceptually analogous to quantization in visual compression, and then employs the Transformer that models token distributions autoregressively, resembling entropy modeling in compression. Nevertheless, traditional AR models suffer from structural degradation caused by the fixed raster-scan order, which limits their generative fidelity and practical applicability.

Visual Auto-Regressive (VAR) model~\cite{tian2024visual} alleviates this issue with a hierarchical coarse-to-fine strategy based on a multi-scale tokenizer, progressively refining visual details via next-scale prediction. Its successors, Infinity~\cite{han2025infinity} and InfinityStar~\cite{liuinfinitystar}, extend this paradigm to text-to-image and text-to-video generation, achieving performance comparable to or surpassing leading diffusion-based approaches. Recently, VAR has been adapted to downstream tasks such as controllable generation~\cite{liu2025scaleweaver,dong2026echogen},  image super-resolution~\cite{quvisual} and image compression~\cite{zhang2026autoregressivebased}.
Despite these advances, the potential of visual autoregressive models for video compression remains unexplored. Notably, the hierarchical next-scale prediction mechanism of VAR naturally lends itself to a progressive and scalable perceptual video compression framework. Motivated by this, we adopt the visual autoregressive model as the backbone of our perceptual video compression framework.

\section{Method}
This section first describes the overall framework (\cref{sec:overallframework}), highlighting on how scalable compression and variable bitrate adaptation are implemented within the visual autoregressive model. We then elaborate on the Multi-scale Residual Quantization module (\cref{sec:quantization}) for multi-scale token maps generation, followed by the Multi-scale Auto-Regressive Context Model (\cref{sec:transformer}) for efficient token transmission.

\subsection{Overall Framework}
\begin{figure*}[t]
    \centering
    \includegraphics[width=0.9\linewidth]{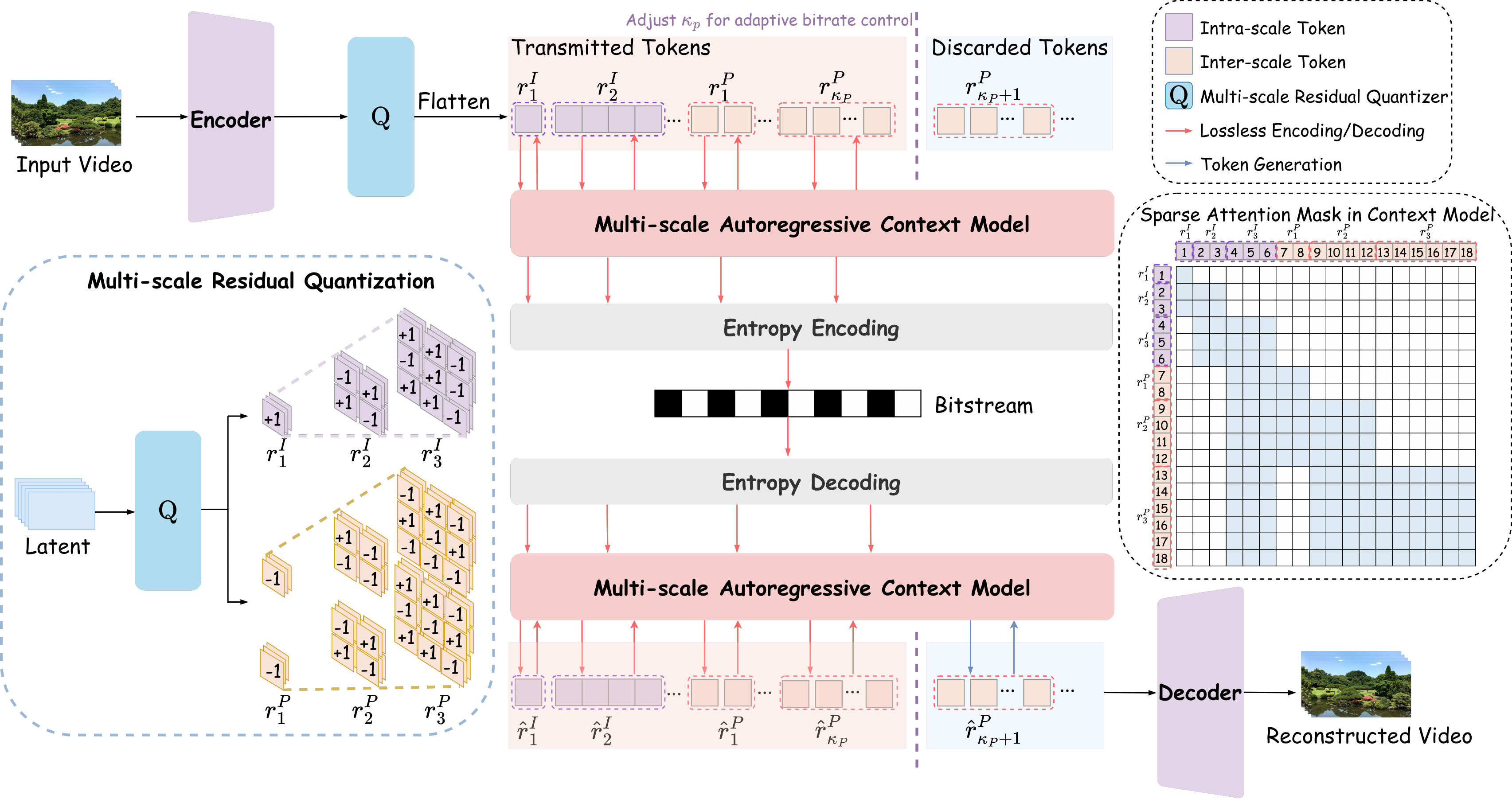}
    \caption{
    \textbf{Overview of the proposed ProGVC video compression framework.} We design a progressive generative video compression framework based on the visual autoregressive model. 
    The left panel illustrates the output pyramid structure of the multi-scale residual quantization. Besides, the sparse attention mask used in the context model is shown on the right.}
    \label{fig:framework}
\end{figure*}

\label{sec:overallframework}
We design ProGVC, a progressive perceptual video compression framework based on the visual autoregressive model, as illustrated in \cref{fig:framework}.
Specifically, a causal video VAE encoder first maps the input video clip into a continuous latent. A multi-scale residual quantizer then discretizes this latent into multi-scale token maps.
During encoding, the intra scales and the first $k$ low-frequency inter token maps are losslessly encoded considering their importance in temporal consistency and global layout construction, while higher-frequency inter scales are discarded and later generated at the decoder.
We can adjust $k$ for adaptive bitrate control.
A multi-scale autoregressive context model estimates the distribution of each token, which is used both for entropy coding of the transmitted tokens and for generation of the discarded scales. Finally, the VAE decoder reconstructs the video from the recovered multi-scale token maps.

\noindent\textbf{Continuous Feature Extraction.} For an input video clip $V \in \Bbb R^{(1+T)\times W\times H\times 3}  $, we regard the first frame as the intra frame, and the remaining frames are inter frames.
A causal VAE encoder $\mathcal{E}$ encodes $V$ into continuous spatio-temporal latent features by $f=\mathcal{E}(V)=\{f_t\}_{t=1}^{1+\frac{T}{4}}$, with $ f_t \in \mathbb{R}^{\frac{W}{4}\times\frac{H}{4}\times 64}$. 
We denote the intra latent features corresponding to the intra frame as $f^{I} = f_1$, and the inter latent features for inter frames as $f^{P}=\{f_t\}_{t=2}^{1+\frac{T}{4}}$. 

\noindent\textbf{Multi-scale Residual Quantization.} 
The employed multi-scale residual quantization converts the continuous latent features into multi-scale residual token maps:
\begin{equation}
R^{(\cdot)}=\mathrm{Q}(f^{(\cdot)})=\{r_k^{(\cdot)}\}_{k=1}^{K},\quad (\cdot) \in \{I, P\}.
\end{equation}
Here, $\mathrm{Q}$ denotes the multi-scale residual quantization, and $R^{I}$ and $R^{P}$ are the resulting multi-scale intra and inter token maps (\ie, scales) respectively. The spatial resolution of the $ k $-th scale $ r_k^{(\cdot)} $ increases with $ k $, with higher scales progressively capturing more high-frequency details. Following Binary Spherical Quantization (BSQ) \cite{zhao2024image}, each token in $ r_k^{(\cdot)} $ is a binary code of length $ L_k $.

As $k$ increases, the upsampling and element-wise addition result of $\{ r_i^{(\cdot)}\}_{i=1}^{k} $  progressively approximates the original continuous features $ f^{(\cdot)}$. The detailed procedure to obtain multi-scale residual token maps is presented in \cref{sec:quantization}.

\noindent\textbf{Scalability and Adaptive Bitrate Control.} 
By transmitting multi-scale token maps from coarse to fine, ProGVC first reconstructs global structure and layout from the base scales, and then progressively adds localized, high-frequency details from the later scales. This progressive structure provides explicit scalability: bitstreams containing more scales yield higher reconstruction quality, while truncating the tail scales produces a valid, lower-rate reconstruction. 
Selecting different subsets of scales for transmission also enables adaptive bitrate control without retraining or modifying the compression model. In practice, we primarily adjust the bitrate by truncating inter scale tokens, since they constitute the majority of transmitted tokens and considering that the intra scales are vital for temporal alignment in modeling inter scales distribution. The discarded higher-frequency inter scales are generated conditioned on the already reconstructed scales at the decoder side. Then the multi-scale summation over all scales is applied to reconstruct the continuous features for final video decoding.

\noindent\textbf{Entropy Coding based on Context Modeling.}
For the transmitted scales, ProGVC further improves compression efficiency through \emph{lossless} entropy coding. Since each token is represented by binary codes, the resulting bit sequence naturally supports arithmetic coding, which reduces the bitrate given accurate probability estimation.
Here we design a multi-scale autoregressive context model that exploits the spatio-temporal priors among scales to estimate the distribution of each scale conditioned on previous scales. Modeling the distributions of intra and inter scales is discussed separately as follows.

\textit{Intra context modeling.}
Following the autoregressive factorization across scales in VAR, the joint distribution of intra scales is factorized as
\begin{equation}
p(r^{I}_{1},\ldots,r^{I}_{K})
=\prod_{k=1}^{K} p\!\left(r^{I}_{k}\mid \{r^{I}_{k'}\}_{k'<k}\right),
\end{equation}
where \(r^{I}_{k}\) denotes the \(k\)-th intra scale. Given the conditional probability \(p(r^{I}_{k}\mid \{r^{I}_{k'}\}_{k'<k})\), the bitstream of the \(k\)-th intra scale, denoted as $\mathcal{B}_k^{I}$, is generated via entropy coding.

\textit{Inter context modeling.}
For inter scales, only the first \(\kappa_P\) scales (\(\kappa_P\le K\)) are transmitted. Let \(r^{P}_{k}\) denotes the \(k\)-th inter scale, and \(R^{I}\!=\!\{r^{I}_{k}\}_{k=1}^{K}\) represents the intra scales. The joint distribution of inter scales is factorized as
\begin{equation}
p(r^{P}_{1},\ldots,r^{P}_{\kappa_P})=
\prod_{k=1}^{\kappa_P} p\!\left(r^{P}_{k}\mid \{r^{P}_{k'}\}_{k'<k},\, R^{I}\right).
\end{equation}
That is, each inter scale is conditioned on all previously reconstructed inter scales as well as the intra scales. Analogously, the bitstream of the \(k\)-th inter scale tokens, $\mathcal{B}_k^{P}$, is generated according to the corresponding conditional distribution.

The detailed architecture of the multi-scale autoregressive context model is described in \cref{sec:transformer}.

\noindent\textbf{Entropy Decoding and Token Generation for Video Decompression.}
At the decoder, the transmitted intra and inter scale tokens are first losslessly recovered from the bitstreams via entropy decoding, denoted as $\{\hat r_k^{I}\}_{k=1}^{K}$ and $\{\hat r_k^{P}\}_{k=1}^{\kappa_P}$, using the same autoregressive probability factorization as in encoding. 

The discarded higher-frequency inter scales are then synthesized through token generation.
Specifically, the discarded inter scales (\(k>\kappa_P\)) are generated conditioned on the already reconstructed scales with the same conditional distribution modeling mechanism, while selecting the most probable entries as
\begin{equation}
\hat r_k^{P}=\arg\max_{\hat r_k^{P}} \; p\!\left(\hat r^{P}_{k}\mid \{\hat r^{P}_{k'}\}_{k'<k},\, \hat R^{I}\right), \quad k=\kappa_P\!+\!1,\ldots,K.
\end{equation}

After obtaining the complete set of hierarchical scales, the intra and inter latent features $\hat f^{I}$ and $\hat f^{P}$ are reconstructed via multi-scale summation. Finally, the reconstructed video $\hat V$ is obtained by feeding these latent features to the VAE decoder \(\mathcal{D}\) as $\hat V=\mathcal{D}(\hat f^{I},\hat f^{P})$.

\subsection{Multi-scale Residual Quantization}
\label{sec:quantization}
To discretize the continuous latent features \(f^{(\cdot)}\), we adopt a multi-scale residual quantization scheme that produces \(K\) hierarchical scales \(\{r_k^{(\cdot)}\}_{k=1}^{K}\). 

Specifically, at scale \(k\), the residual feature \(e_k^{(\cdot)}\), which is initialized with the continous feature \(f^{(\cdot)}\) at $k=1$, is first downsampled into token vectors $\tilde e_k^{(\cdot)}$:
\begin{equation}
\tilde e_k^{(\cdot)} = D_k\!\left(e_k^{(\cdot)}\right),\tilde{e}_k^{(\cdot)} \in \mathbb{R}^{w_k\times h_k \times T'\times L_k}, e_k^{(\cdot)} \in \mathbb{R}^{W/4\times H/4 \times T'\times L_k},
\end{equation}
where \(D_k\) denotes the downsampling operator at scale \(k\), $(w_k, h_k)$ represents the spatial resolution of $k$-th token map, $T'$ and $L_k$ denote the temporal and channel dimensions, respectively.
Each token vector \(v\in\mathbb{R}^{L_k}\) from \(\tilde e_k^{(\cdot)}\) is discretized using Binary Spherical Quantization (BSQ)~\cite{zhao2024image}, which binarizes each element $v_s$ of the token according to its sign and normalizes it by the token dimension:
\begin{equation}
\mathrm{BSQ}(v_s)=\frac{\mathrm{sign}(v_s)}{\sqrt{L_k}}, \qquad
\mathrm{sign}(v_s)\in\{-1,+1\}.
\end{equation}
By applying BSQ to all $k$-th scale tokens, the module obtains discrete residual token map $r_k^{(\cdot)}$. \(r_k^{(\cdot)}\) is then upsampled back to the latent resolution using the scale-specific upsampling operator \(U_k\), and further subtracted from the current residual feature to generate the residual feature $e_{k+1}^{(\cdot)}$ for the next scale:
\begin{equation}
e_{k+1}^{(\cdot)} = e_k^{(\cdot)} - U_k\!\left(r_k^{(\cdot)}\right), \qquad k=1,\ldots,K.
\end{equation}

Through this hierarchical residual quantization process, the latent representation is progressively refined from coarse to fine scales, natively and explicitly supporting a scalable progressive generative video coding paradigm.

\subsection{Multi-scale Auto-Regressive Context Modeling}\label{sec:transformer}

\begin{figure*}[t]
    \centering
    \includegraphics[width=0.96\linewidth]{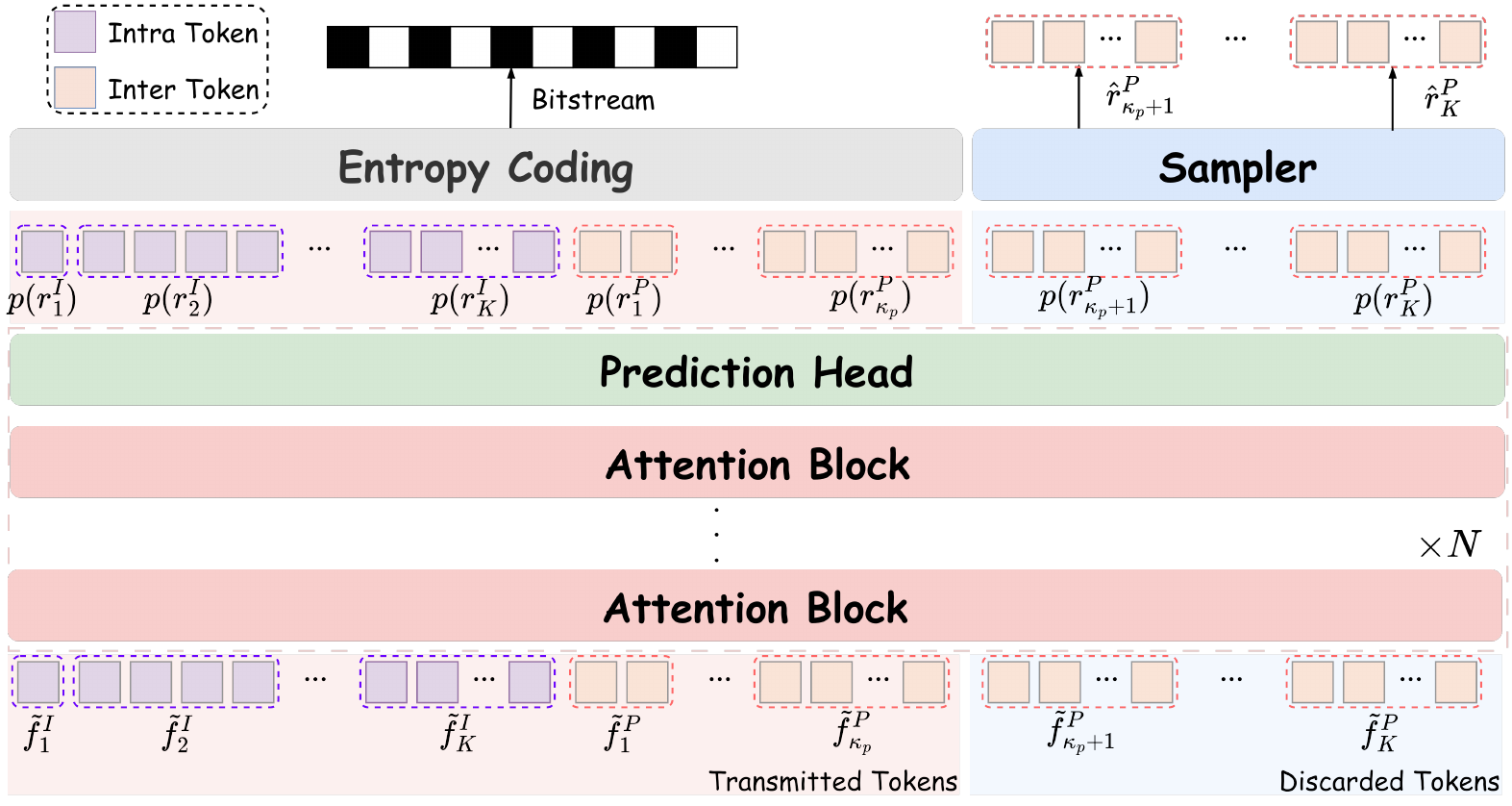}
    \caption{
    \textbf{Overview of the multi-scale autoregressive context model}. The transformer network predicts both intra scale and inter scale tokens in an autoregressive manner. The resulting probability distributions are subsequently utilized for entropy coding and for sampling discarded tokens during reconstruction.}
    \label{fig:transformer}
\end{figure*}

In this stage, our transformer-based autoregressive context model estimates the distribution of each scale in an autoregressive manner. 
\cref{fig:transformer} exhibits the workflow. Specifically, to estimate the probability of the 
$k$-th scale $r_k^{(\cdot)}$, we first construct an aggregated multi-scale representation $\tilde{f}_k^{(\cdot)}$ that serves as input to the autoregressive model for the current scale,

\begin{equation}
    \tilde{f}_k^{(\cdot)}=D_k(\sum_{i=1}^kU_i(r_i^{(\cdot)})).
\end{equation}

The distributions of current scale are then predicted by injecting information from the prefix scales through masked self-attention. In particular, for each attention block, the temporal-spatial priors are incorporated as:

\begin{equation}
\begin{aligned}
\mathcal{Q}&=Z_k^{(\cdot)}\mathit{W^q}, \, \mathcal{K}=\mathrm{concat}\left(Z_k^{(\cdot)}, C_k\right)\mathit{W^k}, \, \mathcal{V}=\mathrm{concat}\left(Z_k^{(\cdot)}, C_k\right)\mathit{W^v}, \\
Z_k^{(\cdot)'}&=\mathrm{Attention}\left(\mathcal{Q,K,V}, \mathrm{Mask}\right)=\mathrm{Softmax}\left(\mathrm{Mask}\left(\mathcal{QK}^\top/\sqrt{d}\right)\right)\mathcal{V},
\end{aligned}
\end{equation}

where $W^q,W^k,W^v$ denote the query, key, and value projection matrices; $Z_{k}^{(\cdot)}$ and $Z_{k}^{(\cdot)'}$ are the input and output of this attention block corresponding to $k$-th scale input $\tilde{f}_k^{(\cdot)}$, respectively. The context set $C_k$ depends on the token type. For intra scales, $C_k$ consists of previous intra scale features $\{Z_{k'}^{I}\}_{k'<k}$; and for inter scales, the context further includes prefix inter scale features $\{Z_{k'}^{P}\}_{k'<k}$ together with intra scale features $\{Z_{k'}^{I}\}_{k'<K}$. The attention mask $\mathrm{Mask}$ determines which tokens in $Z_k^{(\cdot)}$ are visible during attention, restricting each token to attend only to its effective prefix context.
Multiple masked attention blocks are stacked, followed by an MLP-based prediction head to estimate the discrete token distribution. 
For transmitted scales, predicted probabilities are used for lossless entropy coding, whereas truncated scales are generated by selecting the most probable discrete entries according to the estimated distribution.

\begin{figure*}[t]
    \centering
    \includegraphics[width=0.9\linewidth]{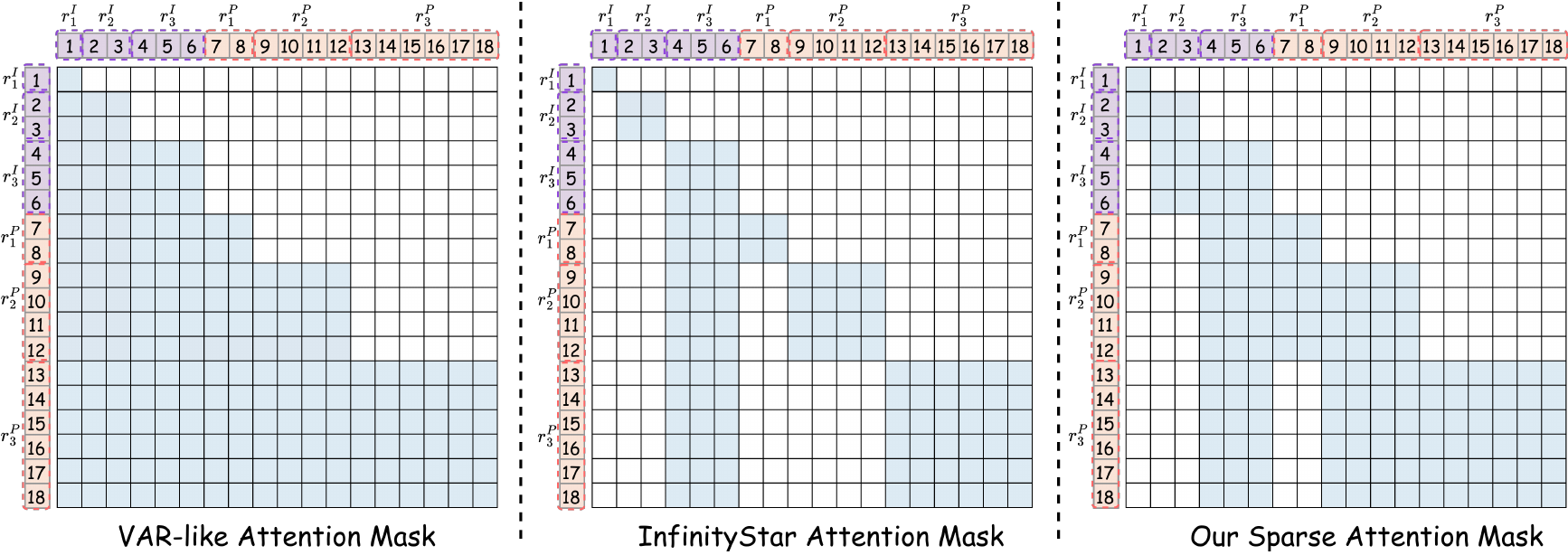}
    \caption{
    \textbf{Illustration of three attention mask designs: VAR-like attention (left), InfinityStar attention (middle), and our sparse attention (right).} For visualization simplicity, we illustrate the masks using three spatial scales (1, 2, 3) and a temporal length of $T=2$ for inter scale tokens.
    }
    \label{fig:mask}
\end{figure*}
\noindent\textbf{Sparse attention}. 
Theoretically, incorporating richer conditional information can reduce conditional entropy~\cite{bengio2003neural}, thereby lowering the expected bitrate. However, naively expanding the conditioning context significantly increases the computational cost of autoregressive video modeling, as the context grows rapidly along both spatial and temporal dimensions. Moreover, the aggregated input \(\tilde f_k^{(\cdot)}\) at scale \(k\) already summarizes information from coarser scales, making full attention over all historical tokens partially redundant~\cite{voronov2024switti}.

As shown in \cref{fig:mask}, Infinity~\cite{han2025infinity} employs a block-wise causal mask that allows each scale to attend to all historical tokens, which becomes computationally expensive for long sequences. InfinityStar~\cite{liuinfinitystar} reduces complexity by adopting a more restrictive attention mask (middle of \cref{fig:mask}); however, the limited receptive field weakens context utilization and may increase the bitrate.

To balance compression efficiency and computational complexity, we design a sparse attention mask used in both the encoding and decoding process, as shown in the right of \cref{fig:mask}. First, this attention mask allows each scale to attend to itself with its immediately preceding scale. By stacking multiple attention layers, long-range dependencies are progressively captured without explicitly attending to the full history. Furthermore, to leverage temporal priors and maintain temporal consistency in reconstructions, inter scale tokens are allowed to attend to the largest intra scale, which retains rich fine-grained details and exhibits stronger temporal coherence with inter scales. As shown in \cref{sec:ablation}, this sparse attention design achieves a favorable trade-off between compression efficiency and computational overhead.

\section{Experiments}

\subsection{Setup}
\noindent\textbf{Dataset}.
We train ProGVC on the Pexels dataset. To construct a high-quality training corpus, we compute the aesthetic and clarity scores for each video and retain only samples with scores above 4.5 and 0.65, respectively. Videos shorter than 10 seconds are discarded, and only those with aspect ratios close to standard 720p resolution are kept, resulting in a large-scale, high-quality corpus of approximately 480K videos. Evaluation is conducted on widely used benchmarks, including Xiph~\cite{xiph}, HEVC Class B~\cite{sullivan2012overview}, and MCL-JCV~\cite{wang2016mcl}. Since ProGVC is trained primarily on 720p content following InfinityStar, the HEVC Class B and MCL-JCV datasets are downsampled to 720p.

\noindent\textbf{Training Details.} The base configurations of the causal VAE and the multi-scale residual quantization are inherited from InfinityStar. We finetune the VAE jointly with the quantization module, largely following prior work~\cite{wang2024omnitokenizer,zhao2024image,liuinfinitystar}. Specifically, training is conducted on $256\times256\times81$ video clips for 10K iterations with a batch size of 2. The training objective combines entropy loss, commitment loss, $l_2$ reconstruction loss, GAN loss, and LPIPS perceptual loss, weighted by 0.1, 0.25, 1, 0.01 and 4, respectively. The learning rate is set to $5\times 10^{-5}$.

We train the multi-scale autoregressive context model for 60k iterations, utilizing the AdamW optimizer~\cite{loshchilov2017decoupled} with batch size of 8. The learning rate is set to $5\times 10^{-5}$, with momentum coefficients as $(\beta_1, \beta_2) = (0.9, 0.97)$. Following the InfinityStar training protocol, we randomly drop the later inter scales to improve training efficiency. All videos are resized and randomly cropped to a resolution of 720p. Each training sample consists of 81 frames.

\noindent\textbf{Metrics.}
To evaluate perceptual quality, full-reference quality metrics are adopted, including DISTS~\cite{ding2020image} and LPIPS~\cite{zhang2018unreasonable}, along with no-reference quality metric of NIQE~\cite{mittal2012making}. PSNR is used to evaluate signal fidelity. And t-LPIPS~\cite{chu2020learning} is adopted to measure temporal consistency. We use kilobits per second (kbps) to measure the bitrate cost for compressing a video sequence within one second.

\noindent\textbf{Baselines.}
We compared ProGVC against three categories of video compression methods: a) traditional video codec based on the latest VVC standard, \ie, VTM‑17.0 \cite{bross2021overview}; b) leading fidelity‑oriented neural video codecs, including DCVC-FM \cite{li2024neural} , DCVC-B \cite{sheng2025bi} and SEVC \cite{bian2025augmented}; c) perceptual video codecs, represented by PLVC \cite{yang2022perceptual}, which serves as the strong reproducible generative baseline. Among these baselines, the fidelity-oriented codecs support variable-rate control but not scalable bitstreams, whereas PLVC supports neither scalability nor adaptive rate control. We adopt the random access configuration with GOP size of 32 and one intra frame for VTM-17.0, while for the other baseline methods we use their default configurations. For each test sequence, we evaluate the first 81 frames. All methods are tested in the RGB domain with the same RGB mode to ensure a fair comparison. The evaluated bitrate range is set to 300-2000 kbps. Additional implementation details are provided in the supplementary material.

\begin{figure*}[t]
  \centering
  \begin{subfigure}{0.25\linewidth}
    \includegraphics[width=\textwidth]{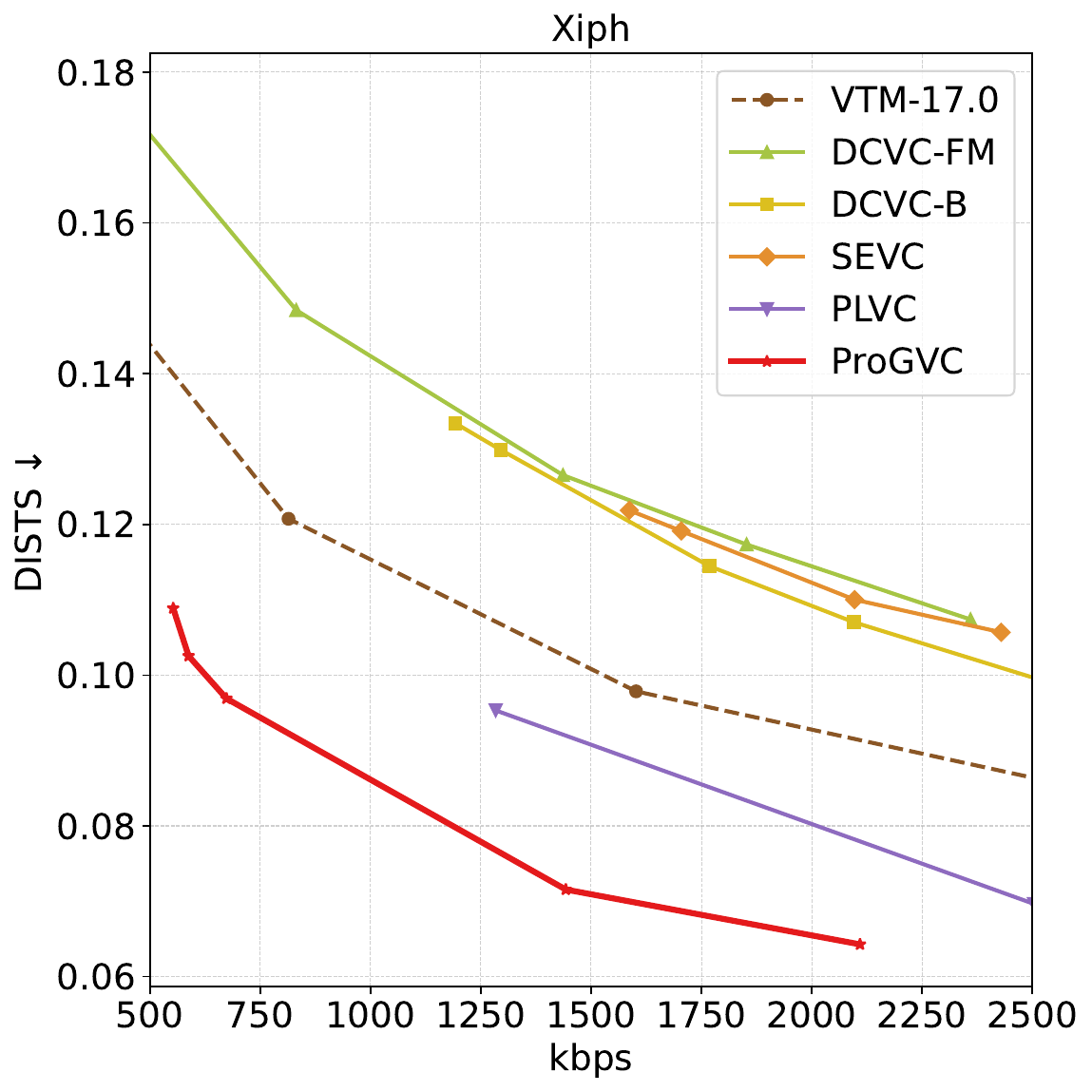}
    \label{fig:Xiph_dists}
  \end{subfigure}
    \hfill
    \hspace{-10pt}
  \begin{subfigure}{0.25\linewidth}
    \includegraphics[width=\textwidth]{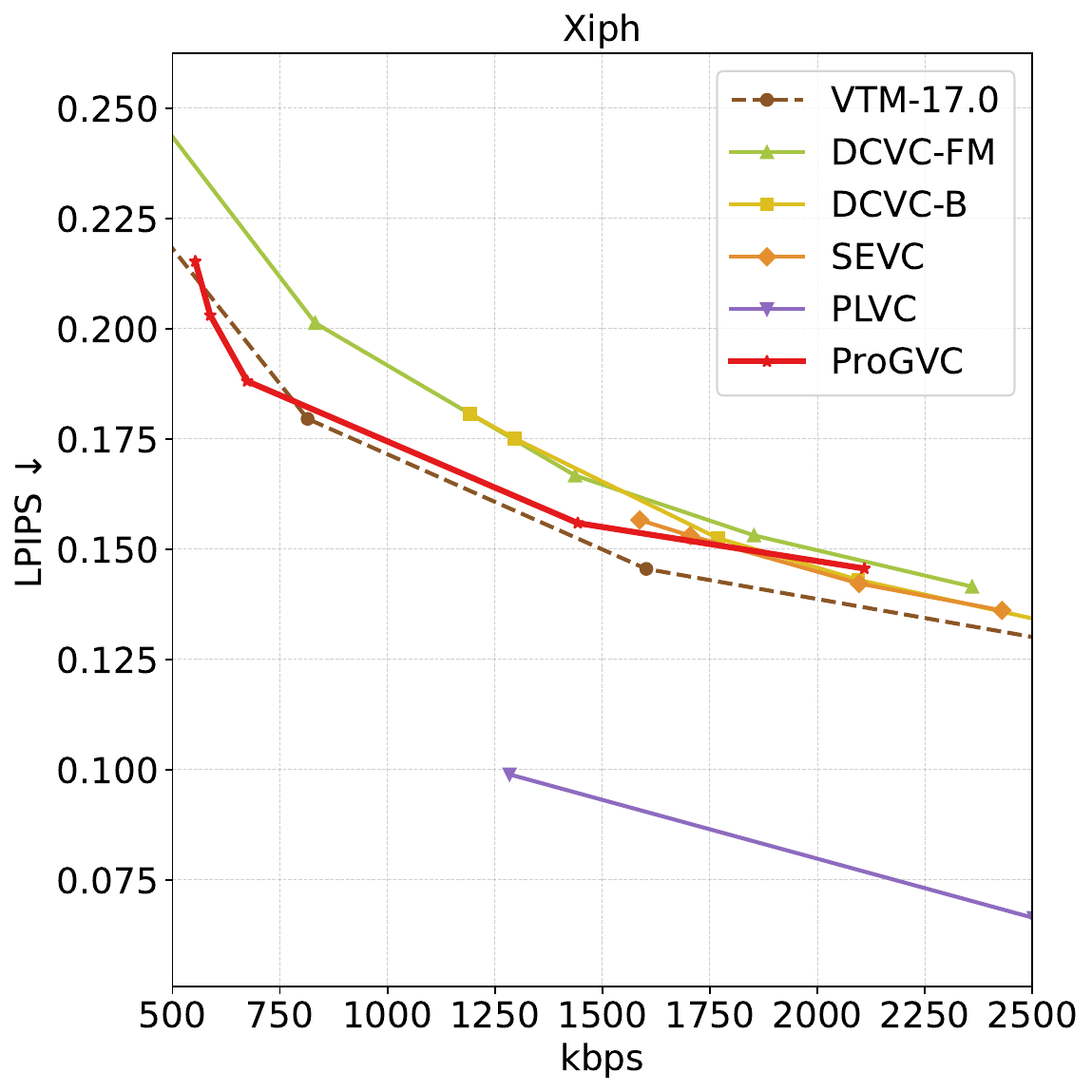}
    \label{fig:Xiph_lpips}
  \end{subfigure}
      \hfill
      \hspace{-10pt}
  \begin{subfigure}{0.25\linewidth}
    \includegraphics[width=\textwidth]{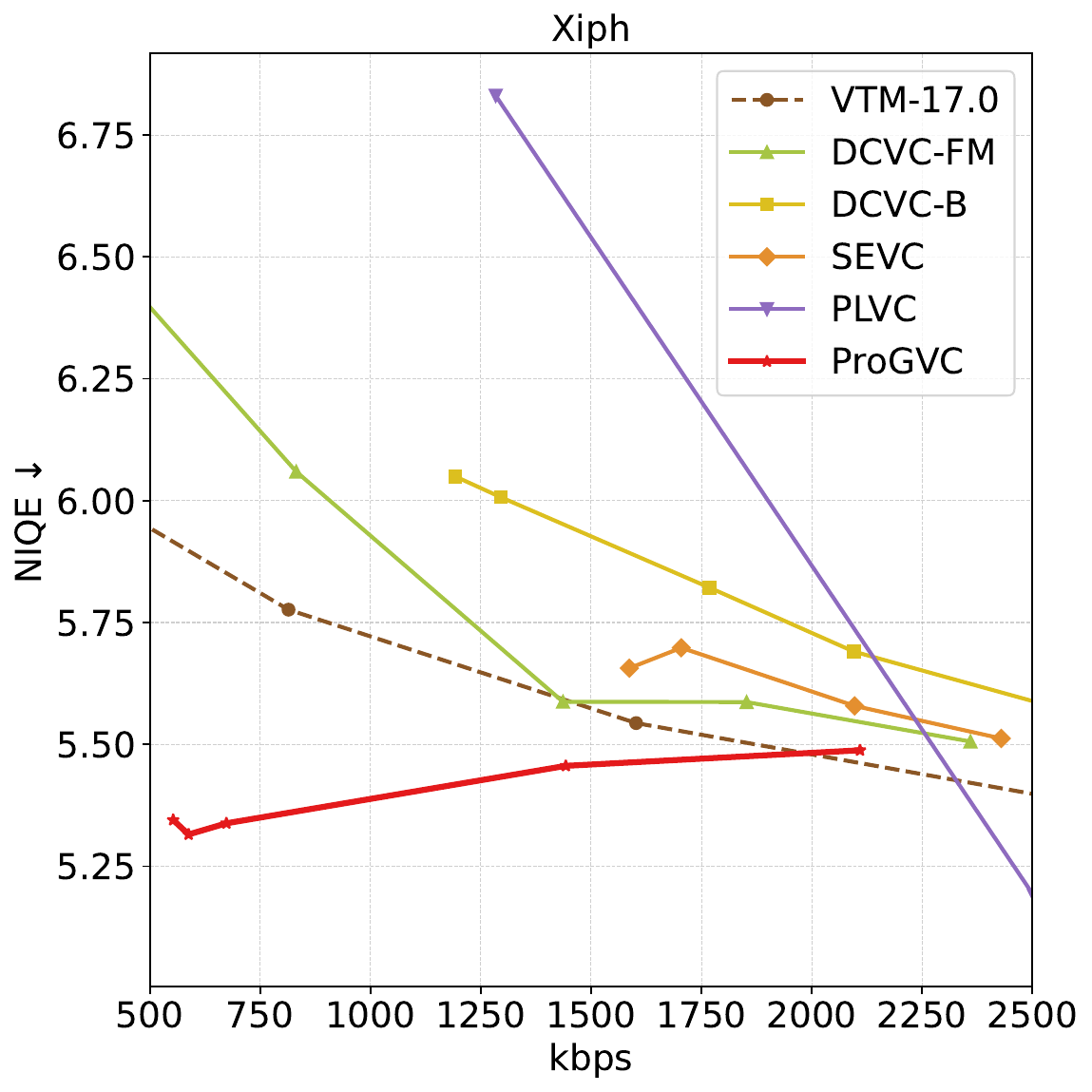}
    \label{fig:Xiph_niqe}
  \end{subfigure}
      \hfill
      \hspace{-10pt}
  \begin{subfigure}{0.25\linewidth}
    \includegraphics[width=\textwidth]{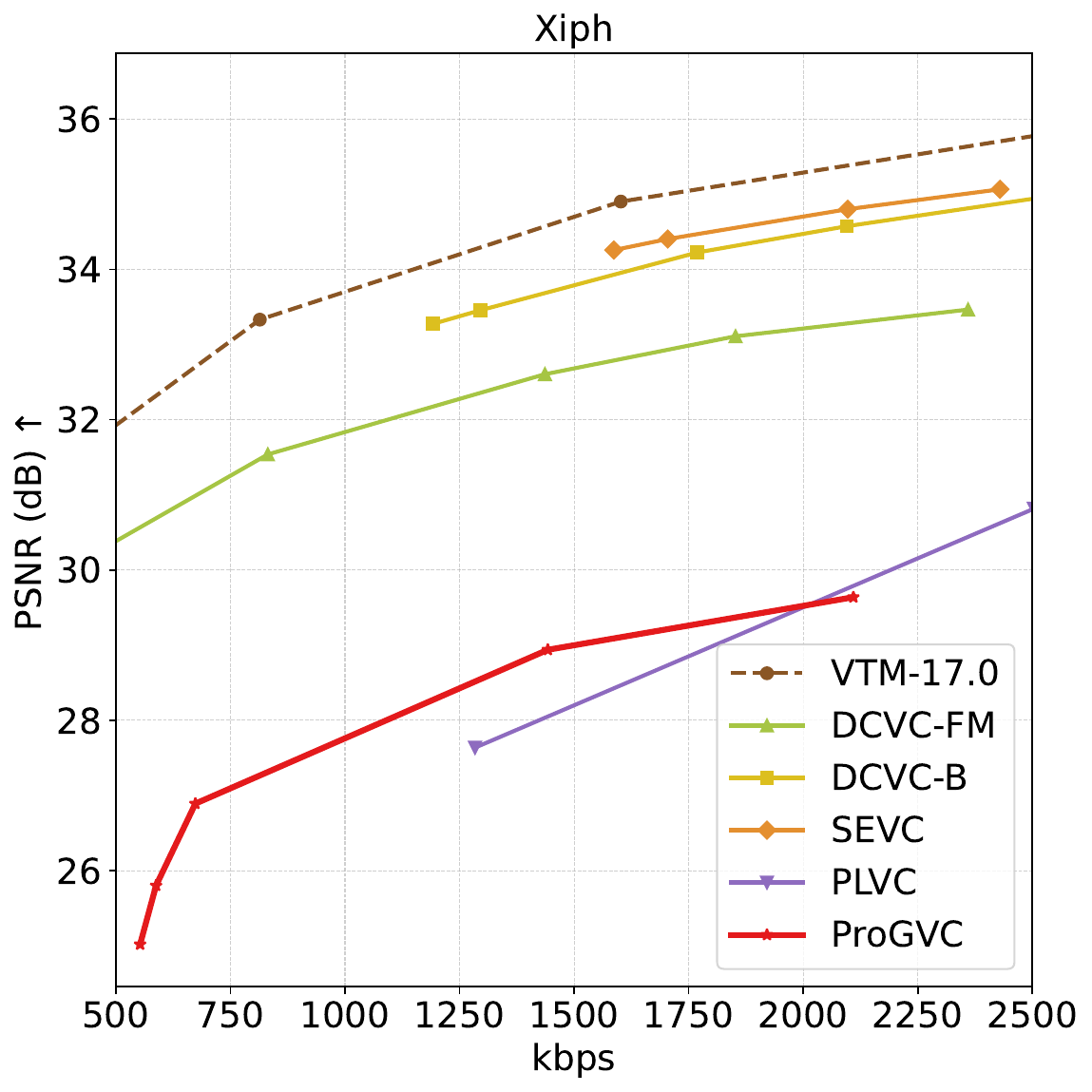}
    \label{fig:Xiph_psnr}
  \end{subfigure}

  \begin{subfigure}{0.252\linewidth}
    \includegraphics[width=\textwidth]{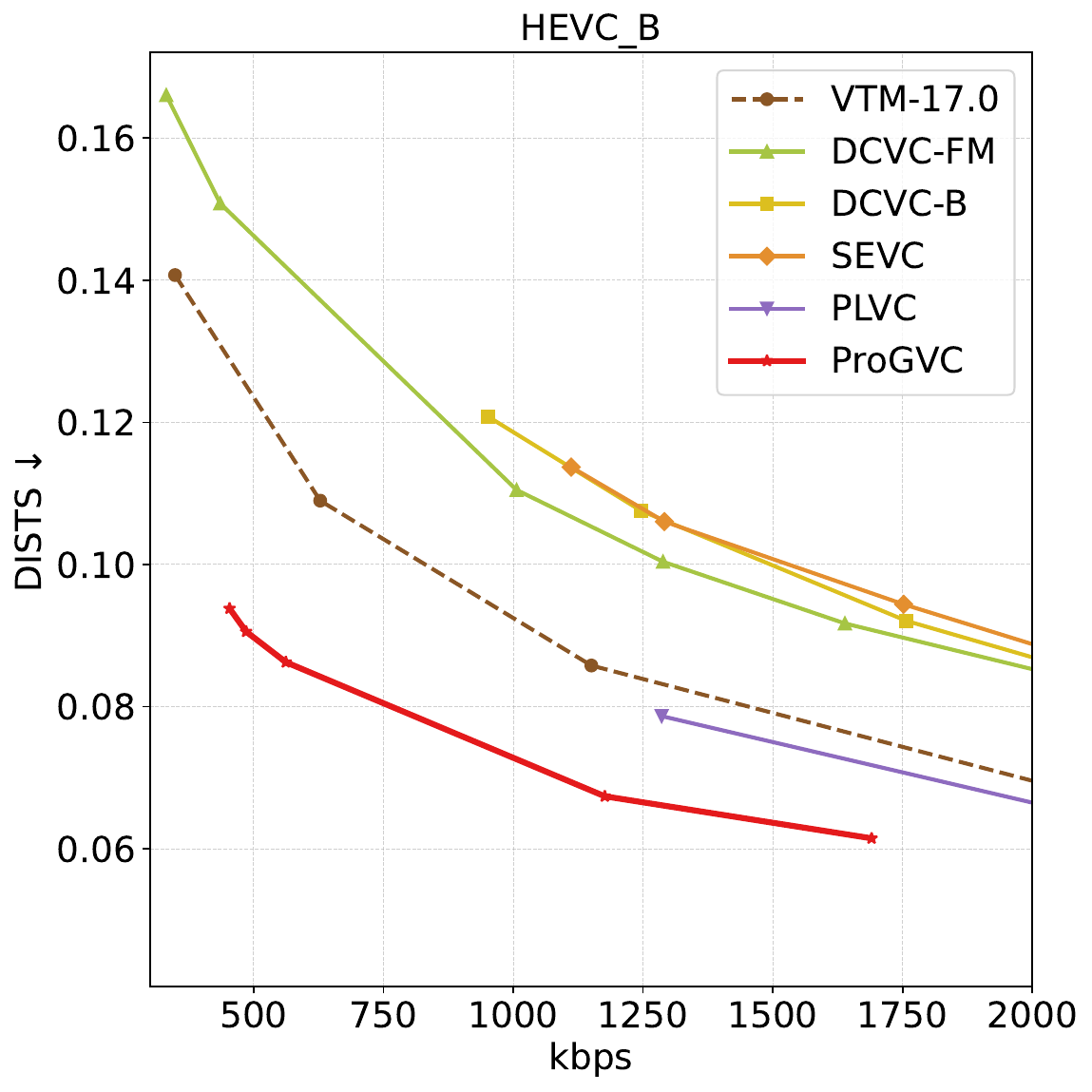}
    \label{fig:HEVCB_dists}    
  \end{subfigure}
    \hfill
    \hspace{-10pt}
  \begin{subfigure}{0.252\linewidth}
    \includegraphics[width=\textwidth]{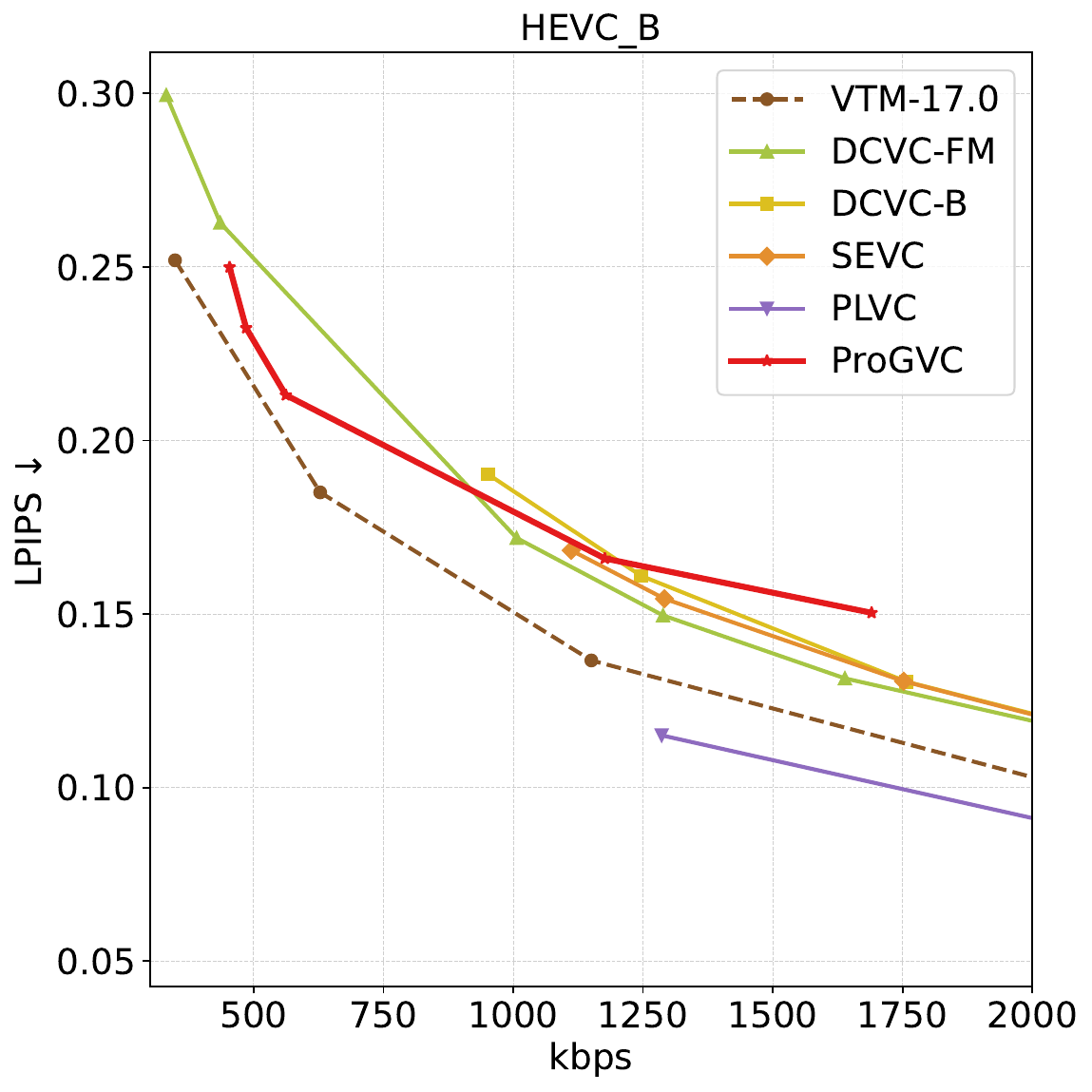}
    \label{fig:HEVCB_lpips}
  \end{subfigure}
      \hfill
      \hspace{-10pt}
  \begin{subfigure}{0.252\linewidth}
    \includegraphics[width=\textwidth]{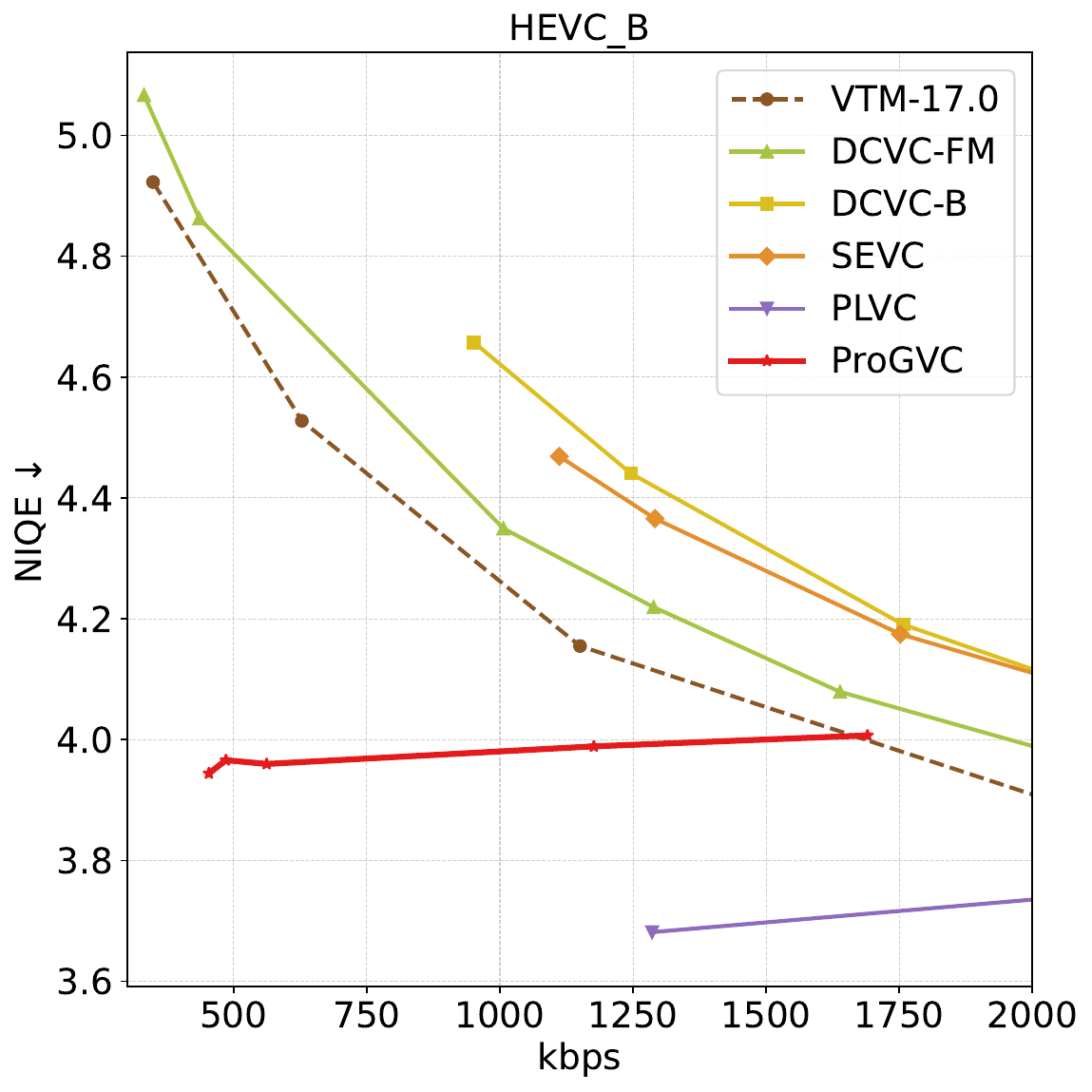}
    \label{fig:HEVCB_niqe}
  \end{subfigure}
      \hfill
      \hspace{-10pt}
  \begin{subfigure}{0.252\linewidth}
    \includegraphics[width=\textwidth]{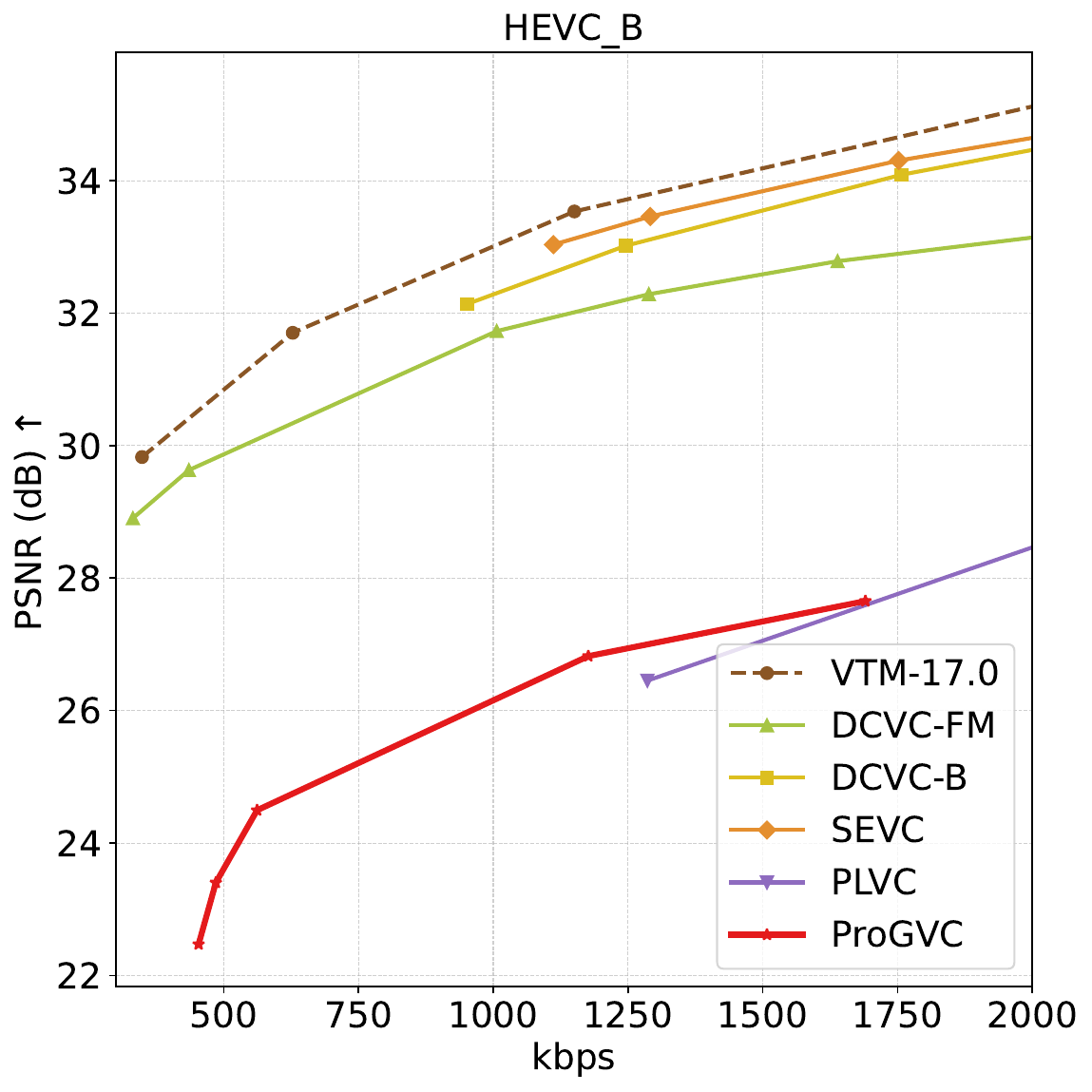}
    \label{fig:HEVCB_psnr}
  \end{subfigure}

  \begin{subfigure}{0.25\linewidth}
    \includegraphics[width=\textwidth]{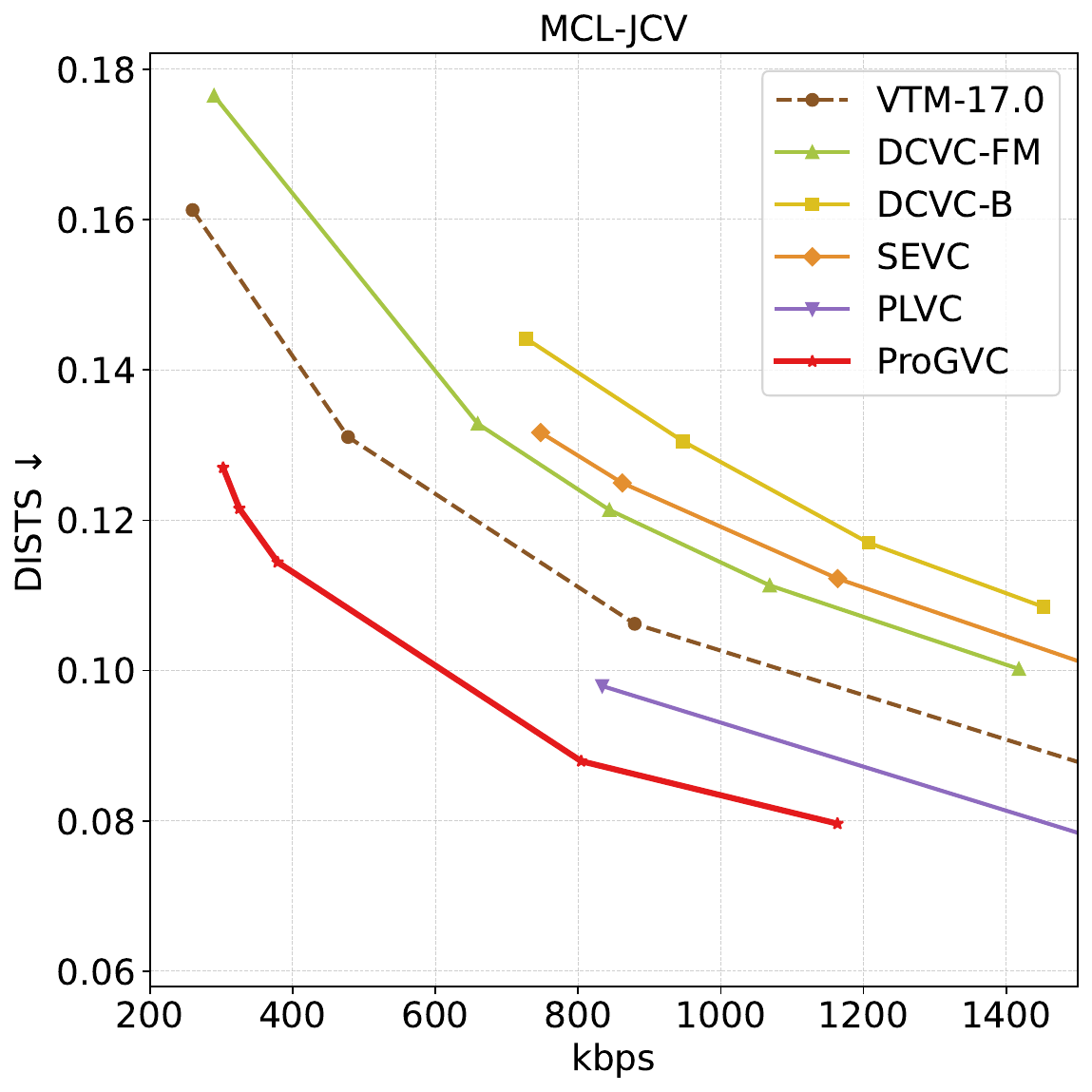}
    \label{fig:MCLJCV720p_dists}    
  \end{subfigure}
    \hfill
    \hspace{-10pt}
  \begin{subfigure}{0.25\linewidth}
    \includegraphics[width=\textwidth]{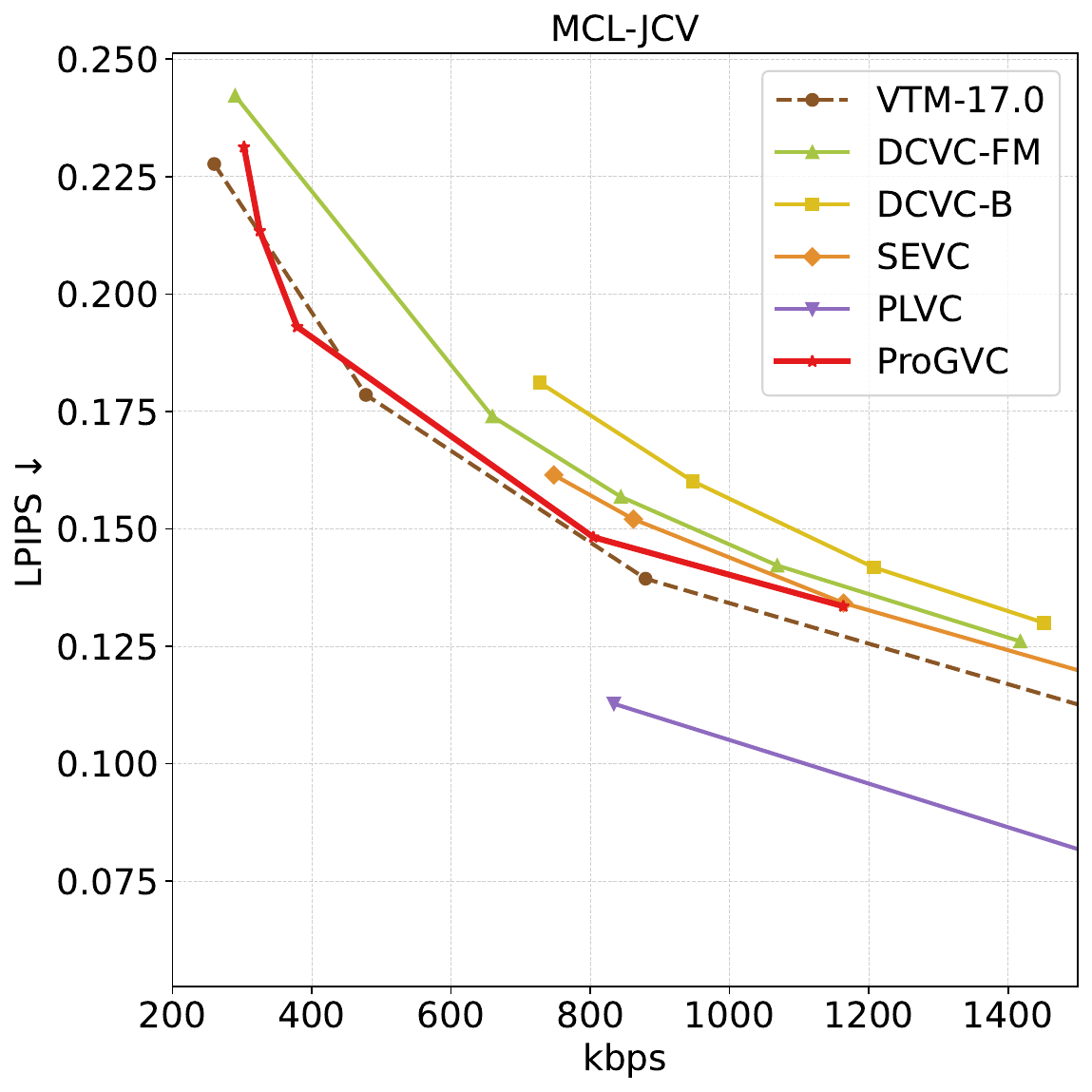}
    \label{fig:MCLJCV720p_lpips}
  \end{subfigure}
      \hfill
      \hspace{-10pt}
  \begin{subfigure}{0.25\linewidth}
    \includegraphics[width=\textwidth]{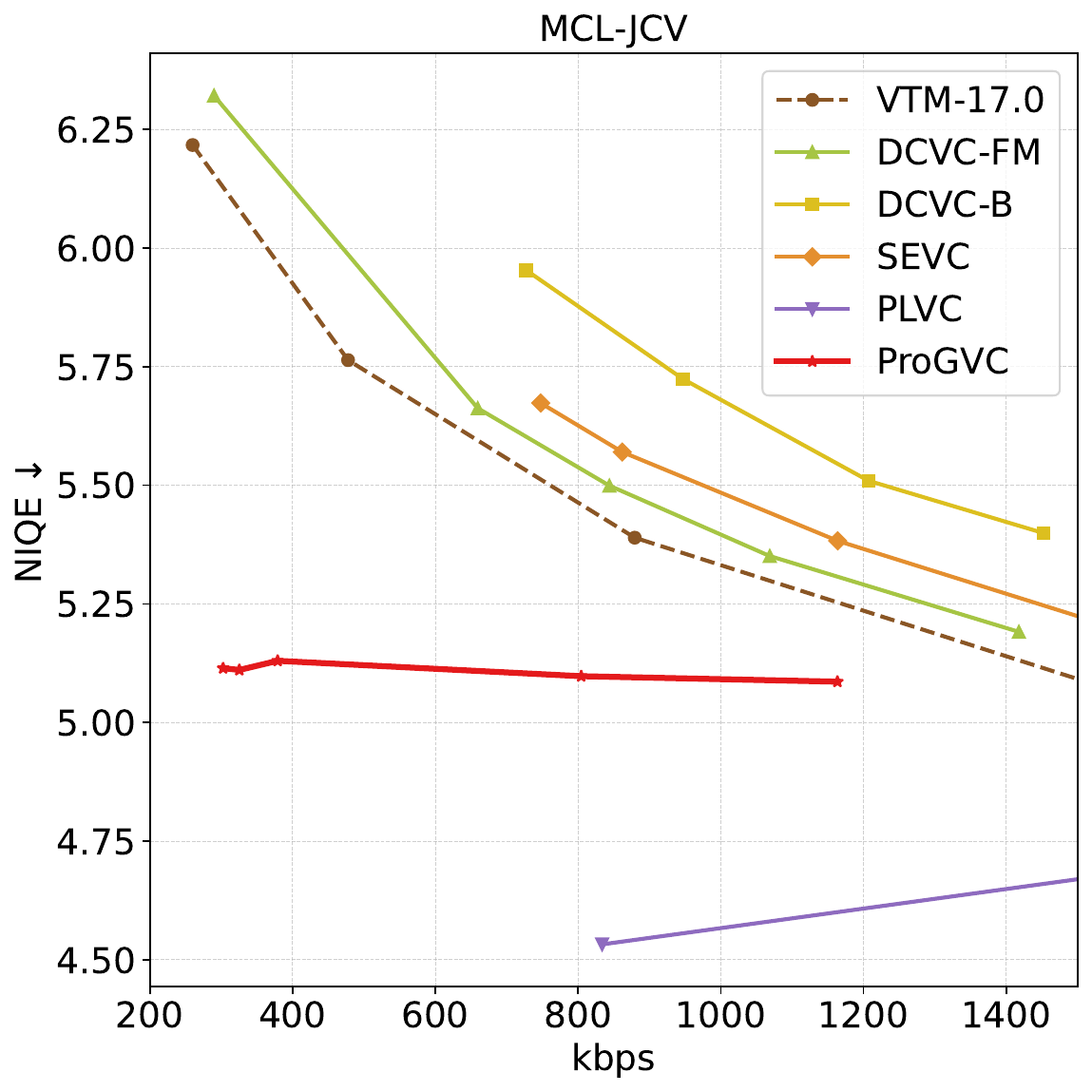}
    \label{fig:MCLJCV720p_niqe}
  \end{subfigure}
      \hfill
      \hspace{-10pt}
  \begin{subfigure}{0.25\linewidth}
    \includegraphics[width=\textwidth]{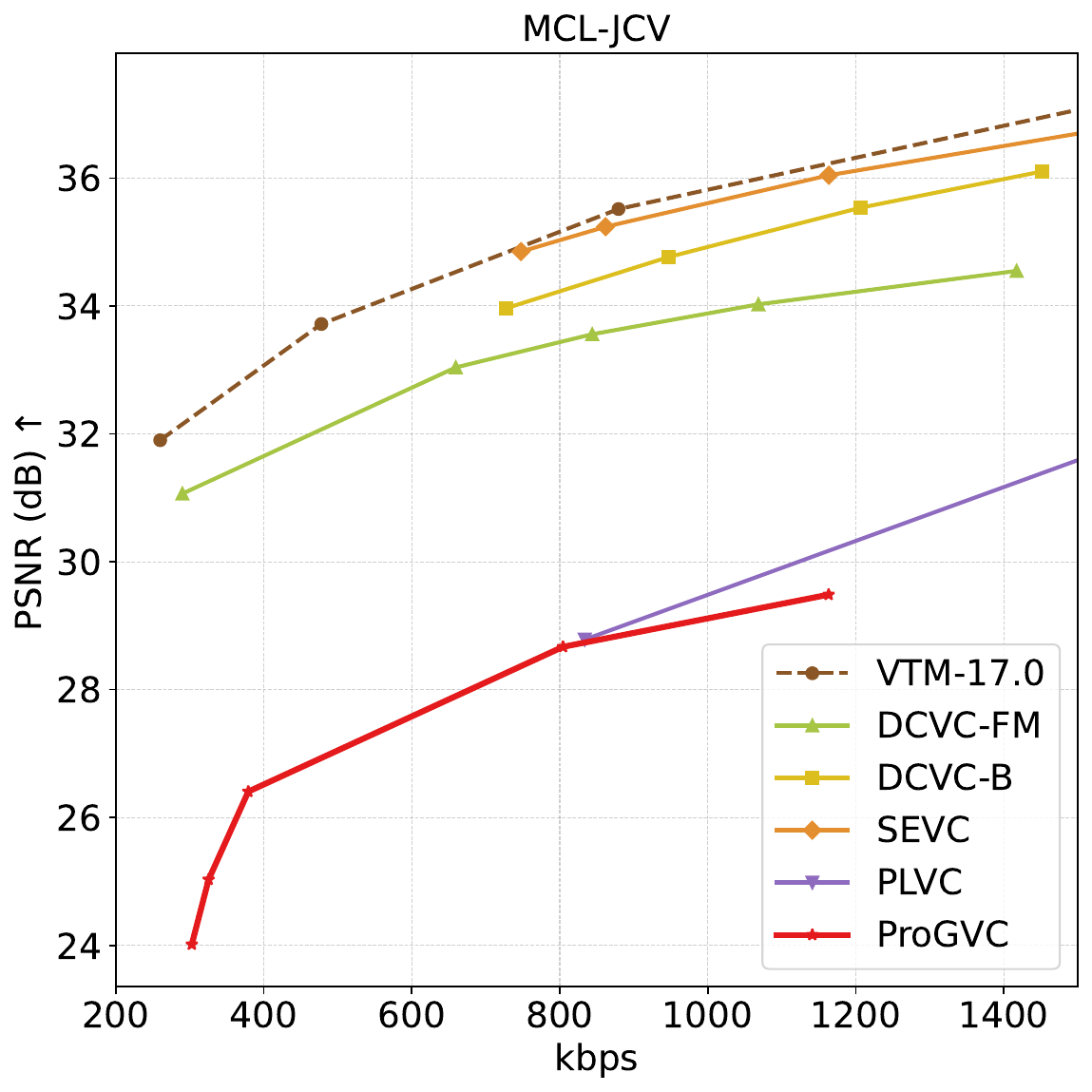}
    \label{fig:MCLJCV720p_psnr}
  \end{subfigure}
  \caption{\textbf{Rate and perception/fidelity curves on Xiph, HEVC B and MCL-JCV datasets.}
  }
  \label{fig:RDcurves}
\end{figure*}

\subsection{Comparison Results}
\noindent\textbf{Quantitative Results.}
As demonstrated in \cref{fig:RDcurves}, ProGVC achieves the best performance on DISTS among all competing methods. It also outperforms traditional and fidelity-oriented codecs under NIQE, while delivering a comparable perceptual efficiency on LPIPS. Compared with the perceptual codec PLVC~\cite{yang2022perceptual}, it further improves NIQE-based compression efficiency on Xiph. \cref{tab:bdrate} reports the BD-rate and BD-metric results using VTM-17.0~\cite{bross2021overview} as the anchor, confirming that ProGVC achieves superior or comparable perceptual quality to other strong baselines while supporting scalability and variable-rate control. \cref{tab:temporalconsistency} further reports t-LPIPS which explicitly evaluating temporal consistency, and ProGVC consistently outperforms all baselines. This is mainly attributable to our hierarchical next-scale autoregressive generation and temporal-spatial attention design, where each token attends to tokens from other frames at current and previous scales, effectively capturing cross-frame temporal dependencies.

\begin{figure*}[t]
    \centering
    {\scriptsize
        \begin{tabular}{ccccc}
            Ground Truth & VTM-17.0~\cite{bross2021overview} & SEVC~\cite{bian2025augmented} & PLVC~\cite{yang2022perceptual} & \textbf{ProGVC} \\
        \end{tabular}
    }

    \includegraphics[width=\textwidth]{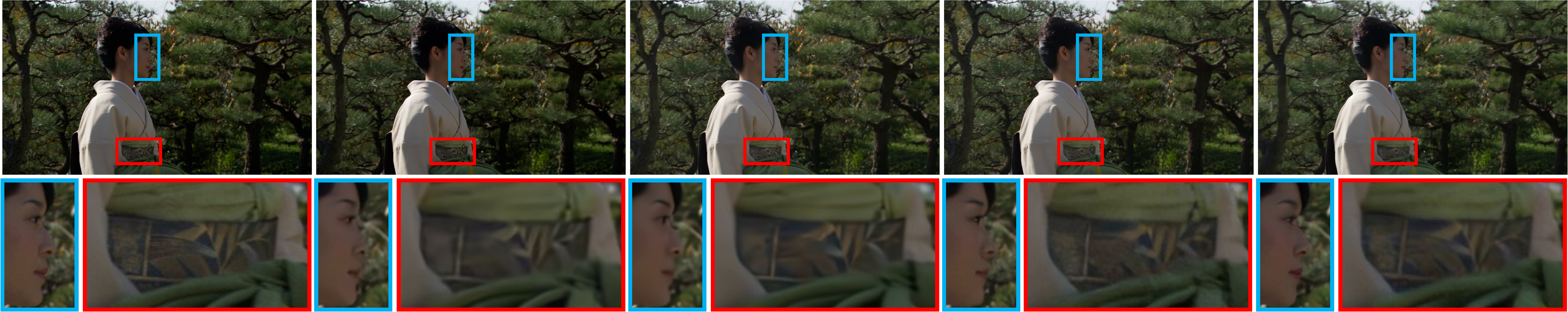}

    {\scriptsize
        \begin{tabular}{ccccc}
            Bitrate $\downarrow$ / DISTS $\downarrow$
            & 444.15 kbps/0.1174
            & 751.66 kbps/0.1330
            & 745.69 kbps/0.0717
            & 355.6 kbps/0.0621 \\
        \end{tabular}
    }

    \includegraphics[width=\textwidth]{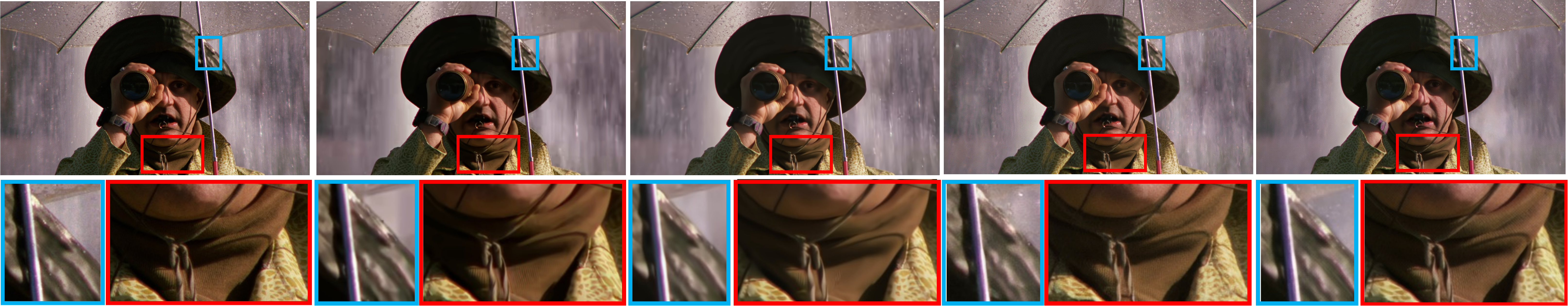}

    {\scriptsize
        \begin{tabular}{ccccc}
            Bitrate $\downarrow$ / DISTS $\downarrow$
            & 302.2 kbps/0.1283
            & 462.72 kbps/0.1259
            & 766.5 kbps/0.0771
            & 300.2 kbps/0.0590 \\
        \end{tabular}
    }

    \caption{\textbf{Visual comparisons with baselines on the \textit{Kimono} sequence in the Xiph dataset and the \textit{videoSRC07} sequence in the MCL-JCV dataset.}}
    \label{fig:visualizationCombined}
\end{figure*}

We note that ProGVC performs more favorably under DISTS and NIQE than under LPIPS. This is likely because our autoregressive context model is trained to capture the natural video distribution, favoring structural coherence and natural textures over strict pixel- or feature-level correspondence. As a result, slight local deviations, as shown in \cref{fig:visualizationCombined}, may be penalized more heavily by LPIPS. Taken together, the multi-metric and qualitative results suggest that ProGVC achieves competitive or superior perceptual quality.

\begin{table*}[t]
\caption{\label{tab:bdrate}
\textbf{BD-rate$\downarrow$ (\%) / BD-metric$\uparrow$ on the Xiph, HEVC B, and MCL-JCV datasets.} \textcolor{red}{\textbf{Red}} and \textcolor{blue}{\textbf{Blue}} indicate the best and the second-best performance in terms of BD-metric, respectively. ``N/A'' indicates that BD-rate cannot be calculated due to the lack of quality overlap.
}
\centering
\scalebox{0.85}{%
\begin{tabular}{llcccc}
\toprule
\multicolumn{1}{l}{Dataset} & \multicolumn{1}{l}{Method} &
\multicolumn{1}{c}{DISTS} & \multicolumn{1}{c}{LPIPS} & \multicolumn{1}{c}{NIQE} & \multicolumn{1}{c}{PSNR(dB)} \\
\midrule

% ===================== Xiph (6 rows) =====================
\multirow{6}{*}{Xiph}
& VTM-17.0~\cite{bross2021overview}
  & 0.0 / 0.0000   & \textcolor{blue}{\textbf{0.0 / 0.0000}}   & \textcolor{blue}{\textbf{0.0 / 0.0000}}     & \textcolor{red}{\textbf{0.0 / 0.0000}} \\
& DCVC-FM \cite{li2024neural}
  & 92.0 / -0.0257  & 40.4 / -0.0228  & 71.3 / -0.2525  & 111.1 / -1.8623 \\
& SEVC~\cite{bian2025augmented}
  & 87.7 / -0.0179  & 19.4 / -0.0062  & 41.8 / -0.1331  & 51.5 / -0.9035 \\
& DCVC-B~\cite{sheng2025bi}
  & 63.2 / -0.0148  & 17.3 / -0.0065  & 94.7 / -0.2306  & 56.8 / -1.0048 \\
& PLVC~\cite{yang2022perceptual}
  & \textcolor{blue}{\textbf{-40.3 / 0.0109}}  & \textcolor{red}{\textbf{-83.7 / 0.0565}}  & 169.9 / -0.0902 & 825.9 / -5.5261 \\
& \textbf{ProGVC}
  & \textcolor{red}{\textbf{-61.5 / 0.0300}}
  & 3.5 / -0.0025
  & \textcolor{red}{\textbf{-62.2 / 0.2670}}
  & 891.2 / -5.8437 \\
\midrule

% ===================== HEVC-B (6 rows) =====================
\multirow{6}{*}{HEVC-B}
& VTM-17.0~\cite{bross2021overview}
  & 0.0 / 0.0000   & \textcolor{blue}{\textbf{0.0 / 0.0000}}   & 0.0 / 0.0000     & \textcolor{red}{\textbf{0.0 / 0.0000}} \\
& DCVC-FM \cite{li2024neural}
  & 62.8 / -0.0231  & 41.0 / -0.0356  & 23.1 / -0.1653  & 50.6 / -1.2807 \\
& SEVC~\cite{bian2025augmented}
  & 88.1 / -0.0193  & 39.2 / -0.0189  & 60.1 / -0.2066  & 20.5 / -0.5422 \\
& DCVC-B~\cite{sheng2025bi}
  & 78.6 / -0.0184  & 41.5 / -0.0209  & 63.9 / -0.2300  & 28.7 / -0.7316 \\
& PLVC~\cite{yang2022perceptual}
  & \textcolor{blue}{\textbf{-18.1 / 0.0046}}  & \textcolor{red}{\textbf{-30.5 / 0.0150}}  & \textcolor{blue}{\textbf{-56.0 / 0.1048}}  & 705.5 / -6.1212 \\
& \textbf{ProGVC}
  & \textcolor{red}{\textbf{-47.7 / 0.0217}}
  & 30.3 / -0.0259
  & \textcolor{red}{\textbf{-52.5 / 0.3509}}
  & 879.6 / -6.8492 \\
\midrule

% ===================== MCL-JCV (6 rows) =====================
\multirow{6}{*}{MCL-JCV}
& VTM-17.0~\cite{bross2021overview}
  & 0.0 / 0.0000   & \textcolor{blue}{\textbf{0.0 / 0.0000}}   & 0.0 / 0.0000     & \textcolor{red}{\textbf{0.0 / 0.0000}} \\
& DCVC-FM~\cite{li2024neural}
  & 49.2 / -0.0215  & 32.9 / -0.0241  & 28.6 / -0.2241  & 68.8 / -1.4598 \\
& SEVC~\cite{bian2025augmented}
  & 52.0 / -0.0164  & 20.1 / -0.0102  & 30.6 / -0.1601  & 11.8 / -0.2967 \\
& DCVC-B~\cite{sheng2025bi}
  & 77.5 / -0.0247  & 44.2 / -0.0224  & 68.4 / -0.3514  & 37.2 / -0.9451 \\
& PLVC~\cite{yang2022perceptual}
  & \textcolor{blue}{\textbf{-28.8 / 0.0108}}  & \textcolor{red}{\textbf{-45.1 / 0.0322}}  & \textcolor{red}{\textbf{N/A / 0.5959}}    & 571.2 / -5.7383 \\
& \textbf{ProGVC}
  & \textcolor{red}{\textbf{-46.2 / 0.0238}}
  & 3.3 / -0.0016
  & \textcolor{blue}{\textbf{-65.5 / 0.5287}}
  & 893.8 / -6.6569 \\
\bottomrule
\end{tabular}
} % end scalebox

\end{table*}

\begin{table}[tb]
\centering
\begin{minipage}[t]{0.55\linewidth}
\centering
\captionof{table}{\textbf{Temporal consistency comparison on Xiph.} BD-rate$\downarrow$ (\%) / BD-metric$\uparrow$ are reported with VTM-17.0 as the anchor.}
\label{tab:temporalconsistency}
\scriptsize
\begin{tabular}{lc}
\toprule
Method & t-LPIPS \\
\midrule
DCVC-FM~\cite{li2024neural} & 46.14 / -0.0019 \\
SEVC~\cite{bian2025augmented} & -15.89 / 0.0004 \\
DCVC-B~\cite{sheng2025bi} & 58.61 / -0.0011 \\
PLVC~\cite{yang2022perceptual} & 709.87 / -0.0070 \\
\textbf{ProGVC} & \textbf{ -20.76 / 0.0007} \\
\bottomrule
\end{tabular}
\end{minipage}
\hfill
\begin{minipage}[t]{0.4\linewidth}
\centering
\captionof{table}{\textbf{Complexity comparison with other neural codecs.}}
\label{tab:complexity}
\scriptsize
\resizebox{\linewidth}{!}{
\begin{tabular}{lcc}
\toprule
Methods & Enc. T (s) $\downarrow$ & Dec. T (s) $\downarrow$ \\
\midrule
DCVC-B\cite{sheng2025bi}     & 1.28 & 1.09 \\
SEVC\cite{bian2025augmented} & 1.04 & \textbf{0.88} \\
DiffVC\cite{ma2025diffusion} & 0.82 & 4.54 \\
\textbf{ProGVC}              & \textbf{0.56} & 1.53 \\
\bottomrule
\end{tabular}
}
\end{minipage}
\end{table}

\noindent\textbf{Qualitative Results.}
\cref{fig:visualizationCombined} illustrates the visual results of different video compression methods. ProGVC produces more visually pleasant reconstructions at low bitrates. Compared to fidelity-oriented baselines, including VTM-17.0~\cite{bross2021overview} and SEVC~\cite{bian2025augmented}, ProGVC produces finer textures and higher structural integrity, mitigating blurring artifacts and fragmented edges. Compared to the perceptual baseline PLVC~\cite{yang2022perceptual}, ProGVC produces more natural textures.

\noindent\textbf{Complexity.}
We compare the encoding and decoding complexity of ProGVC with several neural video codecs, including the fidelity-oriented codecs DCVC-B~\cite{sheng2025bi} and SEVC~\cite{bian2025augmented}, as well as the diffusion-based perceptual codec DiffVC~\cite{ma2025diffusion}. PLVC~\cite{yang2022perceptual} is excluded due to its incompatible implementation with the unified evaluation platform. Since DiffVC is not open-sourced, we implement its architecture following the original paper without training. All experiments are conducted on the same hardware platform, with an NVIDIA H20 GPU and an Intel Xeon Platinum 8469C CPU (192 cores). For ProGVC, we report the average runtime over the lowest and highest bitrate settings.
As summarized in \cref{tab:complexity}, ProGVC achieves the lowest encoding complexity among all baselines, while its decoding time is comparable to DCVC-B~\cite{sheng2025bi} and SEVC~\cite{bian2025augmented}, and clearly lower than that of DiffVC.

\subsection{Ablation Studies}\label{sec:ablation}
We conduct a series of ablation studies on the key components of our framework. For a fair comparison, we report BD-rate and BD-metric results on the Xiph dataset, evaluated across three metrics: DISTS, LPIPS and PSNR.

\noindent \textbf{Multi-scale Auto-Regressive Context Modeling}. 
We evaluate the proposed context modeling procedure for both entropy coding and discarded token generation. To quantify its contribution to lossless entropy coding, we replace the proposed context model with a uniform probability distribution. As reported in \cref{tab:abla_contextmodeling}, the proposed context modeling reduces the bitrate by more than 50\% compared to the uniform baseline. Moreover, the proposed context modeling further improves both perceptual and fidelity metrics via token generation.

\begin{table*}[t]
\caption{\label{tab:abla_contextmodeling}
\textbf{Effect of multi-scale autoregressive context modeling}. ``uniform prob.'' denotes utilizing uniform probability distribution.
}
\centering
\resizebox{0.85\textwidth}{!}{
\begin{tabular}{l|ccc}
\hline
\multirow{2}{*}{Model Variants}
& \multicolumn{3}{c}{BD-rate$\downarrow$(\%) / BD-metric $\uparrow$} \\
\cline{2-4}
& DISTS & LPIPS & PSNR \\
\hline
Base & \textbf{0.0 / 0.0000} & \textbf{0.0 / 0.0000} & \textbf{0.0 / 0.0000} \\
\hline
w/o context model (uniform prob.)   & 122.4 / -0.0259 & 119.6 / -0.0360 & 119.4 / -2.3437 \\
w/o token generation  & N/A / -0.1183 & 280.9 / -0.1273 & 189.9 / -3.8102 \\
% \hline
\bottomrule
\end{tabular}
}
\end{table*}

\noindent \textbf{Attention mask design}. 
We conduct ablation studies to evaluate the effectiveness of the proposed sparse attention mask. In particular, we investigate the trade-off between the bitrate reduction by exploiting sufficient contextual information from previous scales and maintaining computational efficiency.

We begin by analyzing temporally aligned intra–intra and inter–inter attention masks, where tokens attend to context from the same frame(s) at earlier scales. Specifically, we compare three variants: (1) \textit{InfinityStar-like self-only attention}, where each scale attends only to itself; (2) \textit{VAR-like full causal attention}, where each scale attends to all prefix scales; and (3) \textit{our proposed design}, in which each scale attends only to itself together with its immediately preceding scale. As reported in \cref{tab:abla_attn1}, the proposed design achieves a favorable BD-rate–complexity trade-off. Compared with self-only attention, it consistently improves BD-rate across all three metrics. Moreover, it matches the compression performance of full causal attention while substantially reducing computational overhead.

\begin{table*}[t]
\caption{\label{tab:abla_attn1}
\textbf{Effect of three different temporal-aligned attention masks}. The per-frame encoding and decoding times (Enc. T and Dec. T) are averaged over the lowest and highest bitrates.
}
\centering
\resizebox{0.88\textwidth}{!}{
\begin{tabular}{l|ccc|c|c}
\hline
\multirow{2}{*}{Attn. Variants} 
& \multicolumn{3}{c|}{BD-rate$\downarrow$(\%) / BD-metric$\uparrow$}  & \multirow{2}{*}{Enc. T (s) $\downarrow$} & \multirow{2}{*}{Dec. T (s) $\downarrow$}\\
\cline{2-4}
& DISTS & LPIPS & PSNR & & \\
\hline
Ours & 0.0 / 0.0000 & 0.0 / 0.0000 & 0.0 / 0.0000 & 0.56 & 1.53 \\
\hline
Self-only &  6.8 / -0.0021 & 12.8 / -0.0062  & 10.9 / -0.3482 & 0.50 & 1.23 \\
Full causal &  -6.5 / 0.0021 & -3.8 / 0.0019 & 0.5 / -0.0184 & 0.64 & 1.96 \\
\bottomrule
\end{tabular}
}
\end{table*}

To preserve temporal consistency, inter scales should incorporate information from intra scales. As shown in \cref{tab:abla_attn2}, we compare the proposed largest-intra-scale conditioning strategy with three alternatives: no intra scales conditioning, conditioning on the smallest intra scale, and conditioning on intra scales at the same spatial resolution. The results indicate that attending to the largest intra scale consistently improves the compression performance of inter frames.
The largest intra scale, operating at the finest spatial resolution, provides high-frequency details that better match inter scales than low-frequency scales, thereby offering a stronger predictive prior. 
Meanwhile, the computational cost of the proposed strategy remains comparable across all variants. Based on these observations, the largest intra scale is adopted as one of the reference context for inter-scale token modeling.

\begin{table*}[t]
\caption{\label{tab:abla_attn2}
\textbf{Effect of different intra scales to the context modeling of inter scales}. The per-frame encoding and decoding times (denoted as Enc. T and Dec. T) are averaged across the lowest and highest bitrate settings.
}
\centering
\resizebox{0.93\textwidth}{!}{
\begin{tabular}{l|ccc|c|c}
\hline
\multirow{2}{*}{Attn. Variants} 
& \multicolumn{3}{c|}{BD-rate$\downarrow$(\%) / BD-metric $\uparrow$} & \multirow{2}{*}{Enc. T (s) $\downarrow$} & \multirow{2}{*}{Dec. T (s) $\downarrow$}\\
\cline{2-4}
& DISTS & LPIPS & PSNR & & \\
\hline
Ours (largest scale) & \textbf{0.0 / 0.0000} & \textbf{0.0 / 0.0000} & \textbf{0.0 / 0.0000} & 0.56 & 1.53 \\
\hline
No intra reference &  37.3 / -0.0171 & 41.7 / -0.0254 & 46.6 / -1.6698 & 0.53 & 1.46 \\
Smallest scale & 42.7 / -0.0191 & 40.4 / -0.0252 & 46.8 / -1.6803 & 0.53 & 1.46 \\
Same resolution scale & 11.7 / -0.0042 & 12.2 / -0.0064 & 14.4 / -0.5131 & 0.54 & 1.49 \\
\bottomrule
\end{tabular}
}

\end{table*}

\section{Conclusion and Future Work}
In this paper, we introduce ProGVC (Progressive-based Generative Video Compression via Auto-Regressive Context Modeling), the first perceptual video coding paradigm based on the visual autoregressive model that unifies progressive scalability, efficient entropy coding, and detail synthesis within a single framework. By integrating a multi-scale residual quantizer with a task-specific multi-scale autoregressive context model, ProGVC achieves lossless compression of low-frequency global structures while generating high-frequency details. Experimental results demonstrate that our model exhibits consistent improvements in perceptual quality over fidelity-oriented codecs and competitive performance compared to perceptual baselines, while maintaining native scalability, adaptive rate control, and low encoding-decoding latency.

Despite these advantages, ProGVC still leaves room for further optimization. First, the supported video resolution and duration are currently constrained by the training configuration of the autoregressive model. In particular, extending AR-based video generation beyond 720p remains challenging due to the substantially increased computational cost of training at higher resolutions, which mainly arises from the current maturity of the underlying AR video generative backbone. Second, the current adaptive rate control strategy is still coarse-grained, as bitrate is adjusted by transmitting or discarding entire scales, resulting in discrete rather than continuous rate points.
Future work will scale ProGVC to higher resolutions as the underlying AR generative backbones continue to advance, and explore more general resolution- and length-agnostic architectures. In addition, we will investigate finer-grained rate control mechanisms, such as token-level dropping, to enable smoother bitrate adaptation in practical streaming scenarios.

\par\vfill\par
\bibliographystyle{splncs04}
\bibliography{main}

\clearpage
\appendix
\section*{Supplementary Material}

\input{supple_main}

\end{document}

%% file: supple_main.tex
\titlerunning{ProGVC Supplementary Material}

\section{Test Settings}
Most compared methods operate in the RGB color space, whereas the raw videos are stored in YUV420 format. To minimize the impact of color space conversion on the source content, each codec is run in its native color space. Codecs that support YUV format are tested in YUV color space, while the others operate in RGB color space. For evaluation, all decoded outputs are converted to RGB format so that quality metrics are computed consistently across methods.

\subsection{Test Sequences}
The raw videos are stored in YUV420 format and converted to RGB format using the BT.709 standard. For evaluation, we extract the first 81 frames of each sequence. For 1080p sequences, they are further downsampled to the resolution of 720p. For baselines that reconstruct in YUV color space, the decoded YUV420 videos are further converted to RGB color space by applying the BT.709 YCbCr-to-RGB transform, following JPEG-AI standard\cite{jpegai}.

\subsection{Traditional Codecs}
For VTM-17.0\footnote{https://vcgit.hhi.fraunhofer.de/jvet/VVCSoftware\_VTM}, We use the official random-access configuration in the YUV420 domain. Specifically, only the first frame is encoded as I-frame, with all subsequent frames coded as
inter frames to maximize compression efficiency. The detailed parameters to encode each test sequence are set as follows:

-c encoder\_randomaccess\_vtm.cfg  --InputFile=\textit{\{input file name\}} \\
--InputBitDepth=8 --OutputBitDepth=8 --OutputBitDepthC=8\\
--InputChromaFormat=420 --FrameRate=\textit{\{frame rate\}} \\
--FramesToBeEncoded=\textit{\{frame number\}} --SourceWidth=\textit{\{width\}} \\
--SourceHeight=\textit{\{height\}} --IntraPeriod=-1 --QP=\textit{\{qp\}} \\
--BitstreamFile=\textit{\{bitstream file name\}}

\subsection{Neural-based Codecs}
\noindent \textbf{DCVC-FM} \cite{li2024neural} We use the official code and checkpoints provided in the authors’ GitHub repository \footnote{https://github.com/microsoft/DCVC/tree/main/DCVC-family/DCVC-FM}. YUV420 format is used during compression. The GOP size is set to 81.

\noindent \textbf{DCVC-B\cite{sheng2025bi} / SEVC\cite{bian2025augmented}} We use the official implementation and checkpoints from the authors’ GitHub repository\footnote{https://github.com/xhsheng-ustc/DCVC-B}\footnote{https://github.com/EsakaK/SEVC}. The GOP size is 81.

\noindent \textbf{PLVC\cite{yang2022perceptual}}
PLVC is evaluated using the official implementation and pre-trained weights \footnote{https://github.com/RenYang-home/PLVC}. Since PLVC uses HiFiC\cite{mentzer2020high} for I-frame coding, the official HiFiC implementation \footnote{https://github.com/tensorflow/compression/tree/master/models/hific} is adopted.

\noindent \textbf{ProGVC} ProGVC is tested with the transmitted inter scales of 10, 11, 12, 13 and 14. These configuration settings help to keep the bitrate within a practical range.

\section{More Quantitative Results}
\subsection{Additional Testing Dataset} 
\begin{figure*}[tb]
  \centering
  \vspace{-0.2cm}
  \begin{subfigure}{0.25\linewidth}
    \includegraphics[width=\textwidth]{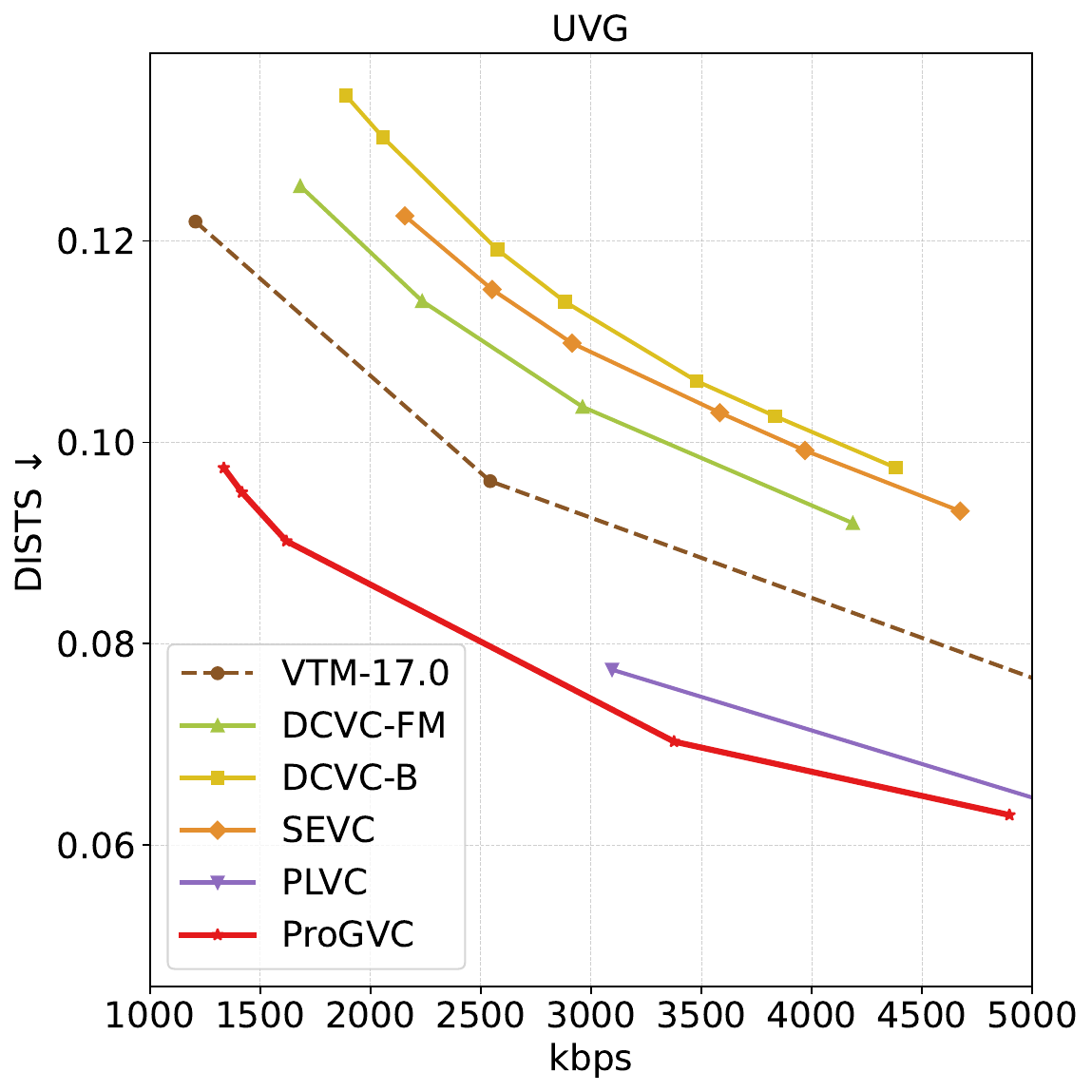}
  \end{subfigure}
    \hfill
    \hspace{-10pt}
  \begin{subfigure}{0.25\linewidth}
    \includegraphics[width=\textwidth]{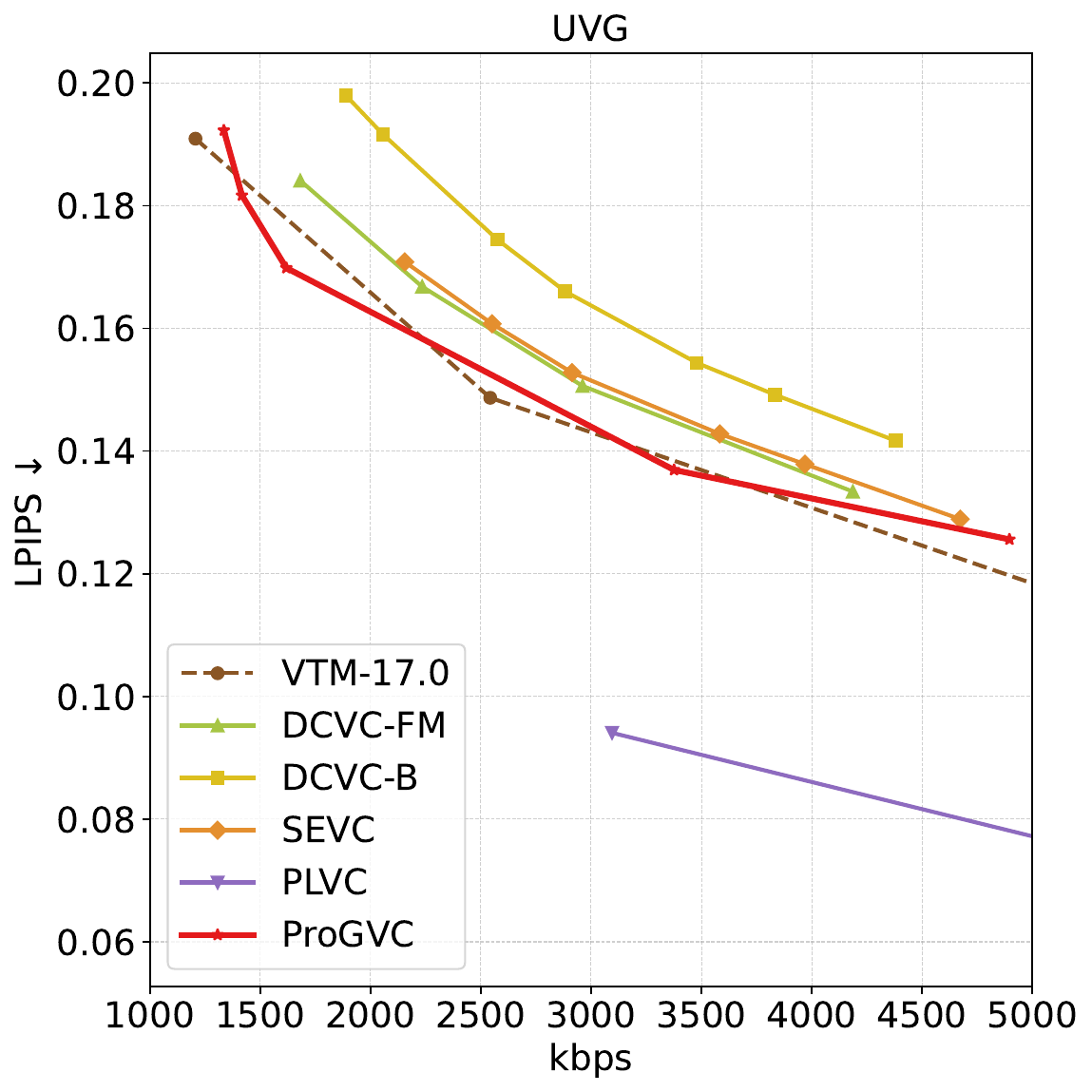}
  \end{subfigure}
      \hfill
      \hspace{-10pt}
  \begin{subfigure}{0.25\linewidth}
    \includegraphics[width=\textwidth]{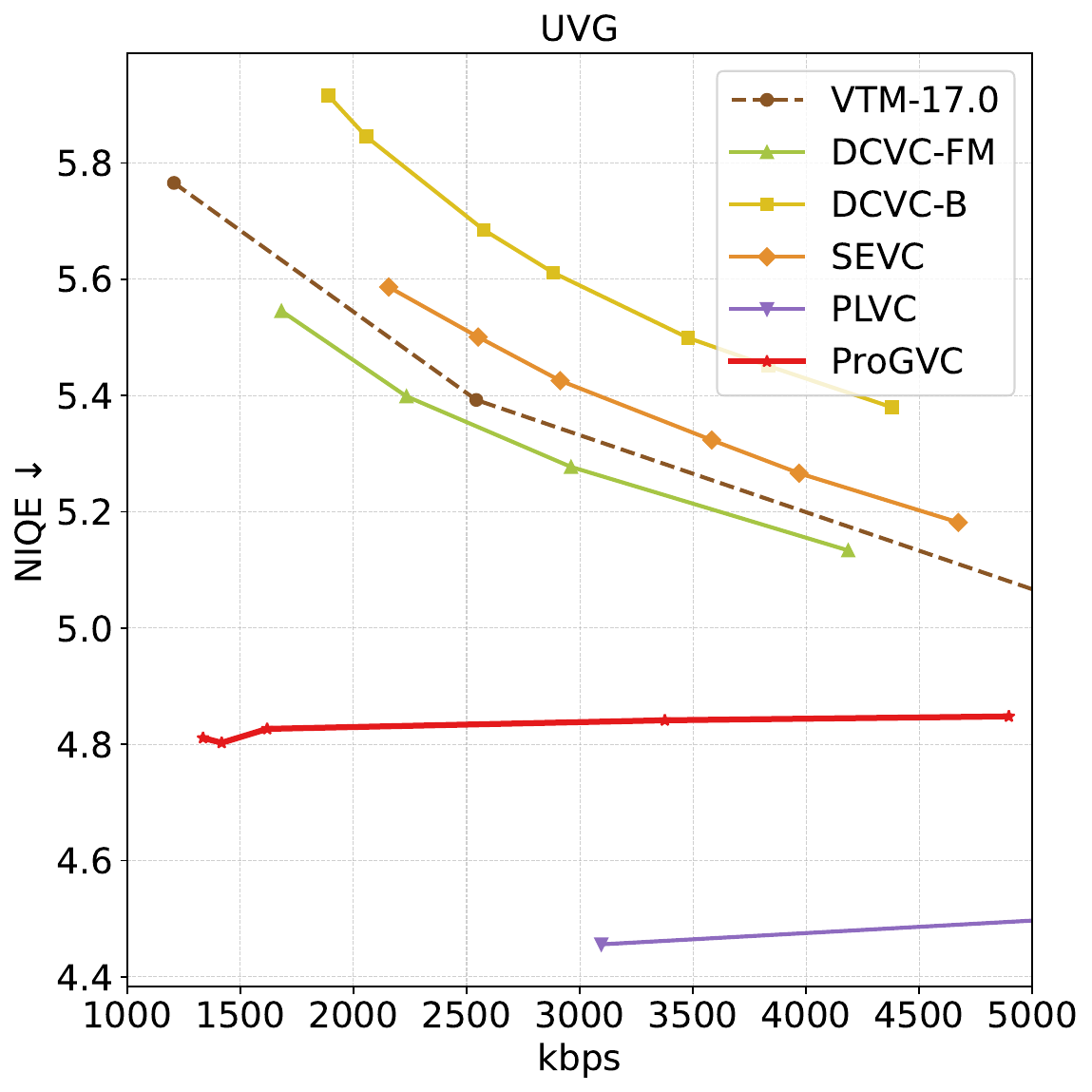}
  \end{subfigure}
      \hfill
      \hspace{-10pt}
  \begin{subfigure}{0.25\linewidth}
    \includegraphics[width=\textwidth]{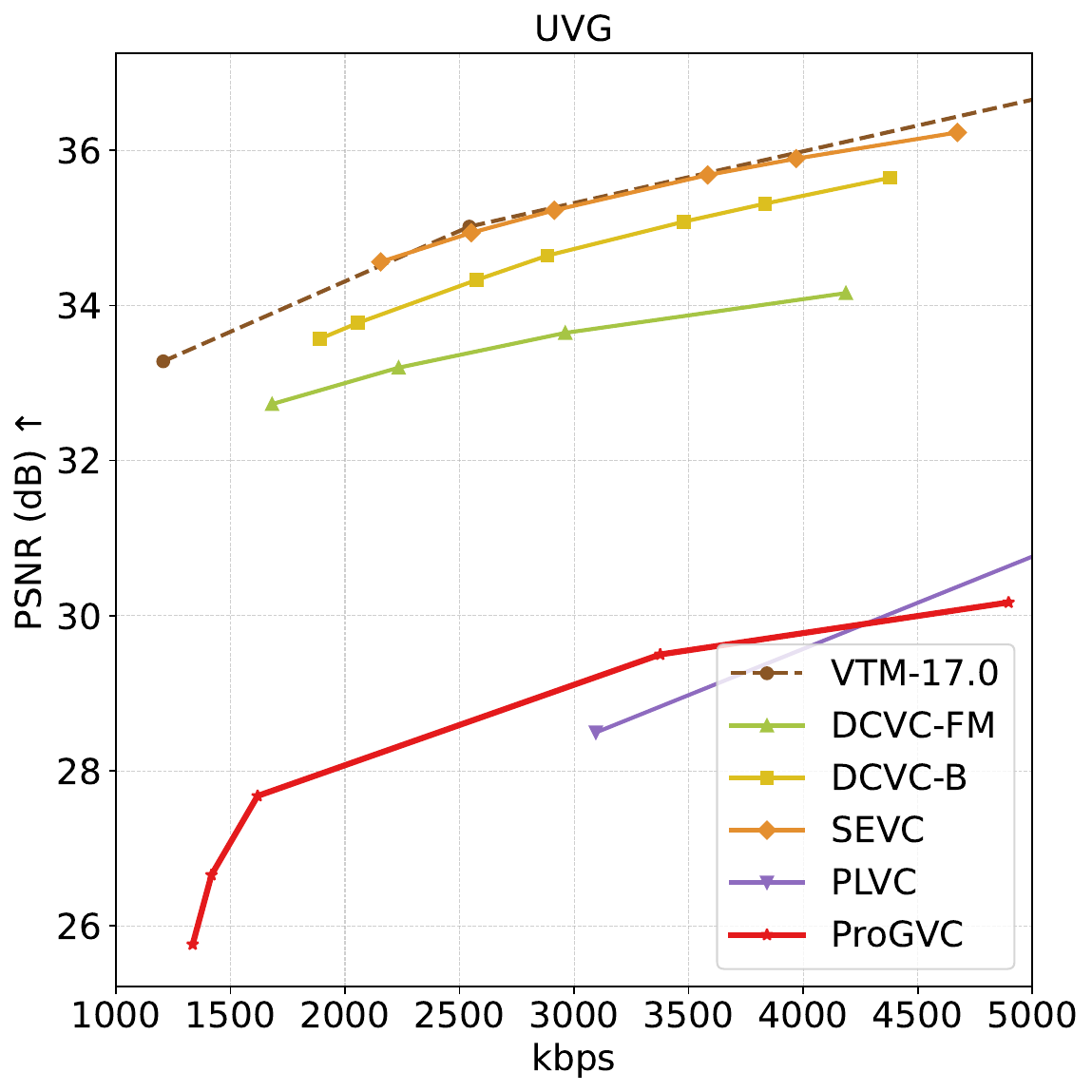}
  \end{subfigure}

    \vspace{-10pt}
  \caption{\textbf{Rate and perception/fidelity curves on UVG datasets.}
  }
  \label{fig:RDcurves_UVG}
  \vspace{-15pt}
\end{figure*}

Following prior work \cite{sheng2025bi,yang2022perceptual}, ProGVC is additionally evaluated on UVG dataset, which is downsampled from the resolution of 1080p to 720p. As shown in \cref{fig:RDcurves_UVG}, ProGVC consistently outperforms all competing methods in terms of DISTS. It also surpasses traditional and fidelity-oriented codecs under NIQE, while achieving comparable perceptual efficiency as measured by LPIPS. Compared to PLVC, ProGVC performs worse on LPIPS and NIQE, but achieves better PSNR performance over a wide bitrate range. Overall, ProGVC demonstrates promising perceptual performance while supporting scalability and variable-rate control.

\subsection{Comparison with Diffusion-based Perceptual Codecs}

We also compare the performance of ProGVC with the latest diffusion-based perceptual codecs, including GNVC-VD\cite{mao2025generative} and DiffVC\cite{ma2025diffusion}. However, ProGVC is evaluated under 720p/81-frame configuration, which differs from the 1080p/96-frame setup used by diffusion-based codecs, and the latter are not open-sourced yet. Therefore, we adopt VTM-17.0 as a common intermediate anchor and obtain the diffusion-based results from their published papers. By comparing each method’s relative compression performance over VTM-17.0, we assess the coding efficiency of ProGVC versus the diffusion-based codecs. Here, to align with the test configurations of GNVC-VD and DiffVC, we adopt VTM-17.0 results under the low-delay configuration as ProGVC's anchor.

As demonstrated in \cref{tab:diffusioncomparison}, ProGVC achieves smaller relative improvements than GNVC-VD on DISTS, but remains competitive with DiffVC. On LPIPS, ProGVC attains the lowest relative improvement among the compared methods. Since GNVC-VD does not report NIQE results, we compare only with DiffVC on this metric, where ProGVC achieves comparable gain. For PSNR metric, ProGVC exhibits a coding-efficiency loss similar to DiffVC, while underperforming GNVC-VD. Overall, GNVC-VD delivers the strongest performance, while ProGVC is generally competitive with DiffVC across both perceptual and fidelity metrics, indicating that ProGVC is a promising new coding paradigm based on autoregressive modeling.

\begin{table*}[tb]
\caption{\label{tab:diffusioncomparison}
\textbf{ Relative BD-rate$\downarrow$ (\%) / BD-metric$\uparrow$ on MCL-JCV, UVG and HEVC B datasets using VTM-17.0 low-delay (LD) configuration as anchor.}
Diffusion-based results are taken from the published papers and are reported under 1080p/96-frame setting, whereas ProGVC is evaluated under 720p/81-frame setting. ``N/A'' indicates that BD-rate cannot be calculated due to the lack of quality overlap. ``-'' means data is not available. \textcolor{red}{\textbf{Red}} marks the best diffusion-based result.
}
\centering
\setlength{\tabcolsep}{1.5pt}
\scalebox{0.75}{
\begin{tabular}{ll lcccc}
\toprule
Dataset & \makecell{Anchor \\ (VTM-17.0 LD)} & Method & DISTS & LPIPS & NIQE & PSNR (dB) \\
\midrule

\multirow{3}{*}{MCL-JCV}
& \multirow{2}{*}{\makecell[l]{1080p, 96f}}
& GNVC-VD\cite{mao2025generative}
& \textcolor{red}{\textbf{-88.3 / 0.0882}} & \textcolor{red}{\textbf{N/A / 0.1470}} & - / - & \textcolor{red}{\textbf{N/A / -2.2600}}\rule{0pt}{2.4ex} \\

&
& DiffVC\cite{ma2025diffusion}
& N/A / 0.0555 & N/A / 0.0946 & \textcolor{red}{\textbf{N/A / 0.9003}} & N/A / -5.7767\rule{0pt}{2.4ex} \\

\cdashline{2-7}
& \makecell[l]{720p, 81f}
& \textbf{ProGVC}
& \textbf{-64.4 / 0.0416} & \textbf{-30.3 / 0.0269} & \textbf{N/A / 0.8172} & \textbf{533.5 / -5.5354}\rule{0pt}{2.4ex} \\

\midrule

\multirow{3}{*}{UVG}
& \multirow{2}{*}{\makecell[l]{1080p, 96f}}
& GNVC-VD\cite{mao2025generative}
& \textcolor{red}{\textbf{-86.5 / 0.0670}} & \textcolor{red}{\textbf{N/A / 0.1263}} & - / - & \textcolor{red}{\textbf{120.5 / -1.8390}} \rule{0pt}{2.2ex}\\

& 
& DiffVC\cite{ma2025diffusion}
& N/A / 0.0423 & N/A / 0.0657 & \textcolor{red}{\textbf{N/A / 0.9939}} & N/A / -5.4513 \rule{0pt}{2.4ex}\\

\cdashline{2-7}
& \makecell[l]{720p, 81f}
& \textbf{ProGVC}
& \textbf{-63.9 / 0.0322} & \textbf{-35.5 / 0.0236} & \textbf{N/A / 0.7499} & \textbf{744.6 / -5.2541}\rule{0pt}{2.4ex} \\

\midrule

\multirow{3}{*}{HEVC-B}
& \multirow{2}{*}{\makecell[l]{1080p, 96f}}
& GNVC-VD\cite{mao2025generative}
& \textcolor{red}{\textbf{N/A / 0.0827}} & \textcolor{red}{\textbf{N/A / 0.1907}} & - / - & \textcolor{red}{\textbf{62.76 / -1.2583}}\rule{0pt}{2.4ex} \\

&
& DiffVC\cite{ma2025diffusion}
& - / - & - / - & - / - & - / -\rule{0pt}{2.4ex} \\

\cdashline{2-7}
& \makecell[l]{720p, 81f}
& \textbf{ProGVC}
& \textbf{-69.5 / 0.0410} & \textbf{-17.3 / 0.0162} & \textbf{-43.6 / 0.6302} & \textbf{569.0 / -5.4982}\rule{0pt}{2.4ex} \\

\bottomrule
\end{tabular}
}
\end{table*}

\subsection{Comparison with More Baselines}

\begin{figure*}[t]
  \centering
  \vspace{-0.2cm}
  \begin{subfigure}{0.25\linewidth}
    \includegraphics[width=\textwidth]{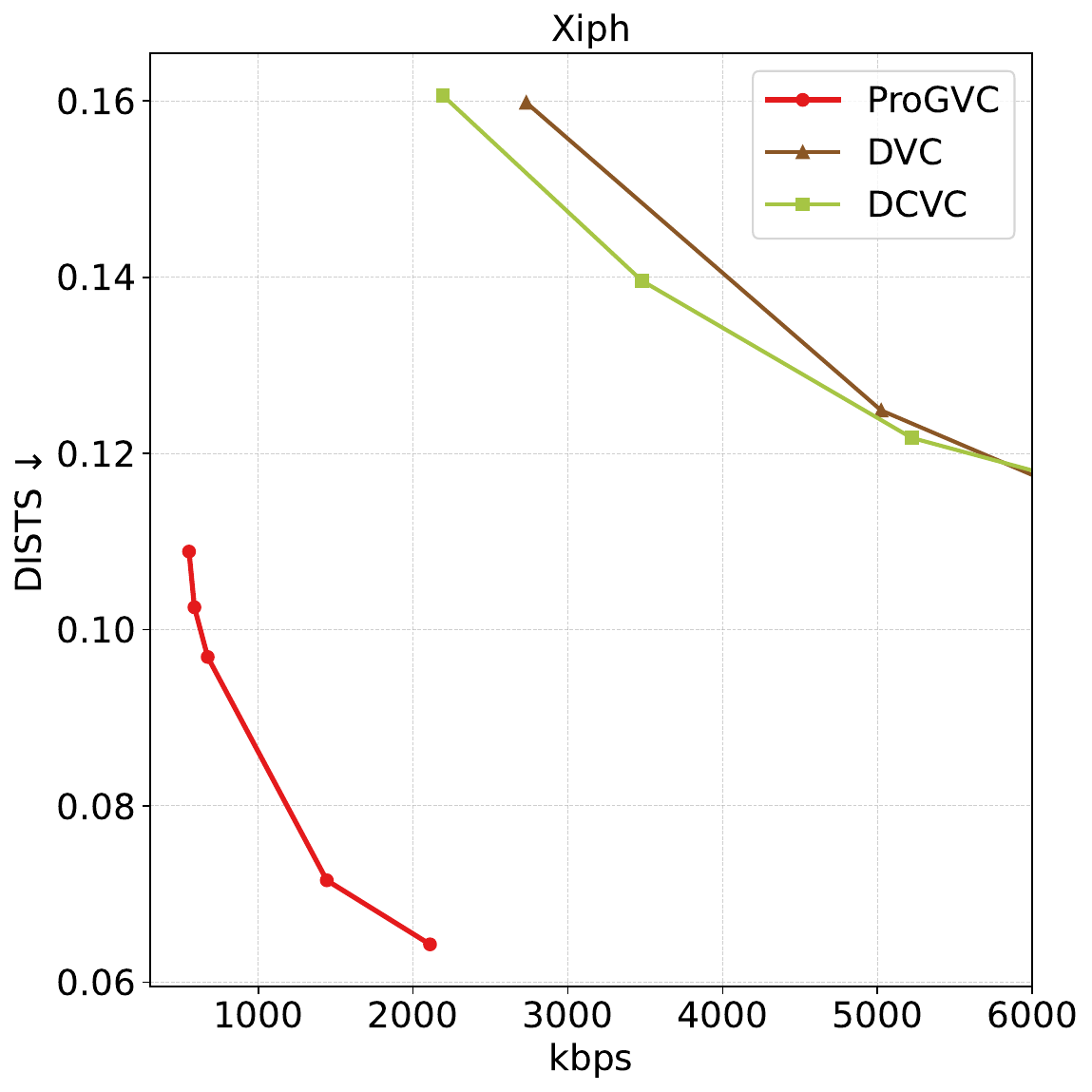}
  \end{subfigure}
    \hfill
    \hspace{-10pt}
  \begin{subfigure}{0.25\linewidth}
    \includegraphics[width=\textwidth]{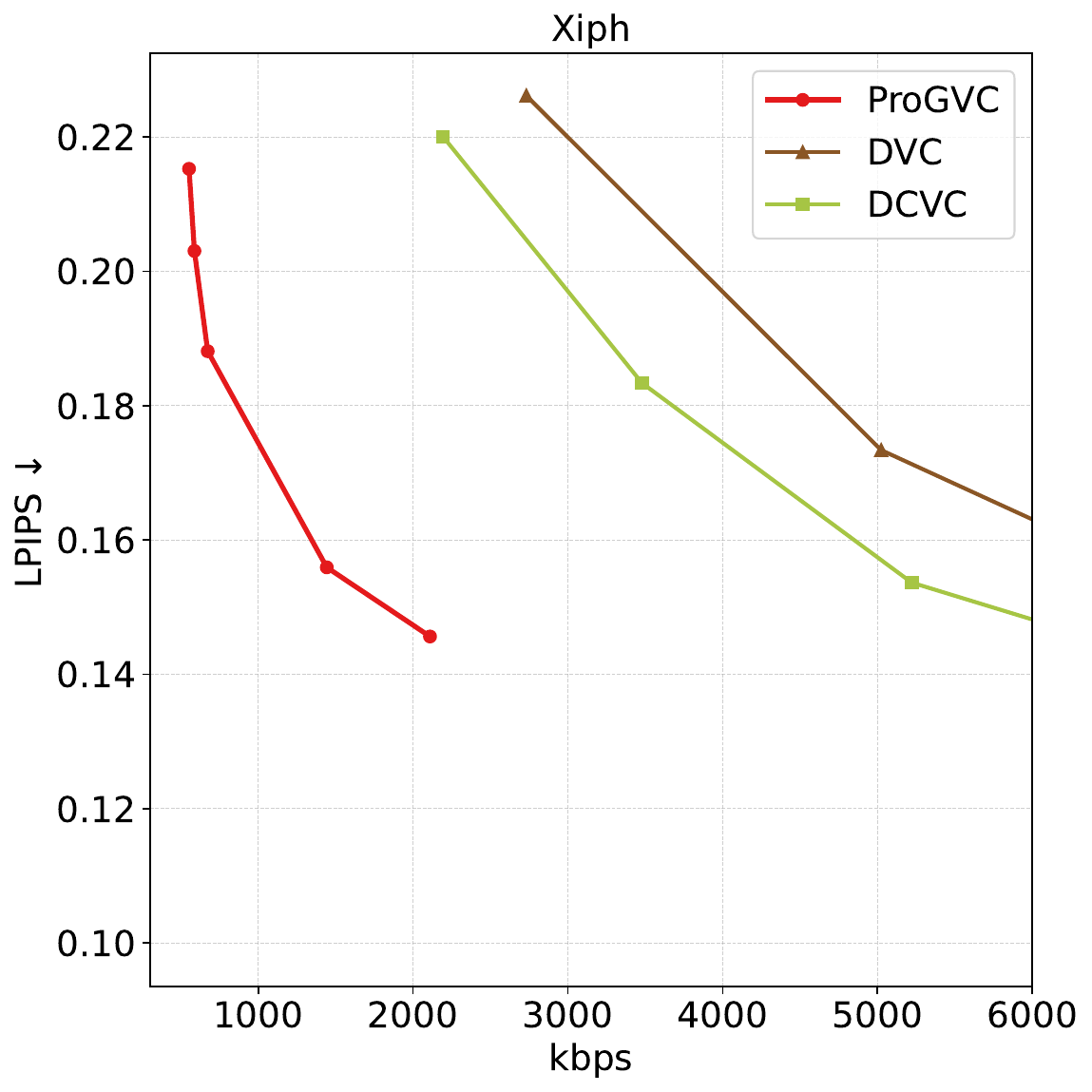}
  \end{subfigure}
      \hfill
      \hspace{-10pt}
  \begin{subfigure}{0.25\linewidth}
    \includegraphics[width=\textwidth]{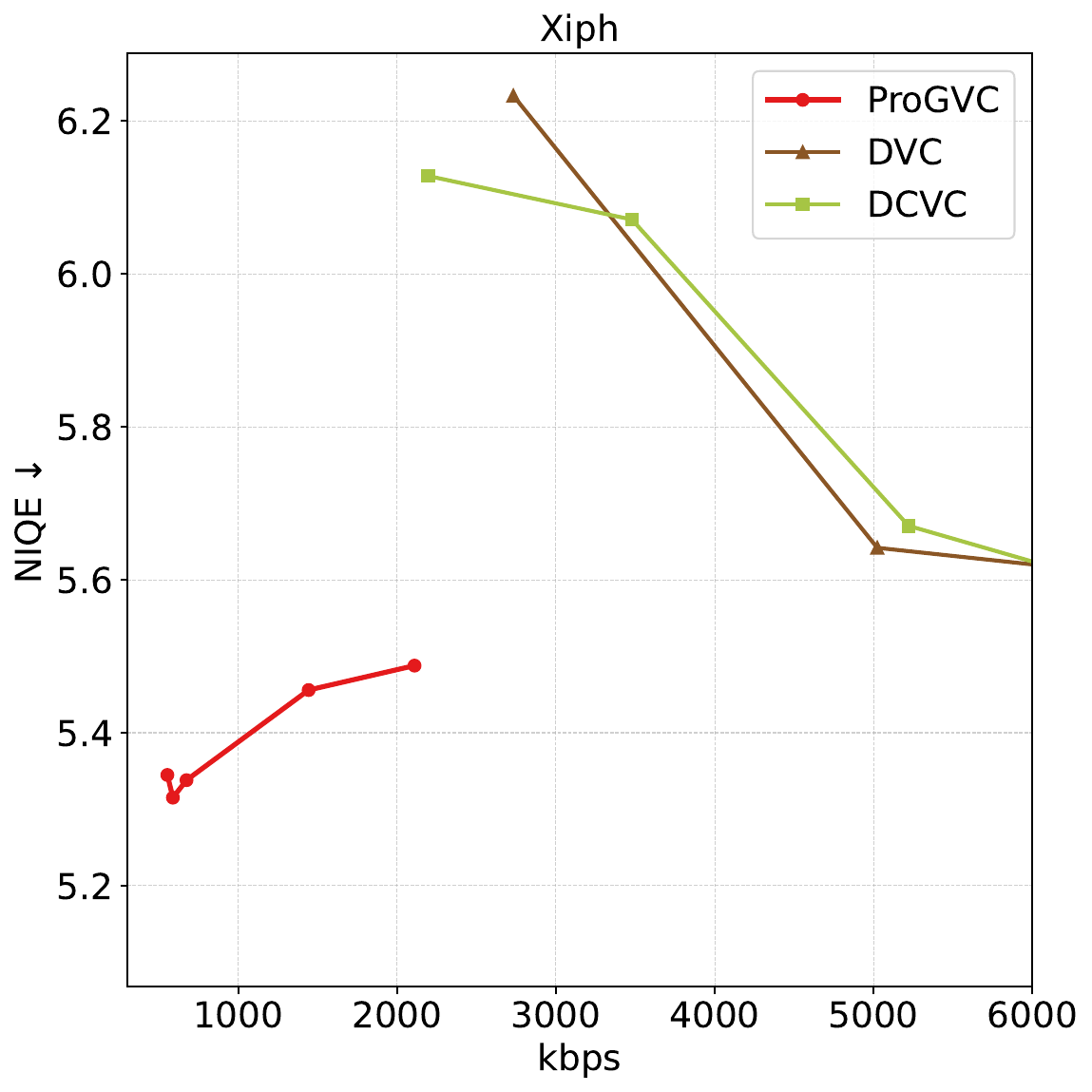}
  \end{subfigure}
      \hfill
      \hspace{-10pt}
  \begin{subfigure}{0.25\linewidth}
    \includegraphics[width=\textwidth]{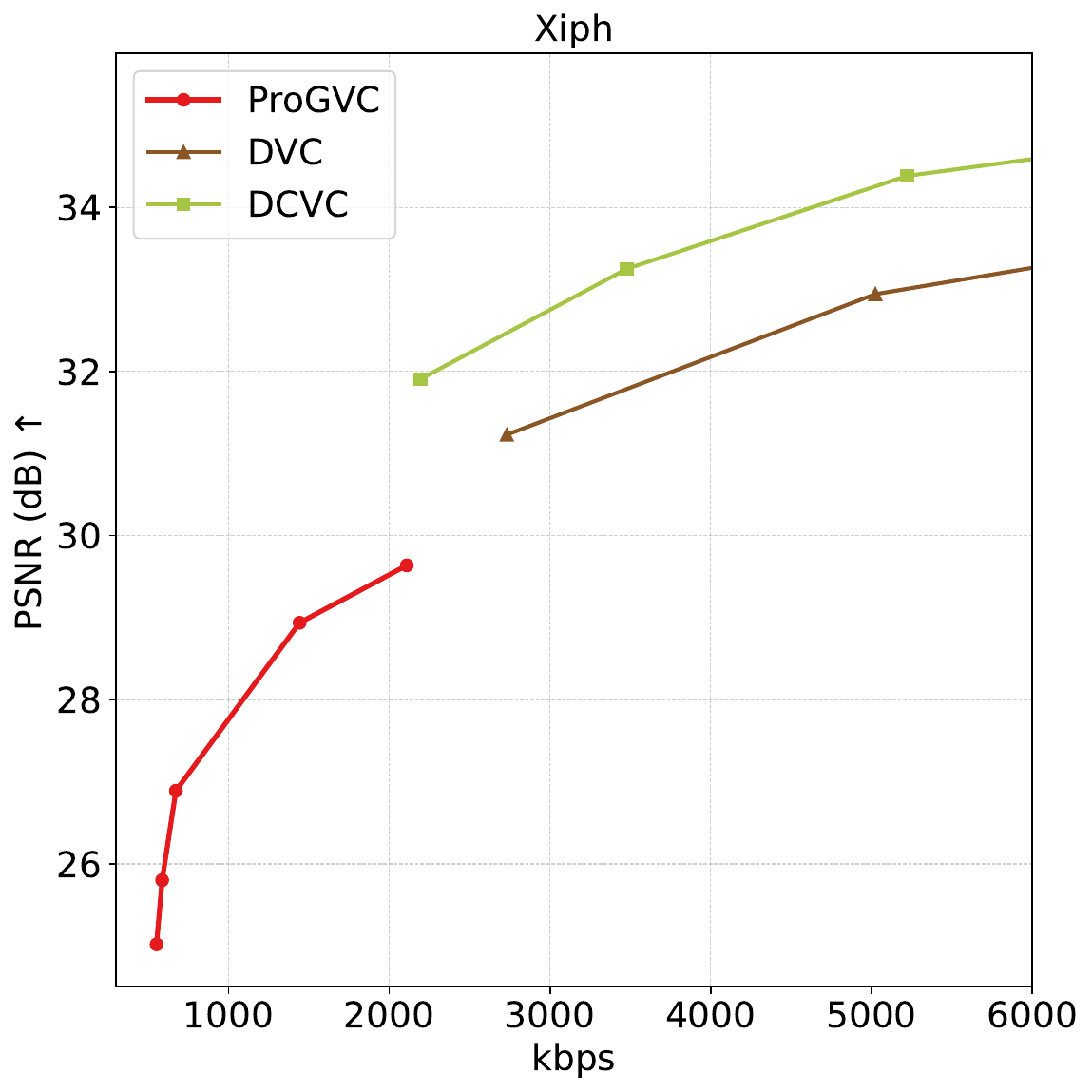}
  \end{subfigure}
    
  \begin{subfigure}{0.25\linewidth}
    \includegraphics[width=\textwidth]{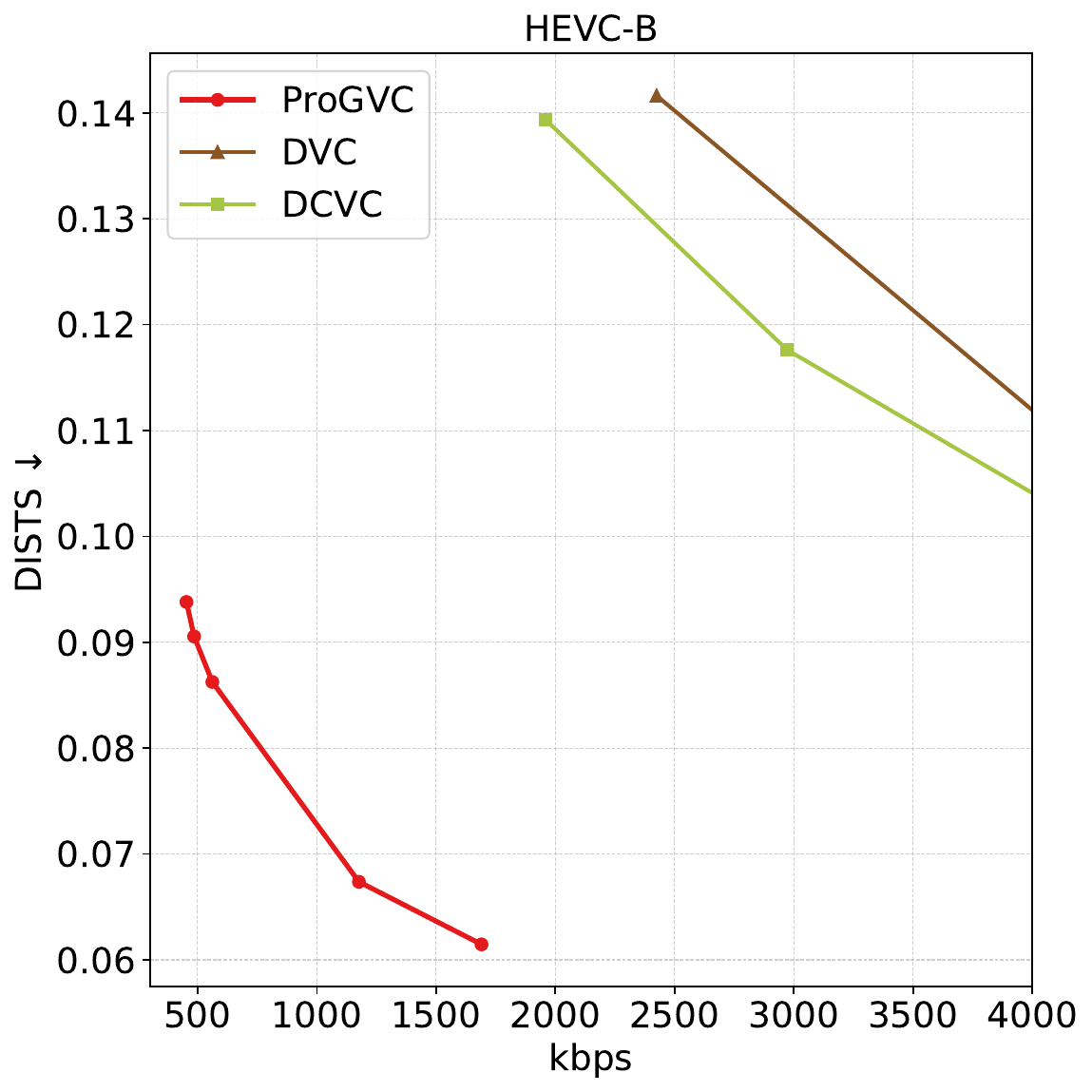}
  \end{subfigure}
    \hfill
    \hspace{-10pt}
  \begin{subfigure}{0.25\linewidth}
    \includegraphics[width=\textwidth]{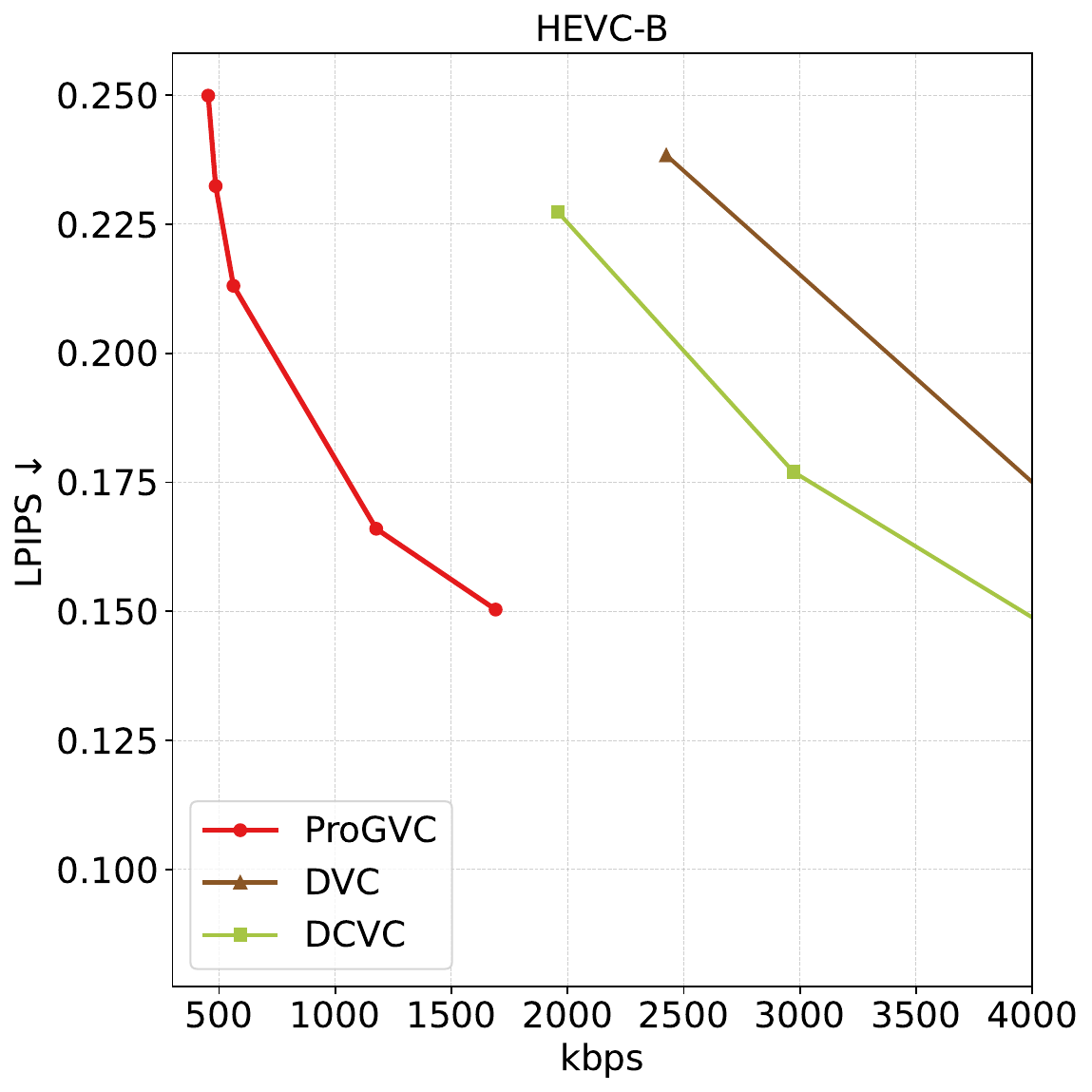}
  \end{subfigure}
      \hfill
      \hspace{-10pt}
  \begin{subfigure}{0.25\linewidth}
    \includegraphics[width=\textwidth]{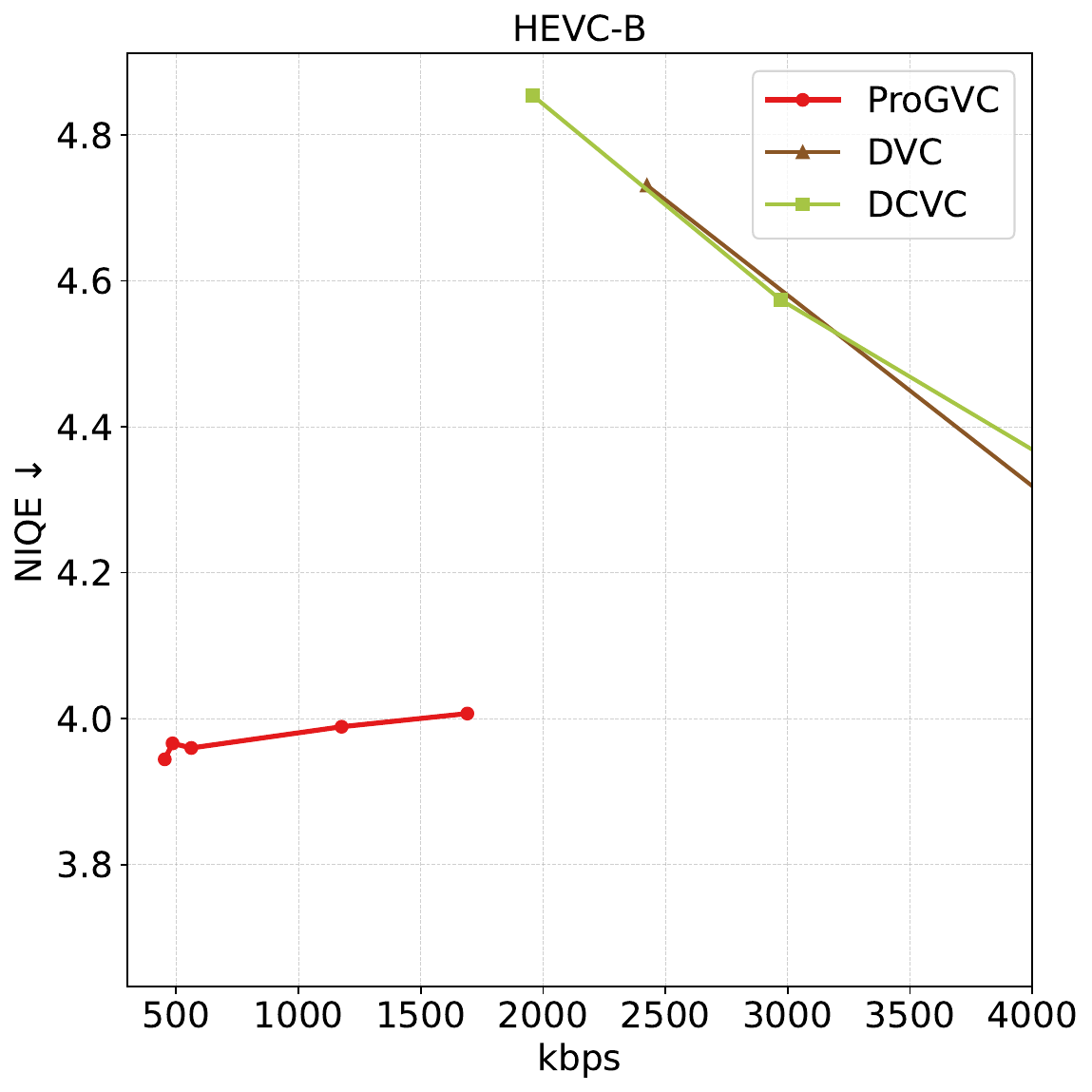}
  \end{subfigure}
      \hfill
      \hspace{-10pt}
  \begin{subfigure}{0.25\linewidth}
    \includegraphics[width=\textwidth]{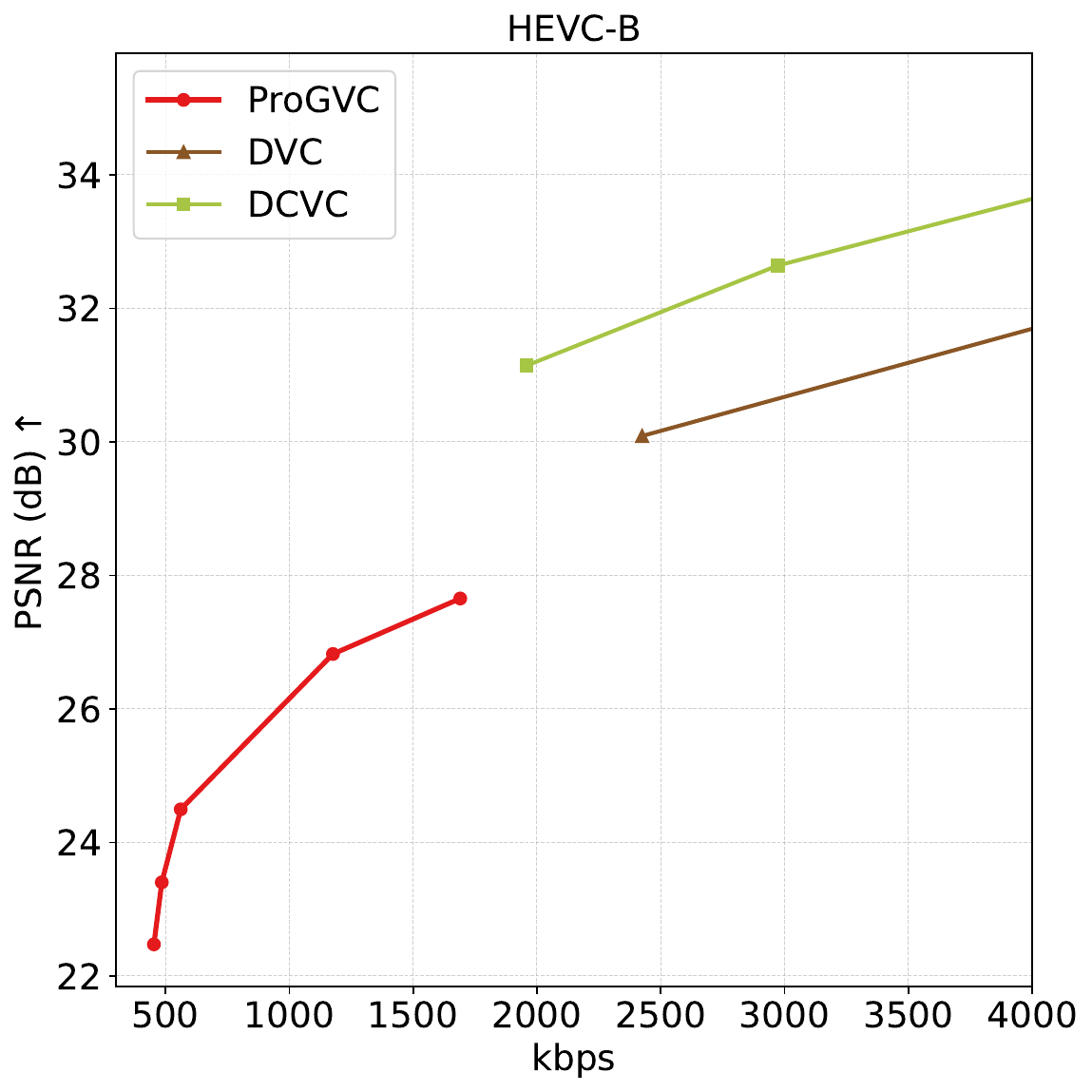}
  \end{subfigure}

  \begin{subfigure}{0.25\linewidth}
    \includegraphics[width=\textwidth]{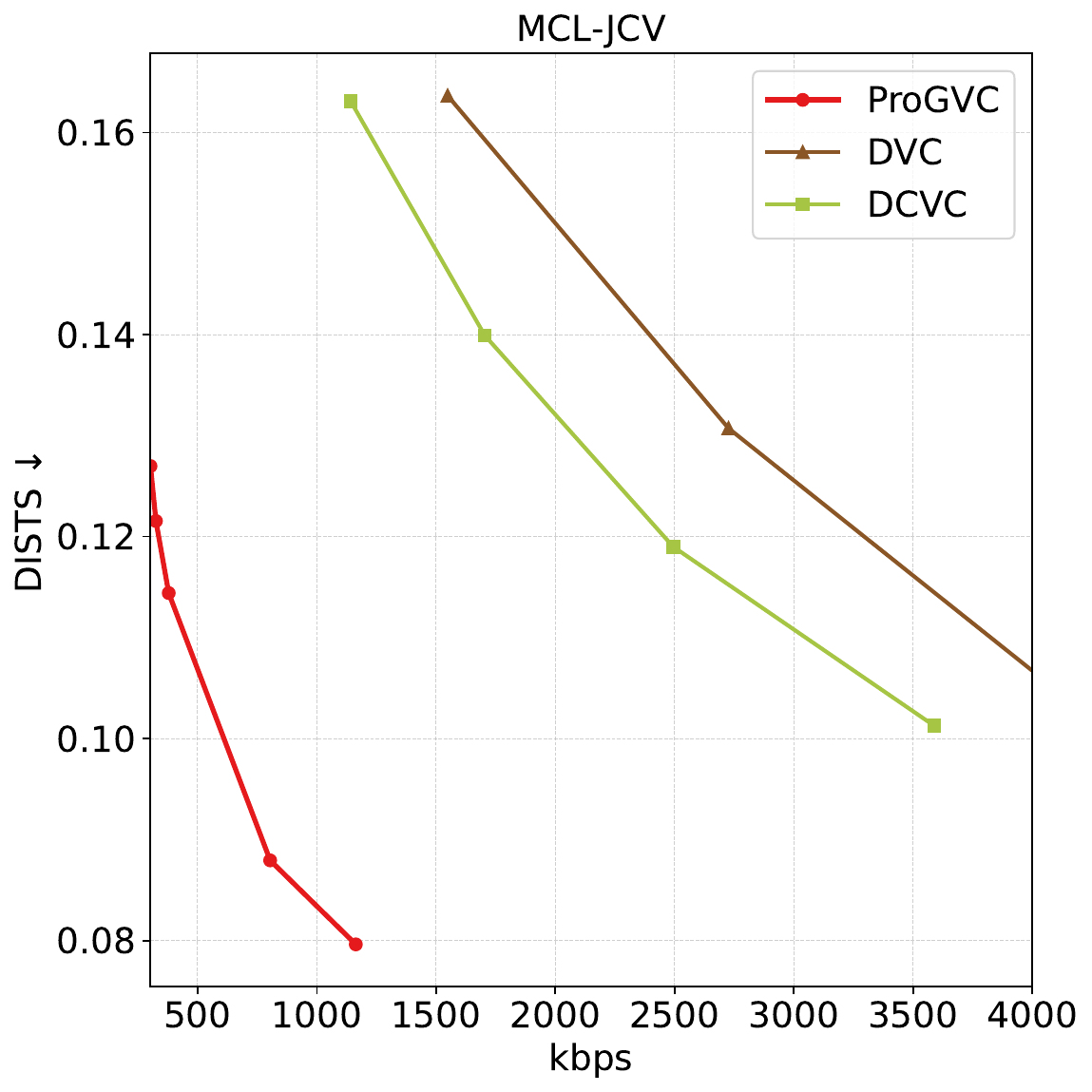}
  \end{subfigure}
    \hfill
    \hspace{-10pt}
  \begin{subfigure}{0.25\linewidth}
    \includegraphics[width=\textwidth]{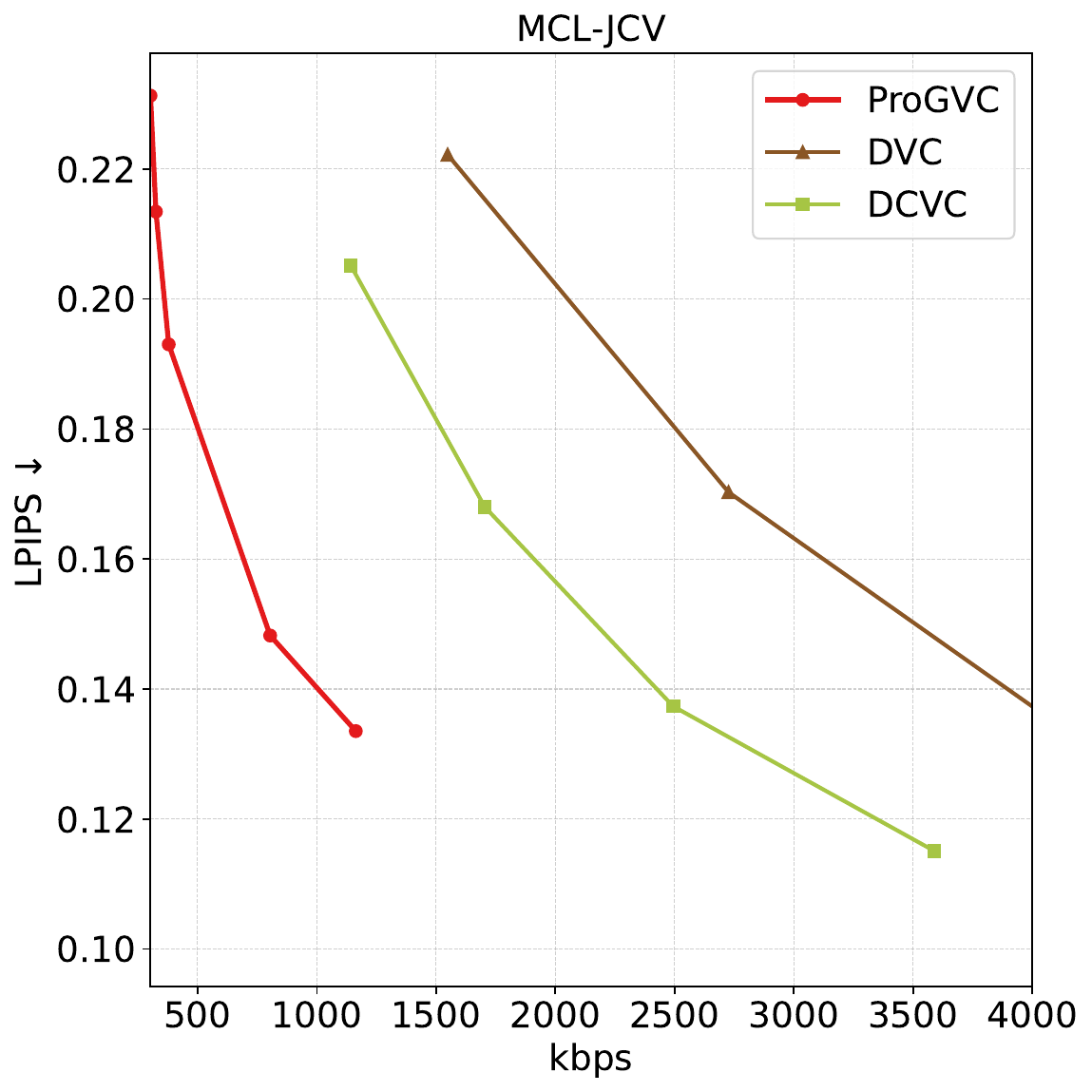}
  \end{subfigure}
      \hfill
      \hspace{-10pt}
  \begin{subfigure}{0.25\linewidth}
    \includegraphics[width=\textwidth]{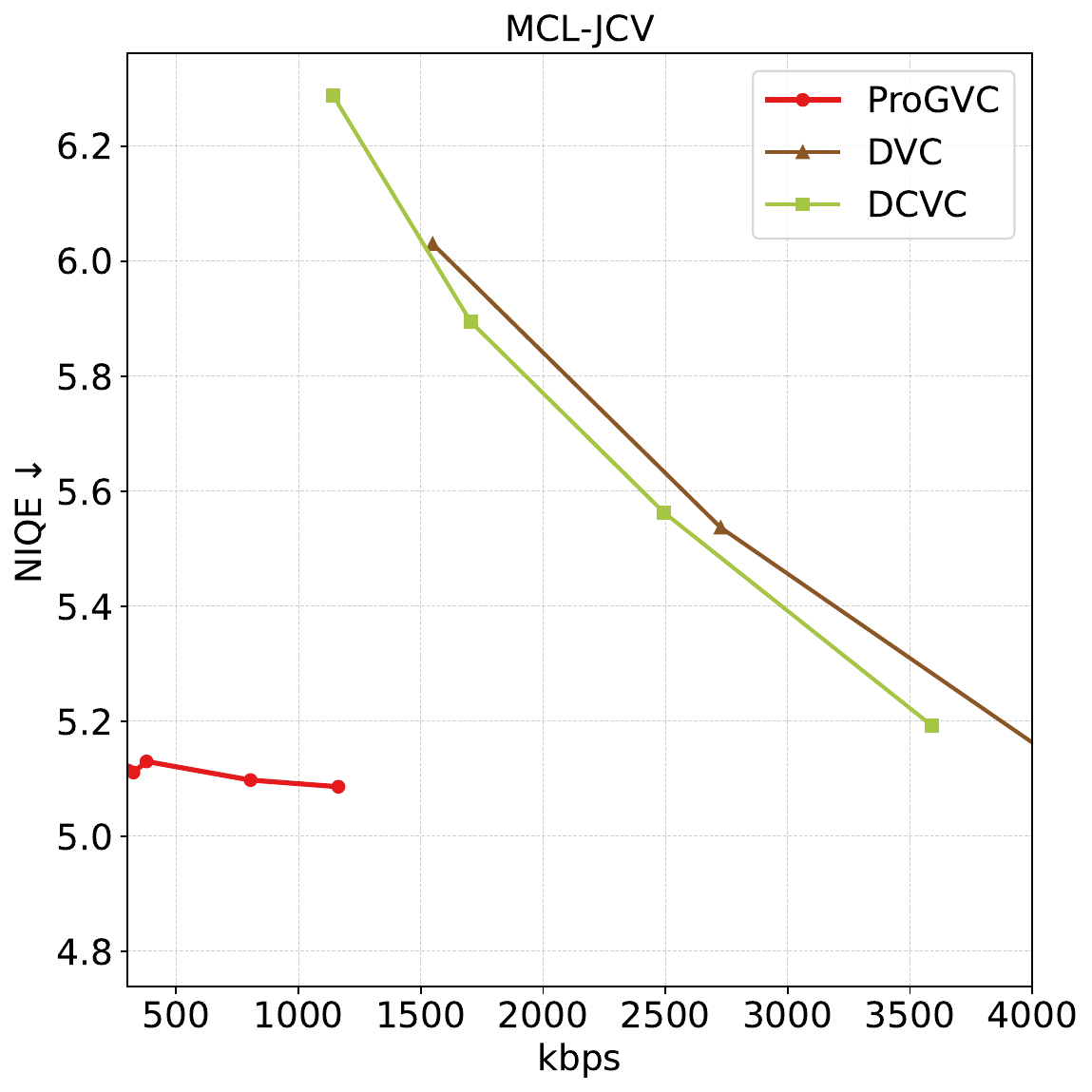}
  \end{subfigure}
      \hfill
      \hspace{-10pt}
  \begin{subfigure}{0.25\linewidth}
    \includegraphics[width=\textwidth]{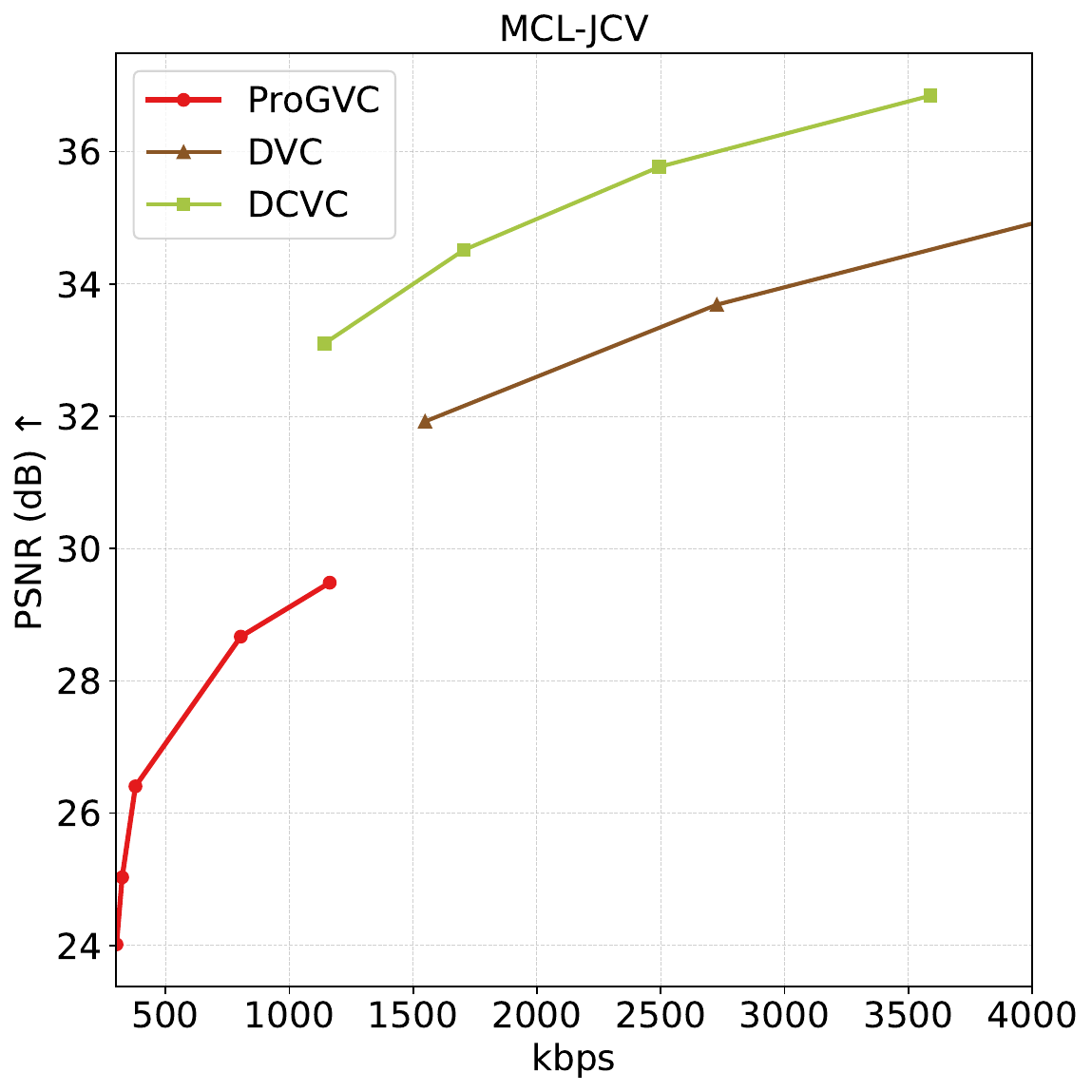}
  \end{subfigure}

  \begin{subfigure}{0.25\linewidth}
    \includegraphics[width=\textwidth]{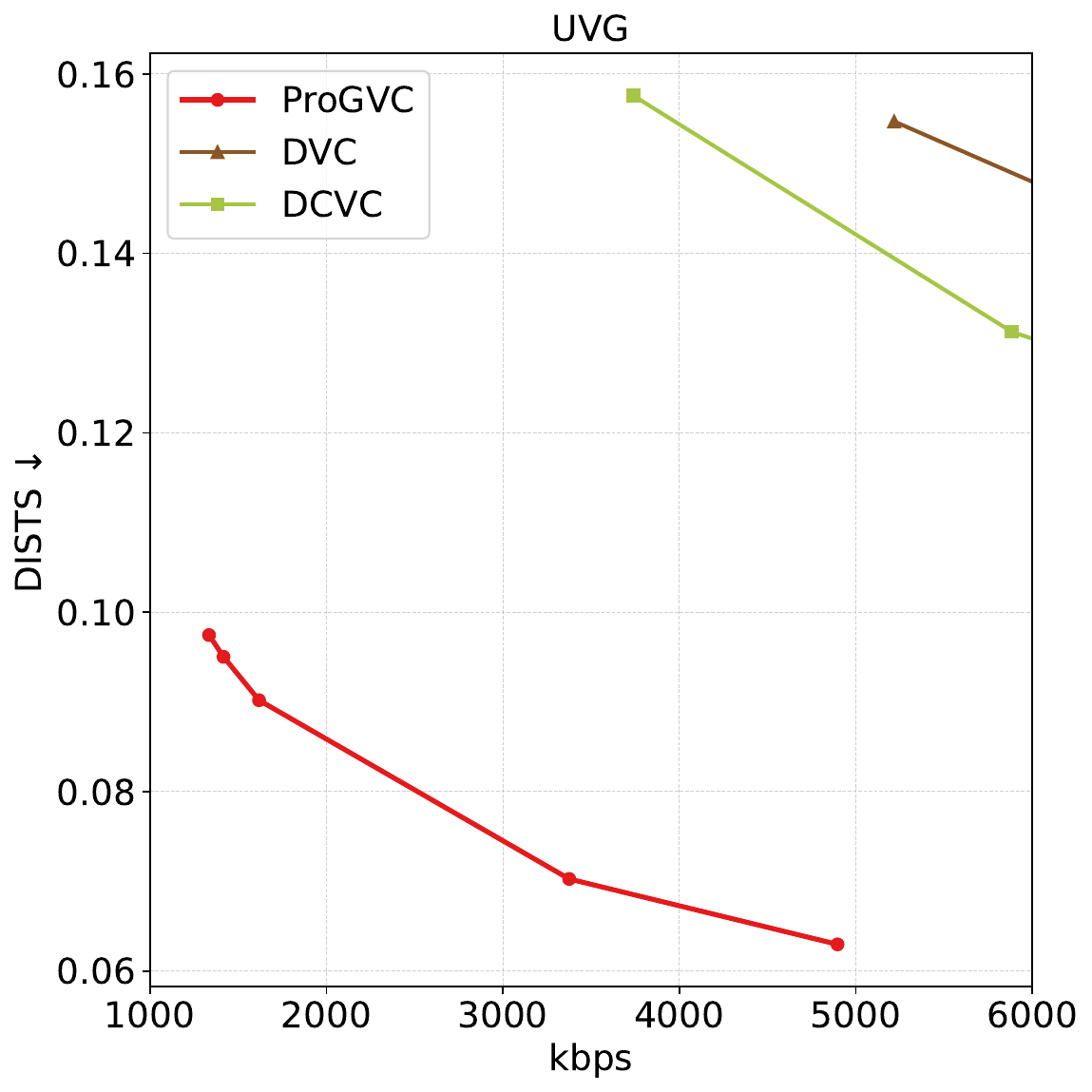}
  \end{subfigure}
    \hfill
    \hspace{-10pt}
  \begin{subfigure}{0.25\linewidth}
    \includegraphics[width=\textwidth]{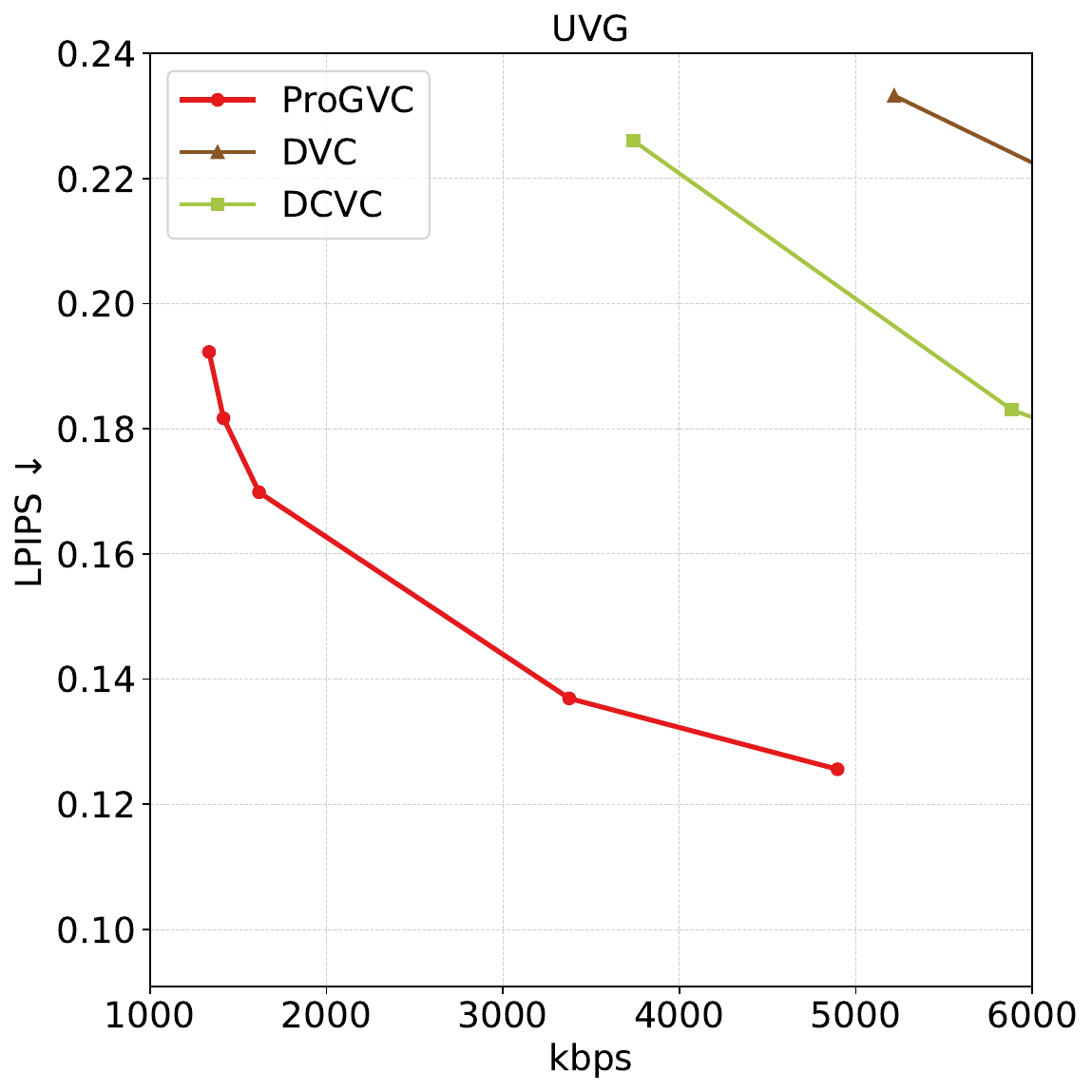}
  \end{subfigure}
      \hfill
      \hspace{-10pt}
  \begin{subfigure}{0.25\linewidth}
    \includegraphics[width=\textwidth]{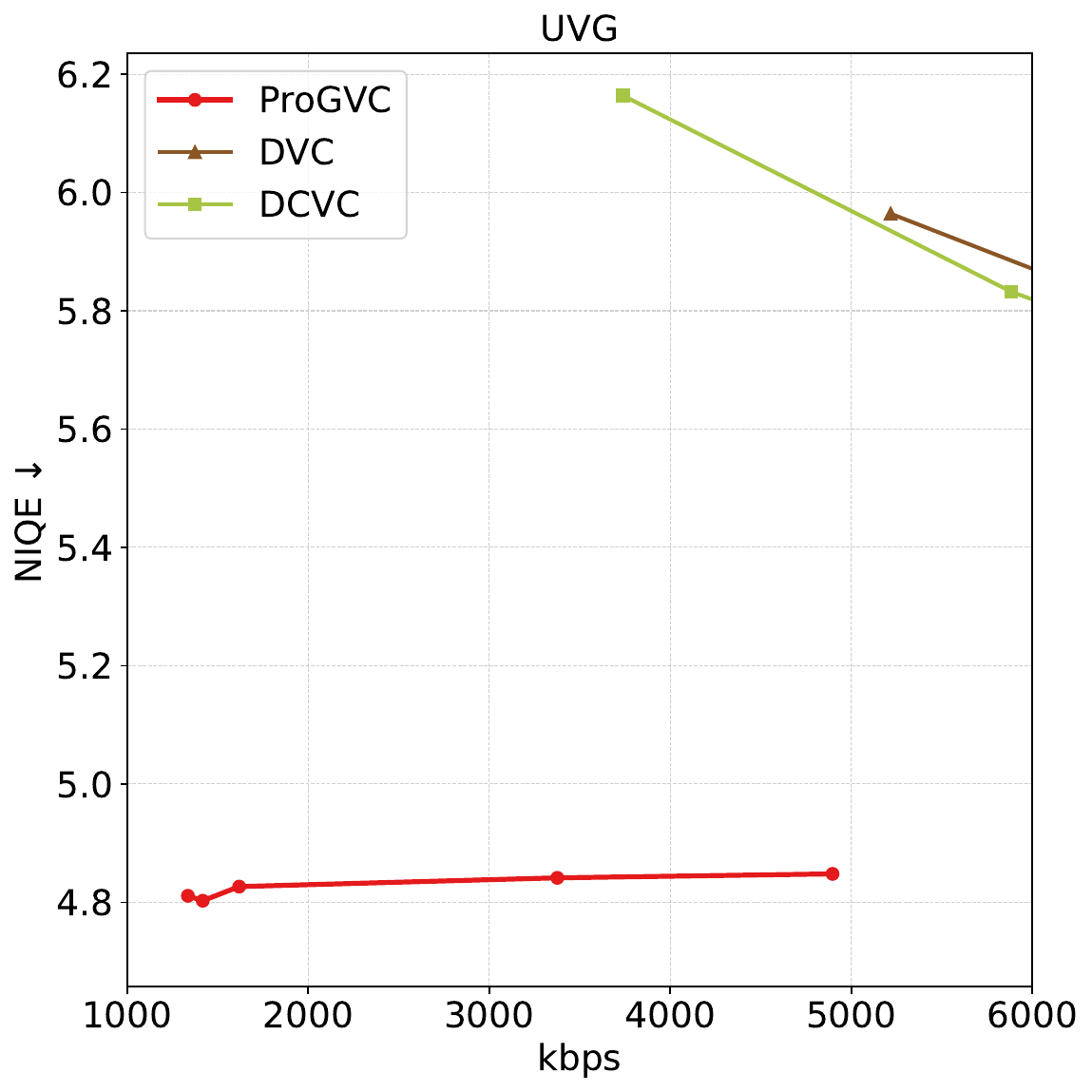}
  \end{subfigure}
      \hfill
      \hspace{-10pt}
  \begin{subfigure}{0.25\linewidth}
    \includegraphics[width=\textwidth]{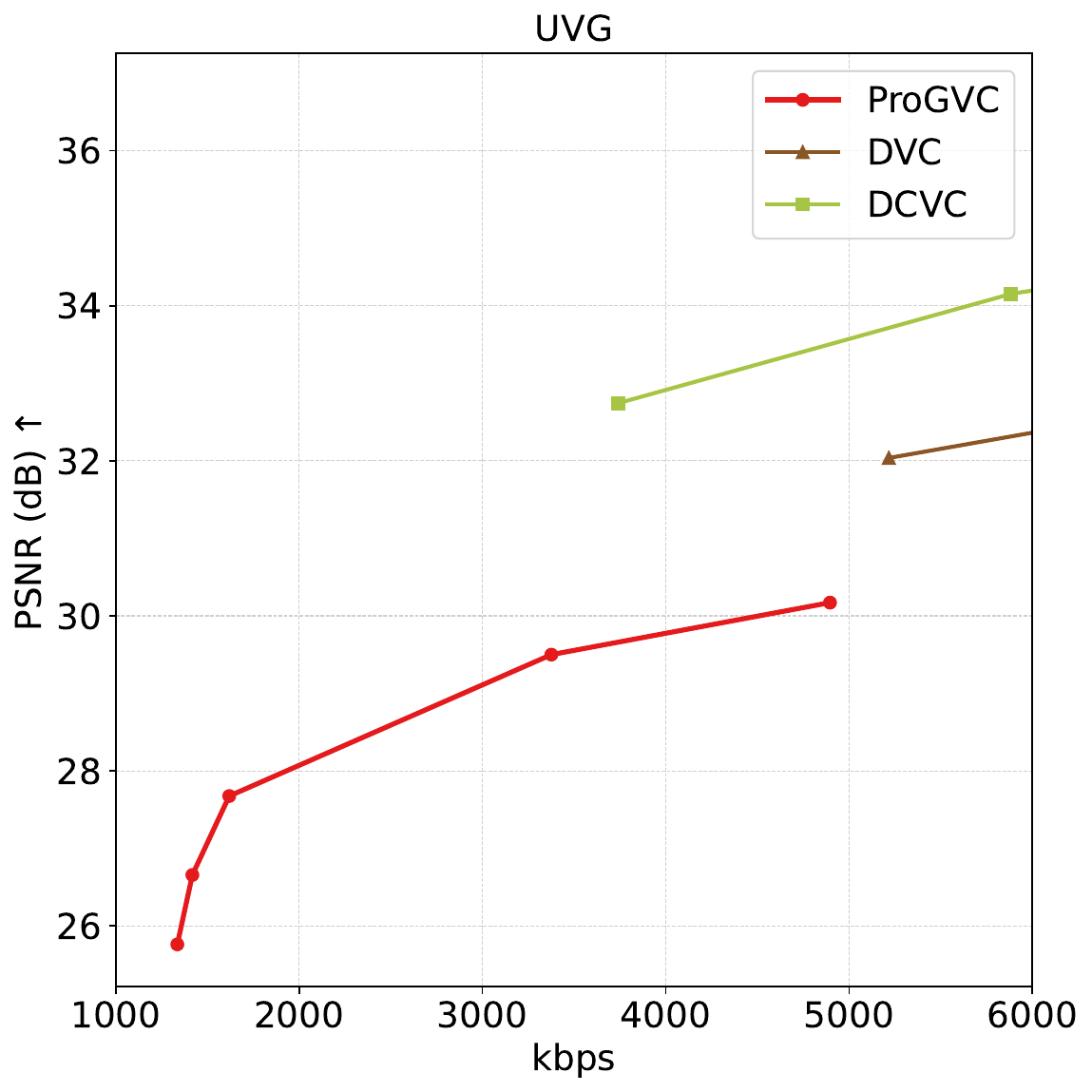}
  \end{subfigure}

  \caption{\textbf{Rate and perception/fidelity curves on Xiph, HEVC B, MCL-JCV and UVG datasets, comparing with previous pioneering neural video codecs.}
  }
  \label{fig:RDcurves_priorworks}
  \vspace{-15pt}
\end{figure*}

The codecs compared in the main manuscript and above are all well-optimized video compression methods. As an early exploration of progressive perceptual video compression with autoregressive modeling, direct comparisons of ProGVC against state-of-the-art, extensively engineered codecs may not fully reflect its longer-term potential.
Therefore, we additionally compare ProGVC with pioneering neural video codecs that have inspired many subsequent designs, aiming to provide a cleaner and more interpretable reference. Specifically, we include DVC~\cite{lu2019dvc}\footnote{https://github.com/RenYang-home/OpenDVC} and DCVC~\cite{li2021deep}\footnote{https://github.com/microsoft/DCVC/tree/main/DCVC-family/DCVC} for comparison.

\cref{fig:RDcurves_priorworks} shows detailed rate--perception/fidelity curves of all methods on the Xiph, HEVC B, MCL-JCV and UVG datasets. As reported, ProGVC consistently outperforms DVC and DCVC in terms of perceptual metrics, including DISTS, LPIPS and NIQE. Meanwhile, its fidelity loss relative to DVC remains controlled, with PSNR dropping by approximately 2 dB at comparable bitrates. These results suggest that ProGVC provides a good starting point for future investigation. \cref{tab:bdrate_priorworks} further presents the detailed BD-rate and BD-metric results.

\begin{table*}[t]
\caption{\label{tab:bdrate_priorworks}
\textbf{BD-rate$\downarrow$ (\%) / BD-metric$\uparrow$ on the Xiph, HEVC B, MCL-JCV and UVG datasets, comparing with previous pioneering neural video codecs.} ``N/A'' indicates that BD-rate cannot be calculated due to the lack of quality overlap. \textcolor{red}{\textbf{Red}} indicates the best performance. The best model in terms of PSNR metric is estimated from \cref{fig:RDcurves_priorworks}.
}
\centering
\scalebox{0.8}{
\begin{tabular}{llcccc}
\toprule
\multicolumn{1}{l}{Dataset} & \multicolumn{1}{l}{Method} &
\multicolumn{1}{c}{DISTS} & \multicolumn{1}{c}{LPIPS} & \multicolumn{1}{c}{NIQE} & \multicolumn{1}{c}{PSNR(dB)} \\
\midrule

\multirow{3}{*}{Xiph}
& \textbf{ProGVC}
  & \textcolor{red}{\textbf{0.0 / 0.0000}}   & \textcolor{red}{\textbf{0.0 / 0.0000}}   & \textcolor{red}{\textbf{0.0 / 0.0000}}     & 0.0 / 0.0000 \\
& DVC \cite{lu2019dvc}
  & 1171.4 / N/A  & 437.40 / N/A  & 1306.9 / N/A  & N/A / N/A \\
& DCVC~\cite{li2021deep}
  & 1352.8 / N/A  & 321.4 / N/A  & 302.3 / N/A  & \textcolor{red}{\textbf{N/A / N/A}} \\
\midrule

\multirow{3}{*}{HEVC-B}
& \textbf{ProGVC}
  & \textcolor{red}{\textbf{0.0 / 0.0000}}   & \textcolor{red}{\textbf{0.0 / 0.0000}}   & \textcolor{red}{\textbf{0.0 / 0.0000}}     & 0.0 / 0.0000 \\
& DVC \cite{lu2019dvc}
  & 937.9 / N/A  & 336.0 / N/A  & 731.3 / N/A  & N/A / N/A \\
& DCVC~\cite{li2021deep}
  & 999.5 / N/A  & 229.1 / N/A  & N/A / N/A  & \textcolor{red}{\textbf{N/A / N/A}} \\
\midrule

\multirow{3}{*}{MCL-JCV}
& \textbf{ProGVC}
  & \textcolor{red}{\textbf{0.0 / 0.0000}}   & \textcolor{red}{\textbf{0.0 / 0.0000}}   & \textcolor{red}{\textbf{0.0 / 0.0000}}     & 0.0 / 0.0000 \\
& DVC \cite{lu2019dvc}
  & 736.4 / N/A  & 396.0 / N/A  & 740.8 / N/A  & N/A / N/A \\
& DCVC~\cite{li2021deep}
  & 612.7 / -0.0828  & 202.2 / -0.0704  &  N/A / -1.1917 & \textcolor{red}{\textbf{N/A / 3.6673}} \\
\midrule

\multirow{3}{*}{UVG}
& \textbf{ProGVC}
  & \textcolor{red}{\textbf{0.0 / 0.0000}}   & \textcolor{red}{\textbf{0.0 / 0.0000}}   & \textcolor{red}{\textbf{0.0 / 0.0000}}     & 0.0 / 0.0000 \\
& DVC \cite{lu2019dvc}
  & 827.0 / N/A  & 433.8 / N/A  & 1120.3 / N/A  & N/A / N/A \\
& DCVC~\cite{li2021deep}
  & 780.7 / -0.0839  & 278.6 /  -0.0831  & N/A / -1.2234  & \textcolor{red}{\textbf{N/A / 3.2310}} \\
\bottomrule

\end{tabular}
} 

\end{table*}

\section{Study on Dropping Intra Scales}
In this part, we explain the reason for not dropping intra scales in the proposed ProGVC. First, as shown in \cref{tab:intrabpp}, intra scale bits account for a relatively small fraction of the total sequence bit cost, with no more than 16\% across all test datasets. This suggests that discarding inter scales is generally a more effective approach for adaptive rate control.

Based on inter scale dropping, we further investigate the effect of additionally dropping intra scales. To remain consistent with the main manuscript, we conduct this study on the Xiph dataset. Because the intra scale conditioning strategy used for inter scale modeling affects the results of intra scale dropping, we evaluate all four types of intra scale conditioning strategies. As demonstrated in \cref{tab:dropintrascales}, compression efficiency consistently degrades as more intra scales are dropped. This is expected because intra scales are crucial for temporal alignment when modeling inter scales' distribution. Discarding intra scales therefore weakens the context for inter scale probability estimation, leading to error propagation in context modeling and less accurate probability prediction. 

Besides, it's noticed that after dropping the largest intra scale, two intra scale conditioning strategies yield compression efficiency improvement, \ie conditioning inter scales on the largest intra scale or on the same resolution scale. However, as indicated by \cref{tab:negativeratio}, more than half of test sequences exhibit BD-rate degradation. This suggests that the average improvement is driven by a subset of sequences and does not generalize reliably. Consequently, intra scale dropping is not adopted in our final design.

\begin{table}[t]
\centering
\setlength{\tabcolsep}{4pt}
\caption{\label{tab:intrabpp}
\textbf{Minimum and maximum ratios of intra scale bit cost to the total sequence bit cost across datasets.}}
\scalebox{0.9}{
\begin{tabular}{lcccc}
\toprule
 & Xiph & HEVC-B & MCL-JCV & UVG \\
\midrule
Min ratio & 11.1\% & 13.8\% & 12.2\% & 13.4\% \\
Max ratio & 15.8\% & 15.4\% & 15.6\% & 15.5\% \\
\bottomrule
\end{tabular}
}
\end{table}

\begin{table}

\centering
\setlength{\tabcolsep}{2pt}
\caption{\label{tab:dropintrascales}
\textbf{BD-rate$\downarrow$ (\%) / BD-metric$\uparrow$ on the Xiph, analyzing the impact of dropping intra scales under different intra scale conditioning strategies.} The anchor is ProGVC, \ie inter scales attend to the largest intra scale and no intra scale is dropped. ``drop1'' and ``drop2'' denote dropping the largest intra scale and the two largest intra scales, respectively. \textcolor{blue}{\textbf{Blue}} indicates method that shows compression efficiency improvement.}
\scalebox{0.9}{
\begin{tabular}{l l ccc}
\toprule
\makecell[l]{Conditioning \\ Strategy}
&  \makecell{Dropping \\ Strategy} & DISTS & LPIPS & PSNR \\
\midrule

\multirow{3}{*}{\makecell{largest }}
& no drop & 0.0 / 0.0000 & 0.0 / 0.0000 & 0.0 / 0.0000  \\
& \textcolor{blue}{\textbf{drop1}} & \textcolor{blue}{\textbf{-13.3 / 0.0020}} & \textcolor{blue}{\textbf{-17.4 / 0.0015}} & \textcolor{blue}{\textbf{-22.5 / 0.1398}} \\
& drop2 & 16.8 / -0.0060 & 9.3 / -0.0092 & 16.4 / -0.9551 \\

\cdashline{2-5}

\multirow{2}{*}{\makecell{no intra }}
& drop1 & 7.2 / -0.0021 & 9.7 / 0.0050 & 13.4 / -0.4636 \\
& drop2 & 1.4 / -0.0010 & 0.6 / -0.0033 & 6.5 / -0.4308 \\

\cdashline{2-5}

\multirow{2}{*}{smallest}
& drop1 & 14.3 / -0.0045 & 9.8 / -0.0050 & 13.9 / -0.4784 \\
& drop2 & 9.1 / -0.0030 & 1.2 / -0.0032 & 7.6 / -0.4549 \\

\cdashline{2-5}

\multirow{2}{*}{\makecell{same resolution \\}}
& \textcolor{blue}{\textbf{drop1}} & \textcolor{blue}{\textbf{-11.2 / 0.0019}} & \textcolor{blue}{\textbf{-13.6 / 0.0009}} & \textcolor{blue}{\textbf{-12.0 / 0.0055}} \\
& drop2 & 5.6 / -0.0037 & -5.0 / 0.0053 & 4.6 / -0.6926 \\

\bottomrule
\end{tabular}
}

\end{table}

\begin{table}[t]
\centering
\caption{\textbf{Ratio of sequences exhibiting BD-rate performance degradation.} ``drop1'' denotes dropping the largest intra scale.}
\label{tab:negativeratio}
\setlength{\tabcolsep}{10pt}
\scalebox{0.9}{
\begin{tabular}{lcccc}
\toprule
\makecell{Conditioning \\ Strategy} & \makecell{Dropping \\ Strategy} & DISTS & LPIPS & PSNR \\
\midrule
largest & drop1 & 50.00\% & 62.50\% & 50.00\% \\
same resolution & drop1 & 56.25\% & 56.25\% & 56.25\% \\
\bottomrule
\end{tabular}
}
\end{table}

\section{Comparison with Progressive Image Compression}
We further compare the proposed progressive video compression paradigm with a frame-wise progressive image compression baseline to analyze the effect of temporal modeling. In the image compression baseline, each frame is compressed independently using the proposed intra scale coding scheme of ProGVC, where the context model only captures cross-scale dependencies within a single frame. In contrast, ProGVC additionally models temporal dependencies across frames.

As reported in \cref{tab:abla_attn1_new}, ProGVC consistently outperforms the frame-wise progressive image compression baseline, achieving approximately 50\% BD-rate gain. This result indicates that incorporating temporal dependencies into the autoregressive context model effectively reduces the bitrate of inter scale tokens. In addition, we compare ProGVC with two alternative attention mask variants that attend to different sets of previous scales. Consistent with the findings in the main manuscript, the results further validate that ProGVC achieves the most favorable BD-rate–complexity trade-off, demonstrating a better balance between context utilization and computational cost than the alternative designs.

With the frame-wise progressive image compression as anchor, we further evaluate different intra scale reference strategies for inter scale coding. As shown in \cref{tab:abla_attn2_new}, conditioning on the largest intra scale, as adopted in ProGVC, again achieves the best compression performance while maintaining comparable computational complexity across all variants. This observation is consistent with the findings in the main manuscript, suggesting that the largest intra scale provides richer details that serve as stronger contextual guidance for predicting inter scale tokens.

\begin{table*}[t]
\caption{
\textbf{Progressive image compression vs. progressive video compression with three temporal-aligned attention masks.} The anchor is frame-wise progressive image compression scheme, where each frame adopts the proposed intra scale coding of ProGVC. ``AVG'' denotes the average BD-rate performance over DISTS, LPIPS and PSNR. The per-frame encoding and decoding times (Enc. T and Dec. T) are averaged over the lowest and highest bitrates.
}
\label{tab:abla_attn1_new}
\centering
\resizebox{0.76\textwidth}{!}{
\begin{tabular}{l|cccc|c|c}
\hline
\multirow{2}{*}{Attn. Variants} 
& \multicolumn{4}{c|}{BD-rate$\downarrow$ (\%)}  
& \multirow{2}{*}{Enc. T (s) $\downarrow$} 
& \multirow{2}{*}{Dec. T (s) $\downarrow$} \\
\cline{2-5}
& \textbf{AVG} & DISTS & LPIPS & PSNR & & \\
\hline
ProGVC
& \textbf{-51.6} & -51.5 & -49.4 & -53.8 & 0.56 & 1.53 \\
\hline
Self-only 
& \textbf{-46.4} & -46.3  & -43.8 & -49.2 & 0.50 & 1.23 \\
Full causal 
& \textbf{-52.9} & -53.4 & -51.2 & -54.0 & 0.64 & 1.96 \\
\bottomrule
\end{tabular}
}
\end{table*}

\begin{table*}[t]
\caption{\label{tab:abla_attn2_new}
\textbf{Effect of different intra scale references for inter scale coding.} The anchor is frame-wise progressive image compression scheme, with each frame utilizes the proposed intra scale coding of ProGVC. ``AVG'' denotes the average BD-rate performance over DISTS, LPIPS and PSNR. The per-frame encoding and decoding times (Enc. T and Dec. T) are averaged over the lowest and highest bitrates. 
}
\centering
\resizebox{0.84\textwidth}{!}{
\begin{tabular}{l|cccc|c|c}
\hline
\multirow{2}{*}{Attn. Variants} 
& \multicolumn{4}{c|}{BD-rate$\downarrow$ (\%)} 
& \multirow{2}{*}{Enc. T (s) $\downarrow$} 
& \multirow{2}{*}{Dec. T (s) $\downarrow$}\\
\cline{2-5}
& \textbf{AVG} & DISTS & LPIPS & PSNR & & \\
\hline
ProGVC (largest scale) 
& \textbf{-51.6}  & -51.5  & -49.4  & -53.8  & 0.56 & 1.53 \\
\hline
No intra reference 
& \textbf{-24.5} & -25.8  & -22.3 & -25.4 & 0.53 & 1.46 \\
Smallest scale 
& \textbf{-24.4}  & -25.9 & -22.1 & -25.1  & 0.53 & 1.46 \\
Same resolution scale 
& \textbf{-45.5} & -45.6  & -43.9  & -46.9 & 0.54 & 1.49 \\
\bottomrule
\end{tabular}
}

\end{table*}

\section{More Qualitative Results}
\begin{figure*}[t]
    \centering

    {\scriptsize
        \begin{tabular}{ccccc}
            Ground Truth & VTM-17.0~\cite{bross2021overview} & SEVC~\cite{bian2025augmented} & PLVC~\cite{yang2022perceptual} & \textbf{ProGVC} \\
        \end{tabular}
    }

    \includegraphics[width=\textwidth]{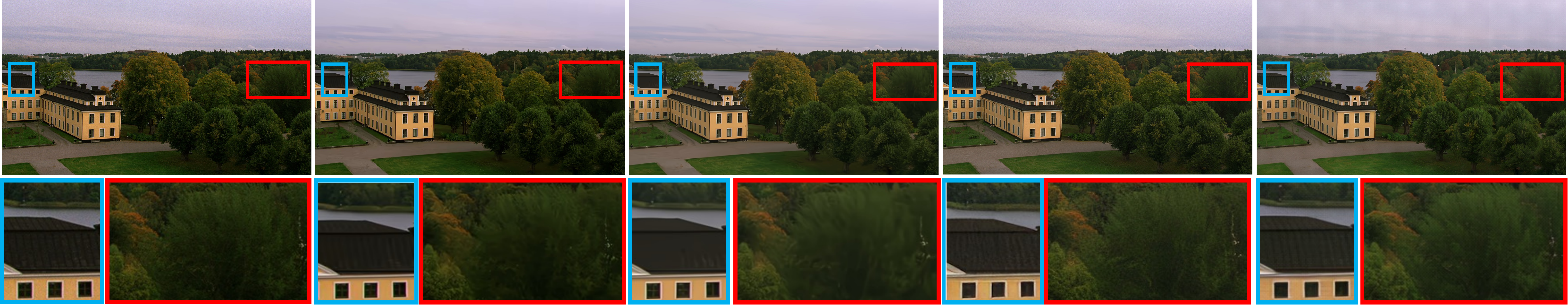}

    {\scriptsize
        \begin{tabular}{ccccc}
            Bitrate $\downarrow$ / DISTS $\downarrow$
            & 632.5 kbps/0.1254
            & 625.6 kbps/0.1544
            & 1035.9 kbps/0.0869
            & 584.8 kbps/0.0784 \\
        \end{tabular}
    }

    \includegraphics[width=\textwidth]{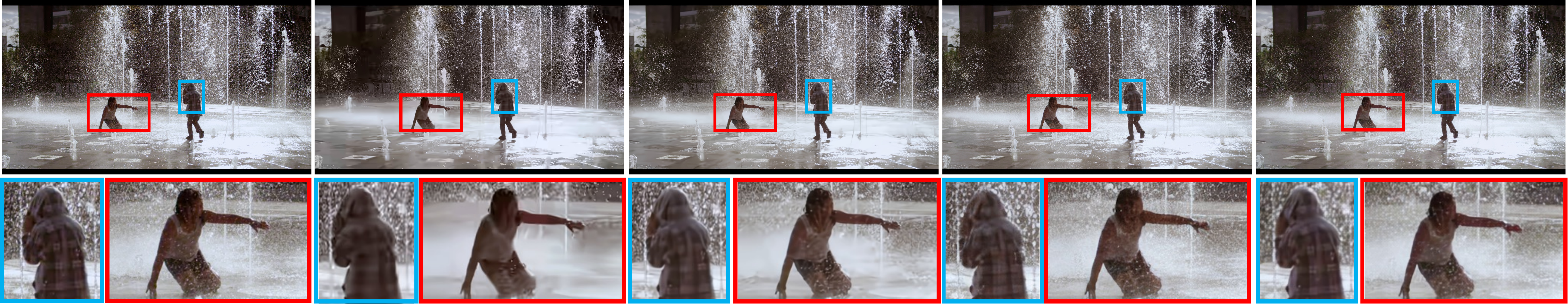}

    {\scriptsize
        \begin{tabular}{ccccc}
            Bitrate $\downarrow$ / DISTS $\downarrow$
            & 371.6 kbps/0.1654
            & 3404.14 kbps/0.1160
            & 1413.93 kbps/0.0575
            & 366.4 kbps/0.0554 \\
        \end{tabular}
    }

    \includegraphics[width=\textwidth]{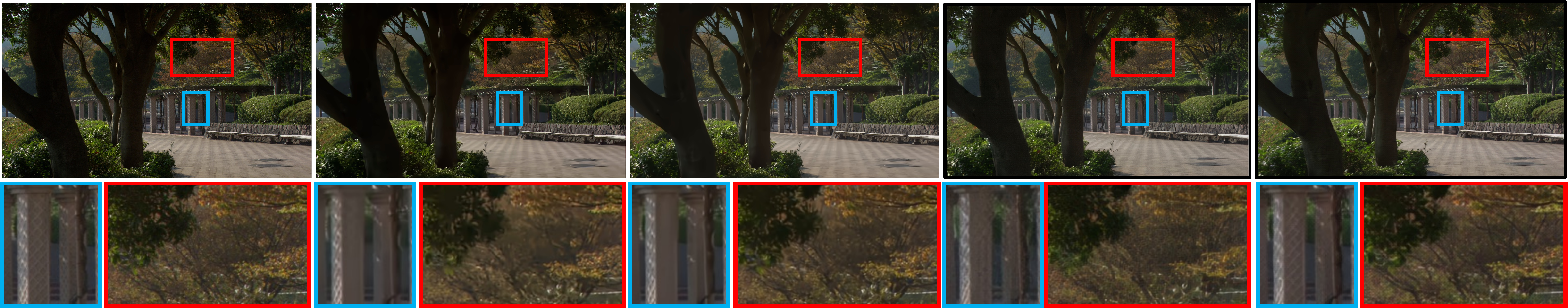}

    {\scriptsize
        \begin{tabular}{ccccc}
            Bitrate$\downarrow$/DISTS$\downarrow$
            & 395.5 kbps/0.1291
            & 754.13 kbps/0.1281
            & 682.1 kbps/0.0956
            & 345.5 kbps/0.0688 \\
        \end{tabular}
    }

    \includegraphics[width=\textwidth]{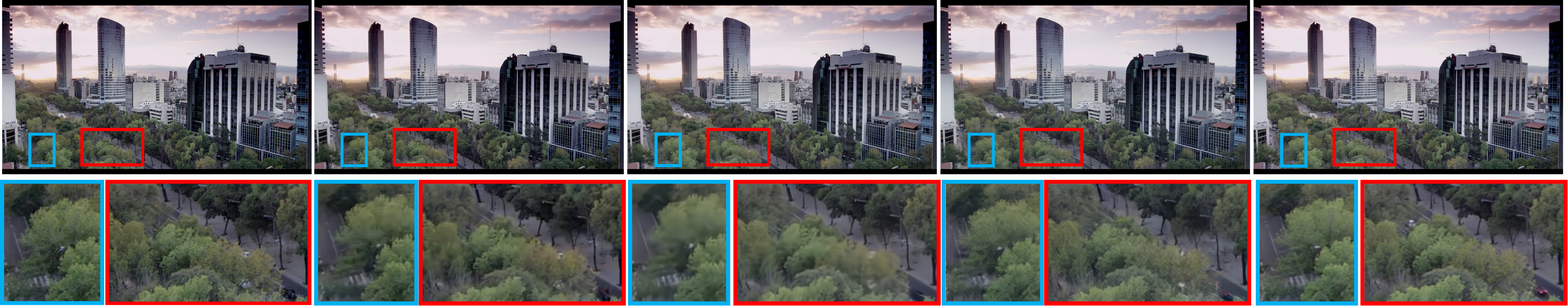}

    {\scriptsize
        \begin{tabular}{ccccc}
            Bitrate $\downarrow$ / DISTS $\downarrow$
            & 475.55 kbps/0.0736
            & 579.04 kbps/0.1097
            & 720.46 kbps/0.1047
            & 390.1 kbps/0.0443 \\
        \end{tabular}
    }

    \includegraphics[width=\textwidth]{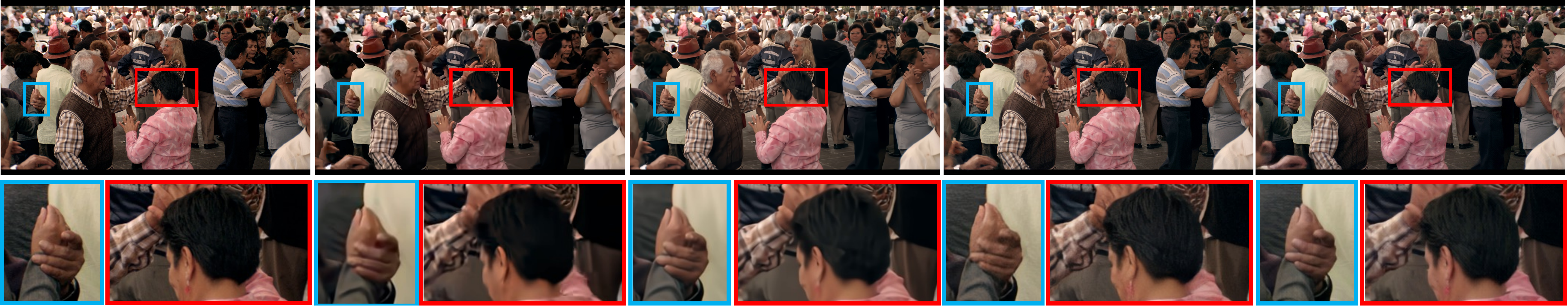}

    {\scriptsize
        \begin{tabular}{ccccc}
            Bitrate $\downarrow$ / DISTS $\downarrow$
            & 450.8 kbps/0.1218
            & 990.92 kbps/0.0937
            & 1024.8 kbps/0.0524
            & 445.02 kbps/0.0382 \\
        \end{tabular}
    }

    \caption{\textbf{Visual comparisons with baselines across several test sequences.} Zoom-in patches highlight area with perceptual differences.}
    \label{fig:visualizationCombined_new}
\end{figure*}

We provide additional qualitative comparisons at below. As demonstrated in \cref{fig:visualizationCombined_new}, ProGVC produces more realistic details and delivers higher reconstruction fidelity with the lowest bitrate cost, outperforming all compared baselines.

%% file: main.bib
@String(ICLR  = {Int. Conf. Learn. Represent.})

@String(IJCAI = {IJCAI})

@String(ICIP  = {IEEE Int. Conf. Image Process.})

@String(TOG   = {ACM Trans. Graph.})

@String(TCSVT = {IEEE Trans. Circuit Syst. Video Technol.})

@String(ICLR  = {ICLR})

@String(ICIP  = {ICIP})

@String(TOG   = {ACM TOG})

@String(TCSVT = {IEEE TCSVT})

@inproceedings{loshchilov2017decoupled,
    title={Decoupled Weight Decay Regularization},
    author={Ilya Loshchilov and Frank Hutter},
    booktitle=ICLR,
    year={2019},
}

@article{van2016conditional,
  title={Conditional image generation with pixelcnn decoders},
  author={Van den Oord, Aaron and Kalchbrenner, Nal and Espeholt, Lasse and Vinyals, Oriol and Graves, Alex and others},
  journal={Advances in neural information processing systems},
  volume={29},
  year={2016}
}

@inproceedings{esser2021taming,
  title={Taming transformers for high-resolution image synthesis},
  author={Esser, Patrick and Rombach, Robin and Ommer, Bjorn},
  booktitle={Proceedings of the IEEE/CVF conference on computer vision and pattern recognition},
  pages={12873--12883},
  year={2021}
}

@article{yan2021videogpt,
  title={Videogpt: Video generation using vq-vae and transformers},
  author={Yan, Wilson and Zhang, Yunzhi and Abbeel, Pieter and Srinivas, Aravind},
  journal={arXiv preprint arXiv:2104.10157},
  year={2021}
}

@article{wang2024omnitokenizer,
  title={Omnitokenizer: A joint image-video tokenizer for visual generation},
  author={Wang, Junke and Jiang, Yi and Yuan, Zehuan and Peng, Bingyue and Wu, Zuxuan and Jiang, Yu-Gang},
  journal={Advances in Neural Information Processing Systems},
  volume={37},
  pages={28281--28295},
  year={2024}
}

@article{tian2024visual,
  title={Visual autoregressive modeling: Scalable image generation via next-scale prediction},
  author={Tian, Keyu and Jiang, Yi and Yuan, Zehuan and Peng, Bingyue and Wang, Liwei},
  journal={Advances in neural information processing systems},
  volume={37},
  pages={84839--84865},
  year={2024}
}

@inproceedings{han2025infinity,
  title={Infinity: Scaling bitwise autoregressive modeling for high-resolution image synthesis},
  author={Han, Jian and Liu, Jinlai and Jiang, Yi and Yan, Bin and Zhang, Yuqi and Yuan, Zehuan and Peng, Bingyue and Liu, Xiaobing},
  booktitle={Proceedings of the Computer Vision and Pattern Recognition Conference},
  pages={15733--15744},
  year={2025}
}

@inproceedings{liuinfinitystar,
  title={InfinityStar: Unified Spacetime AutoRegressive Modeling for Visual Generation},
  author={Liu, Jinlai and Han, Jian and Yan, Bin and Zhu, Fengda and Wang, Xing and Jiang, Yi and PENG, BINGYUE and Yuan, Zehuan and others},
  booktitle={The Thirty-ninth Annual Conference on Neural Information Processing Systems}
}

@inproceedings{van2016pixel,
  title={Pixel recurrent neural networks},
  author={Van Den Oord, A{\"a}ron and Kalchbrenner, Nal and Kavukcuoglu, Koray},
  booktitle={International conference on machine learning},
  pages={1747--1756},
  year={2016},
  organization={PMLR}
}

@inproceedings{
dong2026echogen,
title={EchoGen: Generating Visual Echoes in Any Scene via Feed-Forward Subject-Driven Auto-Regressive Model},
author={Ruixiao Dong and Zhendong Wang and Keli Liu and Li Li and Ying Chen and Kai Li and Daowen Li and Houqiang Li},
booktitle={The Fourteenth International Conference on Learning Representations},
year={2026},
url={https://openreview.net/forum?id=ctmyCjo18u}
}

@article{liu2025scaleweaver,
  title={ScaleWeaver: Weaving Efficient Controllable T2I Generation with Multi-Scale Reference Attention},
  author={Liu, Keli and Wang, Zhendong and Zhou, Wengang and Xu, Shaodong and Dong, Ruixiao and Li, Houqiang},
  journal={arXiv preprint arXiv:2510.14882},
  year={2025}
}

@inproceedings{quvisual,
  title={Visual Autoregressive Modeling for Image Super-Resolution},
  author={Qu, Yunpeng and Yuan, Kun and Hao, Jinhua and Zhao, Kai and Xie, Qizhi and Sun, Ming and Zhou, Chao},
  booktitle={Forty-second International Conference on Machine Learning}
}

@inproceedings{
zhang2026autoregressivebased,
title={Autoregressive-based Progressive Coding for Ultra-Low Bitrate Image Compression},
author={Ziyuan Zhang and Yichong Xia and Bin Chen and Tianwei Zhang and Hao Wang and Han Qiu},
booktitle={The Fourteenth International Conference on Learning Representations},
year={2026},
url={https://openreview.net/forum?id=FXu4G5T5QZ}
}

@article{wiegand2003overview,
  title={Overview of the H. 264/AVC video coding standard},
  author={Wiegand, Thomas and Sullivan, Gary J and Bj{\o}ntegaard, Gisle and Luthra, Ajay},
  journal={IEEE Transactions on circuits and systems for video technology},
  volume={13},
  number={7},
  pages={560--576},
  year={2003},
  publisher={IEEE}
}

@article{sullivan2012overview,
  title={Overview of the high efficiency video coding (HEVC) standard},
  author={Sullivan, Gary J and Ohm, Jens-Rainer and Han, Woo-Jin and Wiegand, Thomas},
  journal={IEEE Transactions on circuits and systems for video technology},
  volume={22},
  number={12},
  pages={1649--1668},
  year={2012},
  publisher={IEEE}
}

@article{bross2021overview,
  title={Overview of the versatile video coding (VVC) standard and its applications},
  author={Bross, Benjamin and Wang, Ye-Kui and Ye, Yan and Liu, Shan and Chen, Jianle and Sullivan, Gary J and Ohm, Jens-Rainer},
  journal={TCSVT},
  volume={31},
  number={10},
  pages={3736--3764},
  year={2021},
  publisher={IEEE}
}

@inproceedings{lu2019dvc,
  title={Dvc: An end-to-end deep video compression framework},
  author={Lu, Guo and Ouyang, Wanli and Xu, Dong and Zhang, Xiaoyun and Cai, Chunlei and Gao, Zhiyong},
  booktitle={Proceedings of the IEEE/CVF conference on computer vision and pattern recognition},
  pages={11006--11015},
  year={2019}
}

@inproceedings{li2023neural,
  title={Neural video compression with diverse contexts},
  author={Li, Jiahao and Li, Bin and Lu, Yan},
  booktitle={Proceedings of the IEEE/CVF conference on computer vision and pattern recognition},
  pages={22616--22626},
  year={2023}
}

@inproceedings{li2024neural,
  title={Neural video compression with feature modulation},
  author={Li, Jiahao and Li, Bin and Lu, Yan},
  booktitle={Proceedings of the IEEE/CVF Conference on Computer Vision and Pattern Recognition},
  pages={26099--26108},
  year={2024}
}

@article{sheng2025bi,
  title={Bi-directional deep contextual video compression},
  author={Sheng, Xihua and Li, Li and Liu, Dong and Wang, Shiqi},
  journal={IEEE Transactions on Multimedia},
  year={2025},
  publisher={IEEE}
}

@inproceedings{bian2025augmented,
  title={Augmented deep contexts for spatially embedded video coding},
  author={Bian, Yifan and Tang, Chuanbo and Li, Li and Liu, Dong},
  booktitle={Proceedings of the Computer Vision and Pattern Recognition Conference},
  pages={2094--2104},
  year={2025}
}

@inproceedings{zhang2021dvc,
  title={Dvc-p: Deep video compression with perceptual optimizations},
  author={Zhang, Saiping and Mrak, Marta and Herranz, Luis and Blanch, Marc G{\'o}rriz and Wan, Shuai and Yang, Fuzheng},
  booktitle={2021 International Conference on Visual Communications and Image Processing (VCIP)},
  pages={1--5},
  year={2021},
  organization={IEEE}
}

@inproceedings{yang2022perceptual,
  title={Perceptual Learned Video Compression with Recurrent Conditional GAN.},
  author={Yang, Ren and Timofte, Radu and Van Gool, Luc},
  booktitle={IJCAI},
  pages={1537--1544},
  year={2022}
}

@inproceedings{li2023high,
  title={High visual-fidelity learned video compression},
  author={Li, Meng and Shi, Yibo and Wang, Jing and Huang, Yunqi},
  booktitle={Proceedings of the 31st ACM International Conference on Multimedia},
  pages={8057--8066},
  year={2023}
}

@inproceedings{rombach2022high,
  title={High-resolution image synthesis with latent diffusion models},
  author={Rombach, Robin and Blattmann, Andreas and Lorenz, Dominik and Esser, Patrick and Ommer, Bj{\"o}rn},
  booktitle={Proceedings of the IEEE/CVF conference on computer vision and pattern recognition},
  pages={10684--10695},
  year={2022}
}

@inproceedings{peebles2023scalable,
  title={Scalable diffusion models with transformers},
  author={Peebles, William and Xie, Saining},
  booktitle={Proceedings of the IEEE/CVF international conference on computer vision},
  pages={4195--4205},
  year={2023}
}

@article{podell2023sdxl,
  title={Sdxl: Improving latent diffusion models for high-resolution image synthesis},
  author={Podell, Dustin and English, Zion and Lacey, Kyle and Blattmann, Andreas and Dockhorn, Tim and M{\"u}ller, Jonas and Penna, Joe and Rombach, Robin},
  journal={arXiv preprint arXiv:2307.01952},
  year={2023}
}

@article{ma2025diffusion,
  title={Diffusion-based perceptual neural video compression with temporal diffusion information reuse},
  author={Ma, Wenzhuo and Chen, Zhenzhong},
  journal={ACM Transactions on Multimedia Computing, Communications and Applications},
  volume={21},
  number={12},
  pages={1--22},
  year={2025},
  publisher={ACM New York, NY}
}

@article{mao2025generative,
  title={Generative Neural Video Compression via Video Diffusion Prior},
  author={Mao, Qi and Cheng, Hao and Yang, Tinghan and Jin, Libiao and Ma, Siwei},
  journal={arXiv preprint arXiv:2512.05016},
  year={2025}
}

@article{ma2025diffvc,
  title={DiffVC-OSD: One-Step Diffusion-based Perceptual Neural Video Compression Framework},
  author={Ma, Wenzhuo and Chen, Zhenzhong},
  journal={arXiv preprint arXiv:2508.07682},
  year={2025}
}

@article{qi2025generative,
  title={Generative latent coding for ultra-low bitrate image and video compression},
  author={Qi, Linfeng and Jia, Zhaoyang and Li, Jiahao and Li, Bin and Li, Houqiang and Lu, Yan},
  journal={IEEE Transactions on Circuits and Systems for Video Technology},
  year={2025},
  publisher={IEEE}
}

@inproceedings{mentzer2022neural,
  title={Neural video compression using gans for detail synthesis and propagation},
  author={Mentzer, Fabian and Agustsson, Eirikur and Ball{\'e}, Johannes and Minnen, David and Johnston, Nick and Toderici, George},
  booktitle={European Conference on Computer Vision},
  pages={562--578},
  year={2022},
  organization={Springer}
}

@article{goodfellow2020generative,
  title={Generative adversarial networks},
  author={Goodfellow, Ian and Pouget-Abadie, Jean and Mirza, Mehdi and Xu, Bing and Warde-Farley, David and Ozair, Sherjil and Courville, Aaron and Bengio, Yoshua},
  journal={Communications of the ACM},
  volume={63},
  number={11},
  pages={139--144},
  year={2020},
  publisher={ACM New York, NY, USA}
}

@article{li2024extreme,
  title={Extreme video compression with pre-trained diffusion models},
  author={Li, Bohan and Liu, Yiming and Niu, Xueyan and Bai, Bo and Deng, Lei and G{\"u}nd{\"u}z, Deniz},
  journal={arXiv preprint arXiv:2402.08934},
  year={2024}
}

@article{bengio2003neural,
  title={A neural probabilistic language model},
  author={Bengio, Yoshua and Ducharme, R{\'e}jean and Vincent, Pascal and Jauvin, Christian},
  journal={Journal of machine learning research},
  volume={3},
  number={Feb},
  pages={1137--1155},
  year={2003}
}

@article{voronov2024switti,
  title={Switti: Designing scale-wise transformers for text-to-image synthesis},
  author={Voronov, Anton and Kuznedelev, Denis and Khoroshikh, Mikhail and Khrulkov, Valentin and Baranchuk, Dmitry},
  journal={arXiv preprint arXiv:2412.01819},
  year={2024}
}

@inproceedings{wang2016mcl,
  title={MCL-JCV: a JND-based H. 264/AVC video quality assessment dataset},
  author={Wang, Haiqiang and Gan, Weihao and Hu, Sudeng and Lin, Joe Yuchieh and Jin, Lina and Song, Longguang and Wang, Ping and Katsavounidis, Ioannis and Aaron, Anne and Kuo, C-C Jay},
  booktitle={2016 IEEE international conference on image processing (ICIP)},
  pages={1509--1513},
  year={2016},
  organization={IEEE}
}

@misc{xiph,
  title        = {Video test media},
  author       = {Xiph.org},
  note         = {\url{https://media.xiph.org/video/derf/}}
}

@article{ding2020image,
  title={Image quality assessment: Unifying structure and texture similarity},
  author={Ding, Keyan and Ma, Kede and Wang, Shiqi and Simoncelli, Eero P},
  journal={IEEE transactions on pattern analysis and machine intelligence},
  volume={44},
  number={5},
  pages={2567--2581},
  year={2020},
  publisher={IEEE}
}

@inproceedings{zhang2018unreasonable,
  title={The unreasonable effectiveness of deep features as a perceptual metric},
  author={Zhang, Richard and Isola, Phillip and Efros, Alexei A and Shechtman, Eli and Wang, Oliver},
  booktitle={Proceedings of the IEEE conference on computer vision and pattern recognition},
  pages={586--595},
  year={2018}
}

@article{mittal2012making,
  title={Making a “completely blind” image quality analyzer},
  author={Mittal, Anish and Soundararajan, Rajiv and Bovik, Alan C},
  journal={IEEE Signal processing letters},
  volume={20},
  number={3},
  pages={209--212},
  year={2012},
  publisher={IEEE}
}

@article{zhao2024image,
  title={Image and video tokenization with binary spherical quantization},
  author={Zhao, Yue and Xiong, Yuanjun and Kr{\"a}henb{\"u}hl, Philipp},
  journal={arXiv preprint arXiv:2406.07548},
  year={2024}
}

@ARTICLE{1284395,
  author={Zhou Wang and Bovik, A.C. and Sheikh, H.R. and Simoncelli, E.P.},
  journal={IEEE Transactions on Image Processing}, 
  title={Image quality assessment: from error visibility to structural similarity}, 
  year={2004},
  volume={13},
  number={4},
  pages={600-612},
  doi={10.1109/TIP.2003.819861}}

@INPROCEEDINGS{1292216,
  author={Wang, Z. and Simoncelli, E.P. and Bovik, A.C.},
  booktitle={The Thrity-Seventh Asilomar Conference on Signals, Systems \& Computers, 2003}, 
  title={Multiscale structural similarity for image quality assessment}, 
  year={2003},
  volume={2},
  number={},
  pages={1398-1402 Vol.2},
  doi={10.1109/ACSSC.2003.1292216}}

@article{jpegai,
  title={JPEG AI: The first international standard for image coding based on an end-to-end learning-based approach},
  author={Alshina, Elena and Ascenso, Joao and Ebrahimi, Touradj},
  journal={IEEE MultiMedia},
  volume={31},
  number={4},
  pages={60--69},
  year={2024},
  _publisher={IEEE}
}

@inproceedings{UVG,
  title={UVG dataset: 50/120fps 4K sequences for video codec analysis and development},
  author={Mercat, Alexandre and Viitanen, Marko and Vanne, Jarno},
  booktitle={ACM MMSys},
  pages={297--302},
  year={2020}
}

@article{mentzer2020high,
  title={High-Fidelity Generative Image Compression},
  author={Mentzer, Fabian and Toderici, George D and Tschannen, Michael and Agustsson, Eirikur},
  journal={Advances in Neural Information Processing Systems},
  volume={33},
  year={2020}
}

@article{li2021deep,
  title={Deep Contextual Video Compression},
  author={Li, Jiahao and Li, Bin and Lu, Yan},
  journal={Advances in Neural Information Processing Systems},
  volume={34},
  year={2021}
}

@inproceedings{tian2024free,
  title={Free-VSC: Free Semantics from Visual Foundation Models for Unsupervised Video Semantic Compression},
  author={Yuan Tian, Guo Lu, Guangtao Zhai},
  booktitle={European Conference on Computer Vision},
  pages={163--183},
  year={2024},
  organization={Springer}
}

@article{tian2026smc,
  title={SMC++: Masked Learning of Unsupervised Video Semantic Compression}, 
  author={Tian, Yuan and Ling, Xiaoyue and Geng, Cong and Hu, Qiang and Lu, Guo and Zhai, Guangtao},
  journal={IEEE Transactions on Pattern Analysis and Machine Intelligence}, 
  volume={48},
  number={2},
  pages={1992-2011},
  year={2026},
  publisher={IEEE}
}

@article{duan2020video,
  title={Video coding for machines: A paradigm of collaborative compression and intelligent analytics},
  author={Duan, Lingyu and Liu, Jiaying and Yang, Wenhan and Huang, Tiejun and Gao, Wen},
  journal={IEEE Transactions on Image Processing}, 
  volume={29},
  pages={8680--8695},
  year={2020},
  publisher={IEEE}
}

@article{chu2020learning,
  title={Learning temporal coherence via self-supervision for GAN-based video generation},
  author={Chu, Mengyu and Xie, You and Mayer, Jonas and Leal-Taix{\'e}, Laura and Thuerey, Nils},
  journal={ACM Transactions on Graphics (TOG)},
  volume={39},
  number={4},
  pages={75--1},
  year={2020}
}
